\UseRawInputEncoding
\documentclass[lettersize,journal]{IEEEtran}
\usepackage{amsmath,amsfonts}
\usepackage{algorithmic}
\usepackage{algorithm}
\usepackage{array}
\usepackage[caption=false,font=normalsize,labelfont=sf,textfont=sf]{subfig}
\usepackage{textcomp}
\usepackage{stfloats}
\usepackage{url}
\usepackage{verbatim}
\usepackage{graphicx}
\usepackage{cite}
\usepackage{booktabs} 

\begin{document}

\title{Implicit Guidance and Explicit Representation of
Semantic Information in Points Cloud: A Survey}

\author{Jingyuan Tang, Yuhuan Zhao, Songlin Sun, Yangang Cai

\thanks{J. Tang, Y. Zhao, S. Sun and Y. Cai are with the School of Information and Communication Engineering, Beijing University of Posts and Telecommunications (BUPT), Beijing, 100876, China; also with Key Laboratory of Trustworthy Distributed Computing and Service (BUPT), Ministry of Education, China. E-mail: tangjingyuan2018@bupt.edu.cn, zhaoyuhuan@bupt.edu.cn, slsun@bupt.edu.cn, caiyangang@bupt.edu.cn.(Corresponding author: Jingyuan Tang)}
 }



\maketitle

\begin{abstract}
Point clouds, a prominent method of 3D representation, are extensively utilized across industries such as autonomous driving, surveying, electricity, architecture, and gaming, and have been rigorously investigated for their accuracy and resilience. The extraction of semantic information from scenes enhances both human understanding and machine perception. By integrating semantic information from two-dimensional scenes with three-dimensional point clouds, researchers aim to improve the precision and efficiency of various tasks. This paper provides a comprehensive review of the diverse applications and recent advancements in the integration of semantic information within point clouds.

We explore the dual roles of semantic information in point clouds, encompassing both implicit guidance and explicit representation, across traditional and emerging tasks. Additionally, we offer a comparative analysis of publicly available datasets tailored to specific tasks and present notable observations. In conclusion, we discuss several challenges and potential issues that may arise in the future when fully utilizing semantic information in point clouds, providing our perspectives on these obstacles. The classified and organized articles related to semantic based point cloud tasks, and continuously followed up on relevant achievements in different fields, which can be accessed through \url{https://github.com/Jasmine-tjy/Semantic-based-Point-Cloud-Tasks}.

\end{abstract}


\section{Introduction}
\IEEEPARstart{T}{he} emergence of 3D acquisition devices \cite{Kähler7165673} and the increasing demand for high-dimensional data have led to a broad interest in the 3D representation of static and dynamic situations, as well as related activities. The current mainstream 3D representations include voxels \cite{Laine5620900}, point clouds \cite{Chia284492}, grids \cite{CATMULL1978350},\cite{Catmull1974ASA}, and neural radiation fields \cite{NeRF58452824}, etc. Among these, the depth sensor can directly capture the point cloud while maintaining the initial data in the three-dimensional scene. Point cloud representation is still the preferred method for many 3D scene activities and applications since it is robust and does not exhibit a significant topological correlation among points. Point clouds are widely used in domains including urban planning, autonomous vehicles, surveying, and mapping. Apart from fulfilling the requirements for 3D scene demonstrate, tasks including segmentation, registration, reconstruction, and compression are also performed in different scenarios, contributing to the advancement of 3D data processing technology.

In 2D image video content, semantic information can be used as a local or global condensed representation. Semantic information, on the one hand, delivers important information to humans in a more direct manner. On the other hand, for terminals with more detailed functions, it can help with machine perception and offer implicit guidance. Currently, two-dimensional representation is the subject of more in-depth research on semantic information than three-dimensional representation. Starting from semantics, this prior information can be used for high-dimensional feature processing such as image and video content retrieval \cite{ALZUBI201520},\cite{Mafla9423139}, content editing \cite{Chai10377456}, and target segmentation \cite{Gu10041185}. 
Semantic information serves as a high-level feature to offer implicit guidance for target functions of optimization, like compression \cite{Chang9428366}, image enhancement \cite{Qi9930878}, and image recognition \cite{Wang10091192}. The divergent exploration and continuous attempts of semantic information in two-dimensional representation make it possible to apply it in three dimensions. Semantic analysis of three-dimensional data has also gained increasing attention due to the construction of large-scale storage libraries and depth scanner data collecting.

Semantically driven tasks based on point cloud representation are still unexplored, despite the fact that semantic information in two-dimensional representation has been extensively examined. Due to the growing maturity of point cloud representation, researchers are also exploring the optimization of traditional point cloud tasks of semantic information, as well as the emerging tasks that may be introduced by combining it with point clouds. The inclusion of semantic information expands the point cloud's content, and the labels that each point receives aid in a more precise understanding and analysis of the three-dimensional scene. 
Furthermore, point cloud data that has been semantically labeled can be more efficiently applied to clustering, retrieval and other tasks, and provide clues for processing more complex three-dimensional data. 
\begin{figure*}[!t]
    \centering
    \includegraphics[width=7.0in]{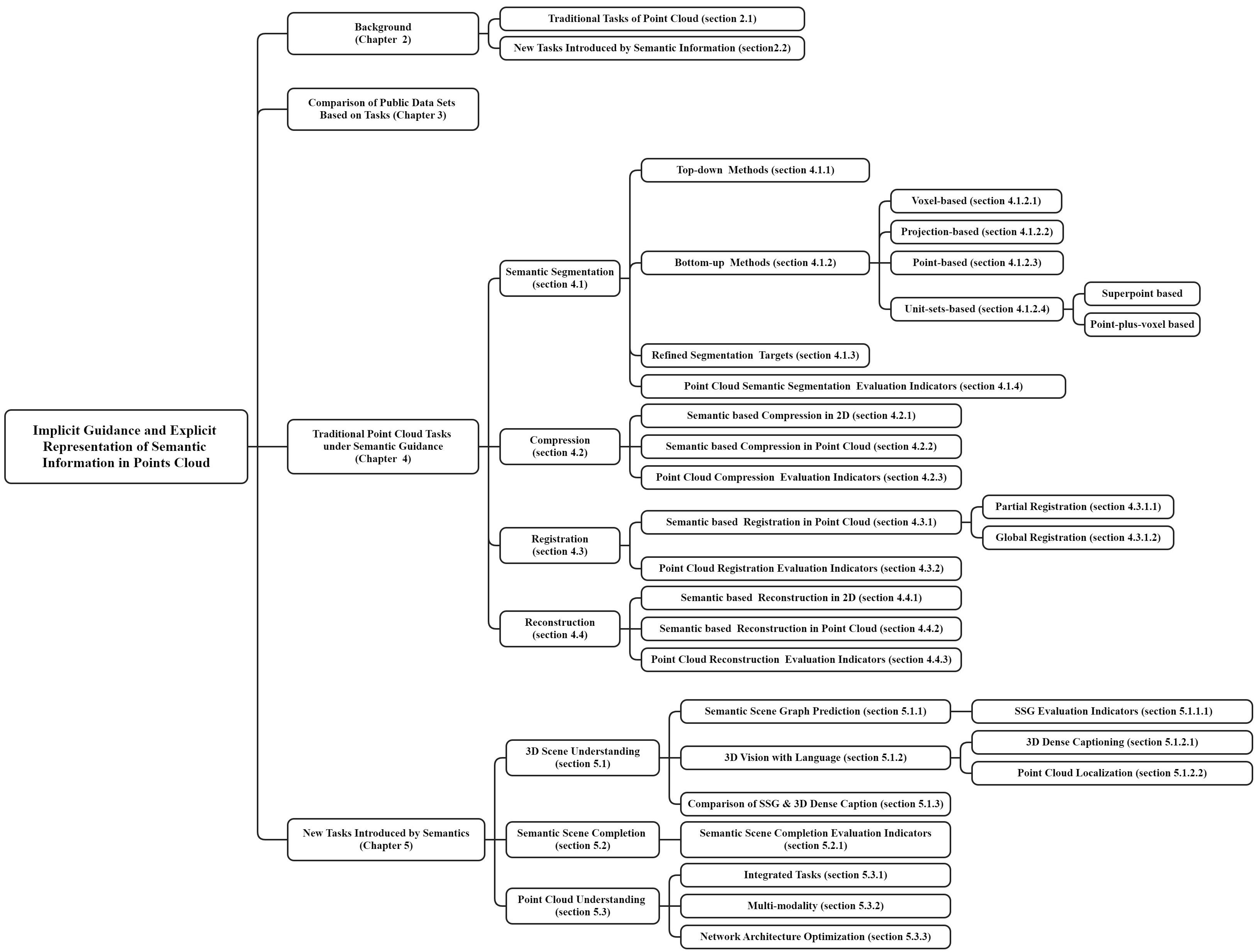}
    \caption{Taxonomy of existing methods in point cloud related to semantic.}
    \label{Organizational structure of this article}
\end{figure*}
Advancements in semantic information within point clouds include the introduction of semantic information into traditional tasks such as instance segmentation and object detection, enabling multi-scene semantic segmentation. Semantic guidance has been utilized to enhance efficiency in traditional tasks such as compression \cite{Zhao9690112}, registration \cite{Shu9860002} and reconstruction \cite{Wang252615362}. Additionally, emerging semantic-based point cloud tasks include 3D dense captioning \cite{Cai9879358}, scene graph prediction \cite{Wang10205194}, semantic scene completion \cite{Li10007036}, and point cloud understanding \cite{Xue10203465} are also burgeoning.

Despite significant advancements, the field of point cloud semantic-related tasks still lacks a comprehensive and systematic survey, hindering a clear and complete understanding of these tasks. The rapid development in this area necessitates a thorough review of the latest research. This paper addresses this need by providing an extensive review from the dual perspectives of implicit derivation and explicit representation, integrating semantic information with both traditional and emerging point cloud tasks.

Implicit derivation in point cloud semantics involves leveraging high-level semantic features to guide task implementation, achieving state-of-the-art performance compared to traditional methods. This category encompasses tasks such as point cloud compression, registration, reconstruction, and semantic scene completion. On the other hand, explicit representation refers to tasks where the final output includes visual semantic tags of the scene or descriptive semantic narration. This category includes semantic segmentation, 3D dense captioning, scene graph prediction, among others. This review aims to elucidate these aspects, offering a comprehensive overview of the current state and future directions of point cloud semantic-related tasks.
\begin{figure}[!t]
    \centering
    \includegraphics[width=3.0in]{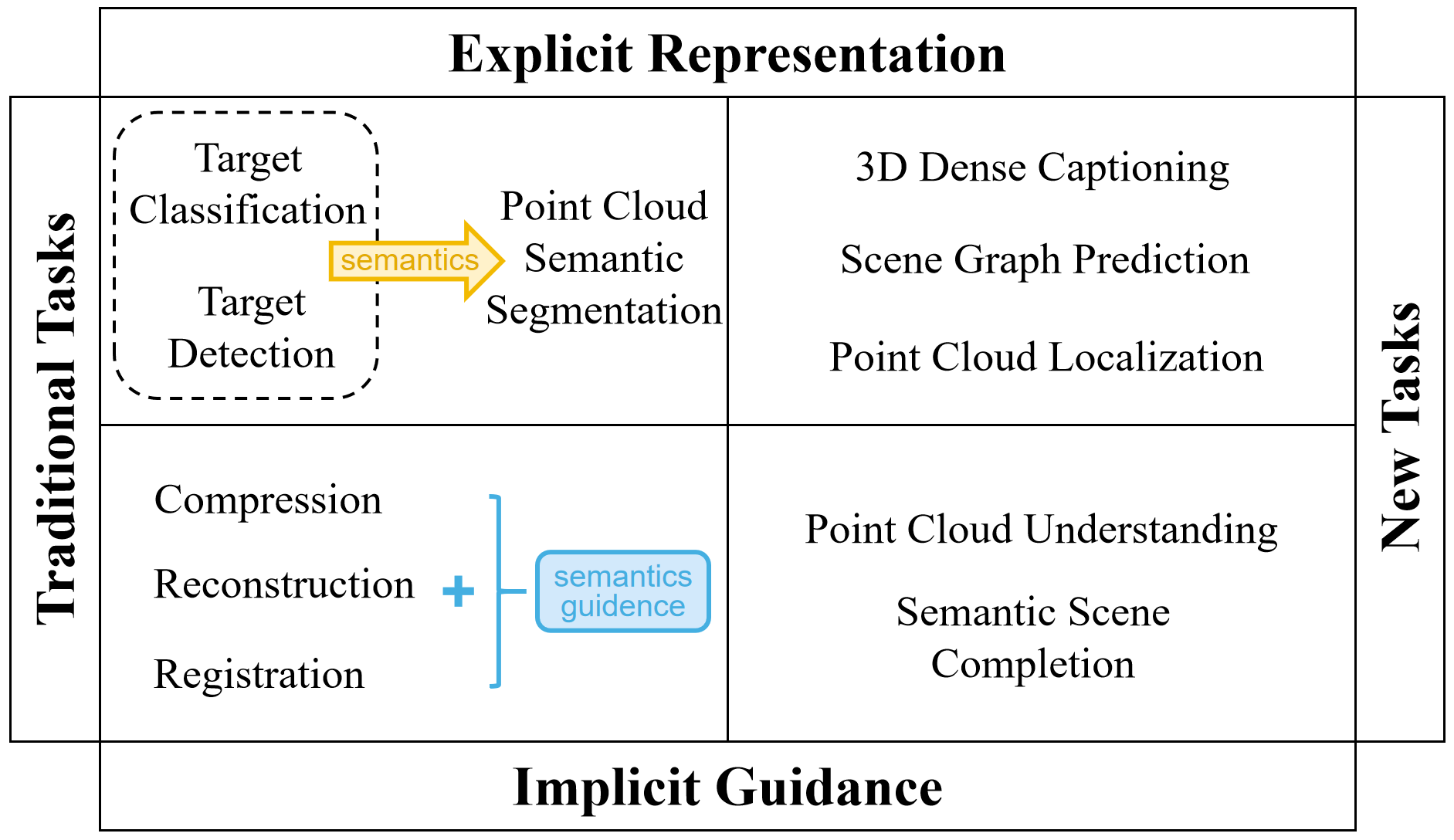}
    \caption{Implicit guidance and explicit representation of semantic information in points cloud.}
    \label{Classification of each task}
\end{figure}
The contributions of this study mainly include threefold. First, it diverges from traditional reviews that focus on specific fields by using "semantics" as a central theme to provide a comprehensive overview of the most advanced point cloud semantic-related tasks, analyzing the advantages and limitations of each area. To our knowledge, this is the first review to address the global semantic analysis of point clouds. Second, we summarize performance evaluation indicators for each task, facilitating multi-dimensional comparisons. Finally, by integrating advancements in 2D semantics with an analysis of the unique characteristics of point clouds, we aim to offer guidelines for future research in this field and discuss potential research directions.

The organizational structure of the remainder of this article is illustrated in Fig.\ref{Organizational structure of this article}. Further details will be discussed in the following subsections.

\section{Background}
As previously mentioned, discussions will be conducted in-depth  on the development of tasks in 3D point clouds from the perspectives of implicit derivation and explicit representation. This analysis is based on the integration of semantic information with both traditional and emerging point cloud tasks. Fig.\ref{Classification of each task} presents the classification of each task.

\subsection{\textbf{Traditional Tasks of Point Cloud} }
Traditional tasks in point cloud processing encompass target classification, detection, compression, registration and reconstruction. 
Point cloud \textbf{target classification} aims to categorize objects represented by a set of 3D points into predefined classes, typically using machine learning classifiers trained on extracted global feature vectors. Recent advancements in 3D point cloud classification have achieved remarkably high accuracy, with performance nearing saturation. However, the diversity of predefined classes and the continuous updating of object datasets may provide new impetus for further improvements \cite{Uy9009007}.

Point cloud \textbf{target detection} primarily employs 3D bounding boxes for annotating targets within a scene \cite{Shi9018080}, using cuboids to enclose detected objects. Unlike target classification, which focuses on distinguishing a single instance, target detection enables the recognition of multiple instances within a scene.
Point cloud \textbf{compression} involves reducing the volume of point cloud data by leveraging its geometric and attribute information, thereby facilitating efficient storage and transmission of the data \cite{cao20193d}.
Point cloud \textbf{registration} involves aligning point clouds captured from various perspectives into a unified coordinate system to create a complete scene point cloud. This process essentially involves measuring and rigidly transforming data points across different coordinate systems \cite{Qin10076895}. 
Point cloud \textbf{reconstruction} refers to the recovery of a continuous, closed 3D surface or shape from discrete point cloud data \cite{Far10301359}. Given that raw point cloud data often contains noise and may be non-uniformly distributed \cite{Huang20202413}, point cloud reconstruction remains a persistent challenge in traditional tasks.

Based on our survey of current research, the above traditional point cloud tasks utilize semantic information as either clues or final visualization outputs. Notably, point cloud target classification and detection have evolved into point cloud semantic segmentation. Meanwhile, traditional tasks such as point cloud registration,compression, and reconstruction continue to be driven by their original optimization goals. Researchers in these fields are integrating semantic attributes and deep learning techniques to achieve optimizations that surpass traditional methods.

\subsection{\textbf{New Point Cloud Tasks Introduced by Semantic Information}}

The growing interest in the intersection of 3D visual understanding \cite{Ding10203570}, natural language processing, and machine perception has spurred the development of new tasks in 3D point cloud representation. These new tasks, driven by semantic information, include 3D dense captioning, scene graph prediction, point cloud localization, semantic scene completion, and point cloud understanding.
The \textbf{3D dense captioning} task involves generating meaningful textual descriptions for detected objects. \textbf{Scene graph prediction} \cite{Chang9661322} leverages the "scene graph" concept from computer graphics to describe instances in 3D scenes and their interrelationships, with each node representing an object and each edge representing a relationship between objects. Although scene graph prediction also maps visual data to natural language, it is distinct from dense captioning, which is more object-focused and emphasizes precise descriptions of object appearance, often overlooking complex geometric relationships between instances. Despite their similar objectives, the differing foundational logics in 2D from which these tasks evolved warrant separate discussions.
\textbf{Point cloud localization} \cite{Xia149514967} Point cloud localization achieves specific location search in 3D point cloud scenes based on user natural language descriptions and instructions.
\textbf{Semantic scene completion} \cite{Li9435044} emphasizes that the semantic and geometric information of the entire scene are jointly inferred. This field has gained tremendous development momentum in recent years. Compared with other tasks, the semantic scene completion task has a relatively targeted dataset, and the details will be introduced in the next section. 
\textbf{Point cloud understanding} is a new integrative task based on traditional point cloud tasks that considers underlying feature sharing between multiple information dimensions (even across dimensions) and modalities to improve the flexibility of the model.

\begin{figure*}[!h]
    \centering
    \includegraphics[width=6.5in]{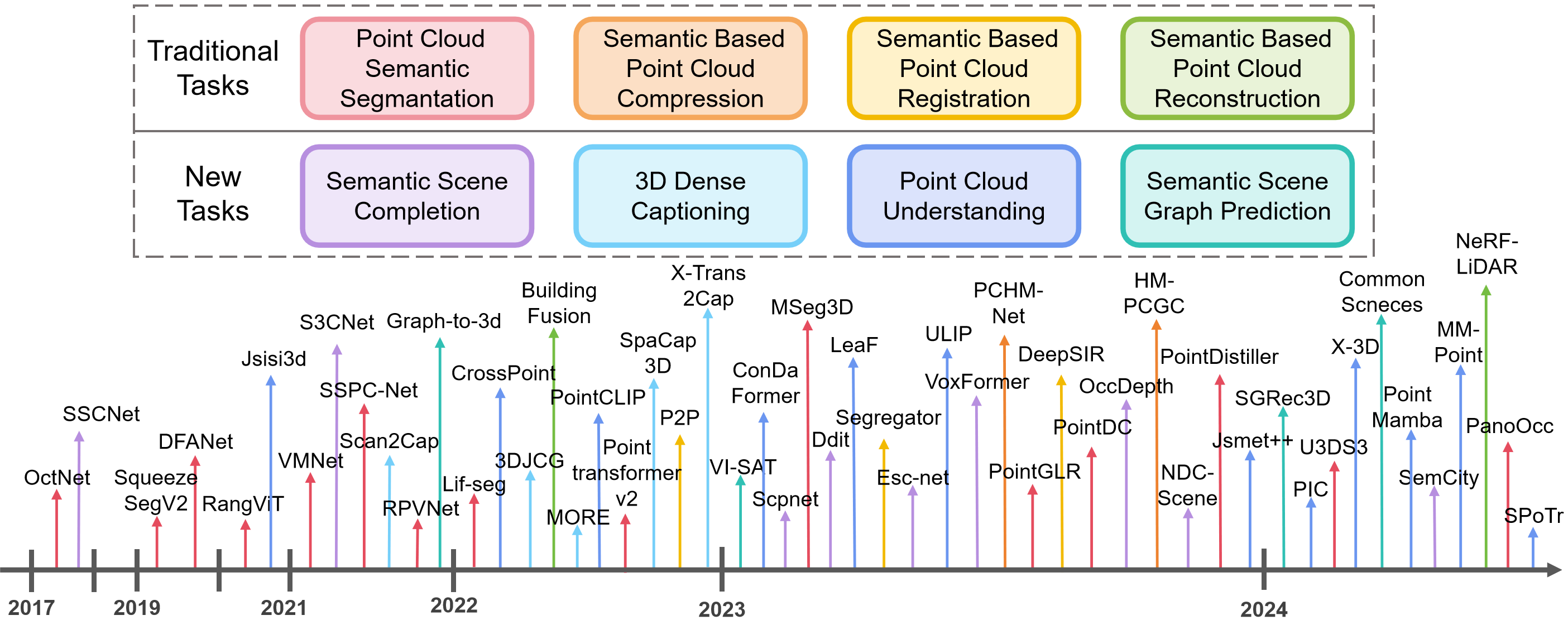}
    \caption{Technical development related to traditional tasks and new applications of point cloud based on semantic information.}
    \label{Fig:background}
\end{figure*}

This article aims to utilize semantic information to provide a comprehensive overview of both traditional applications and emerging tasks in point cloud processing. It focuses on the historical development of related technologies and the direction of new applications. The technological advancements are illustrated in Fig.\ref{Fig:background}.

\section{Comparison of Public datasets Based on Tasks}
In this section, we provide a summary of several widely-used datasets pertinent to various semantic tasks. 

\subsection{\textbf{Point Cloud Semantic Segmentation Related Dataset}}
\begin{itemize}
   \item Semantic3D
  
   The Semantic3D dataset \cite{hacke170403847} offers an extensive collection of annotated 3D point clouds for natural scenes, comprising over four billion individual points. This dataset encompasses a diverse array of urban environments, including churches, streets, railway tracks, squares, villages, football fields, and castles.
   \item SemanticKITTI
  
   The SemanticKitti \cite{behley929719307} dataset is a large-scale outdoor 3D point cloud dataset for autonomous driving applications, dedicated to 3D semantic segmentation tasks. It is built upon the Kitti dataset \cite{geiger12311237} through additional processing steps such as semantic segmentation.
   \item S3DIS
  
   The S3DIS dataset \cite{armeni15341543} comprises six extensive indoor areas, each encompassing various scene types, including offices, storage rooms, corridors, and conference rooms. The dataset includes approximately 273 million points annotated with 13 semantic labels. It is extensively utilized in semantic segmentation and instance segmentation tasks involving indoor point cloud data.

   \item SemanticPOSS
  
   The SemanticPOSS dataset \cite{pan687693} is specifically developed for 3D semantic segmentation tasks, offering a novel resource for this critical aspect of autonomous driving research. It comprises 2,988 complex LiDAR scans, along with a substantial amount of dynamic instance data. The data is formatted consistently with the SemanticKITTI dataset to enable seamless comparison and research across different datasets.

   \item SensatUrban
  
   The SensatUrban dataset \cite{Hu9578083} contains nearly three billion richly annotated 3D points. The dataset covers a large area of three UK cities, totaling approximately 7.6 square kilometers of urban landscape. In the dataset, each 3D point is classified into one of 13 predefined semantic categories.

   \item Matterport3D
  
    The Matterport3D dataset \cite{chang170906158} is primarily used for indoor scene understanding. It contains 10,800 panoramic views of 90 real-world building-scale scenes, constructed from 194,400 RGB-D images. Each scene is annotated with surface structure, camera positions, and semantic segmentation.

   \item SceneNN
  
   The SceneNN dataset \cite{hua92101} contains more than 100 different indoor scenes. These scenarios cover a variety of environments, such as offices, dormitories, classrooms, and food pantries. The dataset's semantic categories encompass 40 different classes. All scenes are reconstructed into triangle meshes, with each vertex and pixel annotated.

   \item ShapeNet

   ShapeNet \cite{Chang2015ShapeNetAI} contains 3D models from various semantic categories, providing extensive semantic annotations for each model, including consistent rigid alignment, parts, bilateral symmetry planes, physical dimensions, keywords, and other annotations. ShapeNet has indexed over 3 million  models, with 220 thousand classified into 3,135 categories.
\end{itemize}
\subsection{\textbf{Point Cloud Compression Related Dataset}}
    The dataset utilized for semantic-guided point cloud compression differs significantly from those used in conventional compression methods, primarily because it requires annotations within the point cloud. As a result, datasets from other tasks are often repurposed, including ModelNet40 \cite{wu19121920}, SemanticKITTI \cite{behley929719307}, and ShapeNet \cite{Chang2015ShapeNetAI}.
    
\subsection{\textbf{Point cloud Registration Related Dataset}}
\begin{itemize}
   \item ModelNet40
   
   The ModelNet40 dataset \cite{wu19121920} has been extensively employed in the domain of point cloud registration. Comprising 12,311 meshed computer-aided design (CAD) models encompassing 40 distinct categories of man-made objects. It is structured into training and test sets, with 9,843 shapes dedicated to training and 2,468 allocated for testing purposes, thereby enabling effective model training and subsequent assessment of performance metrics.

\end{itemize}
\subsection{\textbf{Point Cloud Reconstruction Related Dataset}}
\begin{itemize}
   \item ScanNetV2
  
    The ScanNetV2 dataset \cite{dai58285839} comprises 3D indoor scene reconstructions with extensive annotations, encompassing 1,613 3D scans across 20 different categories. This dataset is also divided into training sets and test sets. The benchmark evaluation targets 20 semantic categories, including 18 distinct object categories.

   \item VASAD
  
   The VASAD dataset \cite{langlois40084015} consists of six complete building models, each with detailed volume descriptions and semantic labels. Collectively, these models provides a substantial scale to facilitate the development and evaluation of learning-based methodologies. The VASAD dataset is designed to advance joint research efforts in point cloud surface reconstruction and semantic segmentation tasks.
\end{itemize}
\subsection{\textbf{Point Cloud Semantic Scene Graph Prediction Related Dataset}}
\begin{itemize}
   \item 3DSSG
  
   The 3DSSG dataset \cite{Wald9156565} is a large 3D dataset for learning 3D semantic scene graphs from 3D indoor reconstructions. This dataset extends 3RScan and contains annotations of semantic scene graphs covering relationships, attributes and category hierarchies. The data structure of 3DSSG is built by defining a set of tuples between nodes and edges, where the nodes represent specific 3D object instances in the 3D scan. These nodes are defined in terms of their semantics, a class hierarchy, and a set of properties that describe the visual and physical appearance of the object instance and its usability. The edges denote the semantic relationships between these nodes.
\end{itemize}

\begin{table*}[htbp]
\renewcommand{\arraystretch}{1.3}
\large
\centering
\caption{Task based Comparison of Benchmark Datasets}
\resizebox{\textwidth}{!}{
\begin{tabular}{ccm{6cm}m{5cm}c}
   \toprule
   Name & Representation & Scale\centering & Source\centering & Task \\
   \midrule
   Semantic3D\cite{hacke170403847} & Point & 4 billion points in 8 class labels\centering & Lidar\centering & Semantic segmentation and classification \\ 
   SemanticKITTI\cite{behley929719307} & Point & 23,201/20,351 scans with 4,549 million points from 28 classes\centering & Lidar\centering & Semantic segmentation, classification and compression \\
   S3DIS\cite{armeni15341543} & Point & 273 million points in 13 semantic labels\centering & RGB-D \centering & Semantic segmentation and classification \\
   SemanticPOSS\cite{pan687693} & Point & 216 million points, 2,988 scenes, 14 semantic labels\centering & Lidar\centering & Semantic segmentation and classification \\
   SensatUrban\cite{Hu9578083} & Point & 3 billion points, 13 predefined semantic classes\centering & RGB-D\centering & Semantic segmentation and classification \\
   Matterport3D\cite{chang170906158} & Mesh & 10,800 views from 90 scenes\centering & RGB-D\centering & Semantic segmentation and classification \\
   SceneNN\cite{hua92101} & Mesh & 100 scenes, 40 semantic classes\centering & RGB-D\centering & Semantic segmentation and classification \\
   ShapeNet\cite{Chang2015ShapeNetAI} & Mesh & More than 3,000,000 models, 3,135 categories\centering & CAD model\centering & Semantic segmentation, classification and compression\\
   ModelNet40\cite{wu19121920} & Mesh & 40 artificial objects categories\centering & CAD model\centering & Point cloud matching and compression \\
   ScanNetV2\cite{dai58285839} & lmage & 1,513 scenes, 20 semantic classes\centering & RGB-D\centering & Point cloud reconstruction\\
   VASAD\cite{langlois40084015} & Mesh & Approximate 62,000 \begin{math}m^2\end{math} of building floors\centering & CAD model\centering & Point cloud reconstruction\\
   3DSSG\cite{Wald9156565} & Image & 478 scenes, 534 classes\centering & RGB-D\centering & 3D semantic scene graphs prediction\\
   Nr3D\cite{achlioptas422440} & Mesh and image & 41,500 free-form natural (human) referential utterances\centering & CAD model and RGB-D\centering & 3D dense captioning\\
   ScanRefer\cite{chen202221} & Image & 800scenes, 11,046 objection, 51,583free-form natural (human) referential utterances\centering & RGB-D\centering & 3D dense captioning\\
   NYUv2\cite{silberman746760} & Image & 1,449 RGBD images consisting 464 diverse scenes across 26 scene classes\centering & RGB-D\centering & Semantic scene completion\\
   SUNCG\cite{song808816} & Image & 45,622 indoor scenes\centering & RGB-D\centering & Semantic scene completion\\
   Scan2CAD\cite{avetisyan26142623} & Mesh and image & 97,607 annotated keypoint pairs\centering & CAD model and RGB-D\centering & Semantic scene completion\\
   \bottomrule 
\end{tabular}
}
\end{table*}
\subsection{\textbf{Point Cloud 3D Dense Captioning Related Dataset}}
\begin{itemize}
   \item Nr3D
  
   The Nr3D dataset \cite{achlioptas422440} integrates three-dimensional object models from ShapeNet and ScanNet, encompassing a wide range of object categories. Each object model is provided with corresponding 3D point cloud data and semantic segmentation labels. The Nr3D dataset includes 41,500 free-form human descriptions, facilitating the enhancement of computational understanding and processing of objects within three-dimensional scenes.

   \item ScanRefer
  
   The ScanRefer dataset \cite{chen202221} is built based on ScanNet dataset. It is an important dataset focusing on the field of 3D scene understanding. It is dedicated to promoting the interaction between natural language and 3D scenes. The 3D scene understanding dataset. This dataset contains 11,046 objects in 800 ScanNet scenes with a total of 51,583 natural language descriptions. Each scene contains detailed information such as point clouds, geometric images, and semantic segmentation labels. The natural language description part involves instructions related to objects, locations, and actions in the scene. The goal of the ScanRefer dataset is to perform object localization directly in three-dimensional space through the use of natural language expressions so that machines can understand and execute these instructions.
\end{itemize}
\subsection{\textbf{Point Cloud Semantic Scene Completion Related Dataset}}
\begin{itemize}
   \item NYUv2
  
   The NYUv2 dataset \cite{silberman746760} consists of 1449 RGBD images, which include 464 different indoor scenes in 26 scene categories. Each image is densely labeled pixel-by-pixel, and each instance has a unique instance label. The support annotation of each image consists of a set of 3-tuple $(R_i, R_j, type)$, where $R_i$ is the area ID of the supported object, $R_j$ is the area ID of the supported object, and type indicates the direction of support . For example, the supporting object is a table, the supported object is a water glass, and the direction is upward.

   \item SUNCG

   The SUNCG dataset \cite{song808816} contains 45,622 indoor scenes. On a technical level, this dataset allows the acquisition of depth images and semantic scene volumes by setting different camera directions. As a large synthetic 3D scene dataset, SUNCG provides dense voxel annotation. This dataset is suitable for training algorithms to perform scene understanding tasks such as semantic segmentation, depth estimation, visual navigation, etc.

   \item Scan2CAD
  
   The Scan2CAD dataset \cite{avetisyan26142623} consists of 1506 ScanNet scans and 14225 CAD models from ShapeNet. This dataset contains 97,607 keypoint pairs that connect the CAD model and their corresponding objects in the scan. The CAD models in the dataset mainly include common objects in indoor scenes such as chairs, tables, and cabinets.
\end{itemize}

\section{Traditional Point Cloud Tasks under Semantic Guidance}
\subsection{\textbf{Point Cloud Semantic Segmentation}}
Point cloud semantic segmentation endeavors to transform the input 3D point cloud representation into a visualized output with a highlighted region of interest mask. This process involves a series of technical operations that achieve instance segmentation while assigning a class label to each point. Originating from instance segmentation and target detection, point cloud semantic segmentation encompasses both top-down and bottom-up methodologies. Given that each point in a 3D point cloud inherently possesses discrete attributes, bottom-up methods, which are based on clustering principles, have become predominant.

Furthermore, point cloud semantic segmentation must adapt to the requirements of various scenarios. Consequently, some researchers have optimized segmentation techniques to cater to diverse refined segmentation objectives. Beyond enhancing the semantic segmentation process, studies have also focused on aspects such as generalizability and real-time performance. Thus, the latter part of this section will analyze specific methods and models of point cloud semantic segmentation, guided by the overarching goals of the field.

\subsubsection*{\bf Top-down Semantic Segmentation Method}

In the domain of natural image processing, state-of-the-art methods predominantly adhere to a top-down paradigm \cite{NEURIPS2020_cd3afef9},\cite{Wang9536421}. Typically, this process involves detecting candidate instances followed by pruning through non-maximum suppression (NMS). However, in the context of 3D point cloud representation, top-down segmentation methods generally underperform compared to bottom-up approaches. Consequently, the majority of researchers concentrate on enhancing the technical performance of bottom-up methods, while exploration of top-down segmentation methods remains relatively limited.

In response to this context, Sun et al. \cite{Sun10030147} introduced an innovative top-down segmentation method termed NeuralBF. This approach generates instance suggestions based on a given query by leveraging spatial and semantic affinities for instance discrimination. NeuralBF redefines the point cloud segmentation problem as a point-to-point affinity calculation issue, employing a neural bilateral filter to model this calculation. Similar to conventional segmentation methods, NeuralBF utilizes the NMS method during the proposal pruning phase. Despite addressing the bottleneck in the proposal generation process, NeuralBF, as a top-down segmentation method, still encounters certain limitations.

\subsubsection*{\bf Bottom-up Semantic Segmentation Method}

In the context of bottom-up approaches, methods are typically categorized into voxel-based, projection-based, and point-based techniques. Recently, there have also been proposals for methods based on unit-sets, such as superpoints and point-plus-voxel combinations.

\begin{enumerate}
    \item Voxel-based Semantic Segmentation

    Voxel-based semantic segmentation involves initially converting unstructured point clouds into regular voxel forms, with each voxel encompassing a set of associated points. This conversion process can result in a loss of resolution. Dense voxel methods store data for each voxel, whereas sparse voxel methods store only non-empty voxel data to conserve space.

    VoxNet \cite{Matu7353481} represents one of the earliest attempts at voxel-based 3D deep learning, demonstrating notable performance across various three-dimensional representations such as LiDAR, RGBD, and CAD. Pure volume networks face cubic growth in voxel count, which severely limits the resolution of dense voxel methods due to memory constraints. To address this, Le et al. \cite{Le8579057} introduced an approach that uses a linearly expanded number of points to generate grid cells, integrating point and grid data to achieve computational memory savings and enhanced performance without necessitating high-resolution networks. Sparse voxel-based methods mitigate computational demands by disregarding empty voxels. Riegler et al. \cite{Rieg8100184} proposed OctNet, a convolutional network leveraging point cloud sparsity. This method dynamically adjusts the depth of the octree during recursive voxel segmentation based on data density, facilitating the processing of high-resolution point cloud inputs via deep learning. Since voxelization can overlook geometric complexities at scales smaller than the voxel size, Hu et al. 
    \cite{Hu9710530} proposed VMnet, which combines sparse voxel and grid characteristics for three-dimensional semantic segmentation. Despite voxel-based methods achieving high accuracy, their substantial memory and computational requirements hinder their application in large-scale scenarios. Moreover, Wang \cite{Wang171517168} et al. adopted a learning approach from coarse to fine, which solved the problem of memory overhead and significantly improved efficiency.

    To address the voxelization bottleneck in large-scale scenes, Tang et al. \cite{Tang220919732} introduced the SPVConv lightweight module, which enhances the accuracy of identifying small objects by simultaneously considering neighborhoods and local details. Zhang \cite{Zhang20010028} et al. proposed a voxel guided dynamic point network that utilizes prior knowledge from voxel features to guide high-quality spatial feature extraction.

    While voxel-based methods address data regularity issues to some extent, their computational complexity in handling large-scale dense point clouds poses limitations for practical applications.
    \item{Projection-based Semantic Segmentation}
    \label{Projection-based semantic segmentation} 
    
    In the early stages of 3D representation, a commonly adopted strategy involved performing dimensionality reduction on 3D data. Point cloud based semantic segmentation technology has inherited and developed this concept, attempting to transform a 3D point cloud dataset into multiple 2D planes via multi-view projection. After completing semantic segmentation on the 2D plane, the resulting annotations are remapped and reintegrated into the original 3D space using  reconstruction algorithm. Consequently, projection-based semantic segmentation technology is also referred to as a multi-view processing method, with related efforts documented in studies such as \cite{lawin2095107},\cite{Rong9430559}, \cite{Rong10158507}, and \cite{alons5435439}.

    Projection methods are primarily categorized into spherical projection and bird's-eye view projection (orthogonal projection). For instance, spherical projection has been utilized in researches such as \cite{Milio8967762}, \cite{Wu8462926}, \cite{Wu8793495}, \cite{xu28846293}, \cite{cortinhal207222}, and \cite{wang20185323} to achieve dimensionality reduction of data, though its applicability has been predominantly confined to LiDAR data, necessitating broader applicability. On the other hand, bird's-eye view projection, employed by studies like \cite{Rong10119167}, \cite{Jiang10220057}, \cite{ng20205242}, \cite{chen2033331}, and \cite{Liu10124335}, facilitates feature migration from 3D views to planes, though such method can encounter area distortion problems when dealing with large-scale data.

    Projection-based methods can effectively leverage processing algorithms with proven performance in the image domain. For example, Ando et al. \cite{Ando10204428} proposed RangViT, which combines visual large models (ViTs) \cite{dosov201011929} to harness their powerful representation learning capabilities, thereby enhancing semantic segmentation performance. Additionally, projection-based methods benefit from efficient computational performance, making them a crucial approach for achieving real-time performance.

    Despite their advantages, projection-based methods face two significant limitations. First, since the 2D view only provides an approximate representation of the original 3D scene, geometric structure loss occurs during the reconstruction process. Second, in complex scenes, occlusion problems substantially increase the difficulty of segmenting small objects.
    \begin{figure}[t]
    \centering
    \includegraphics[width=3.4in]{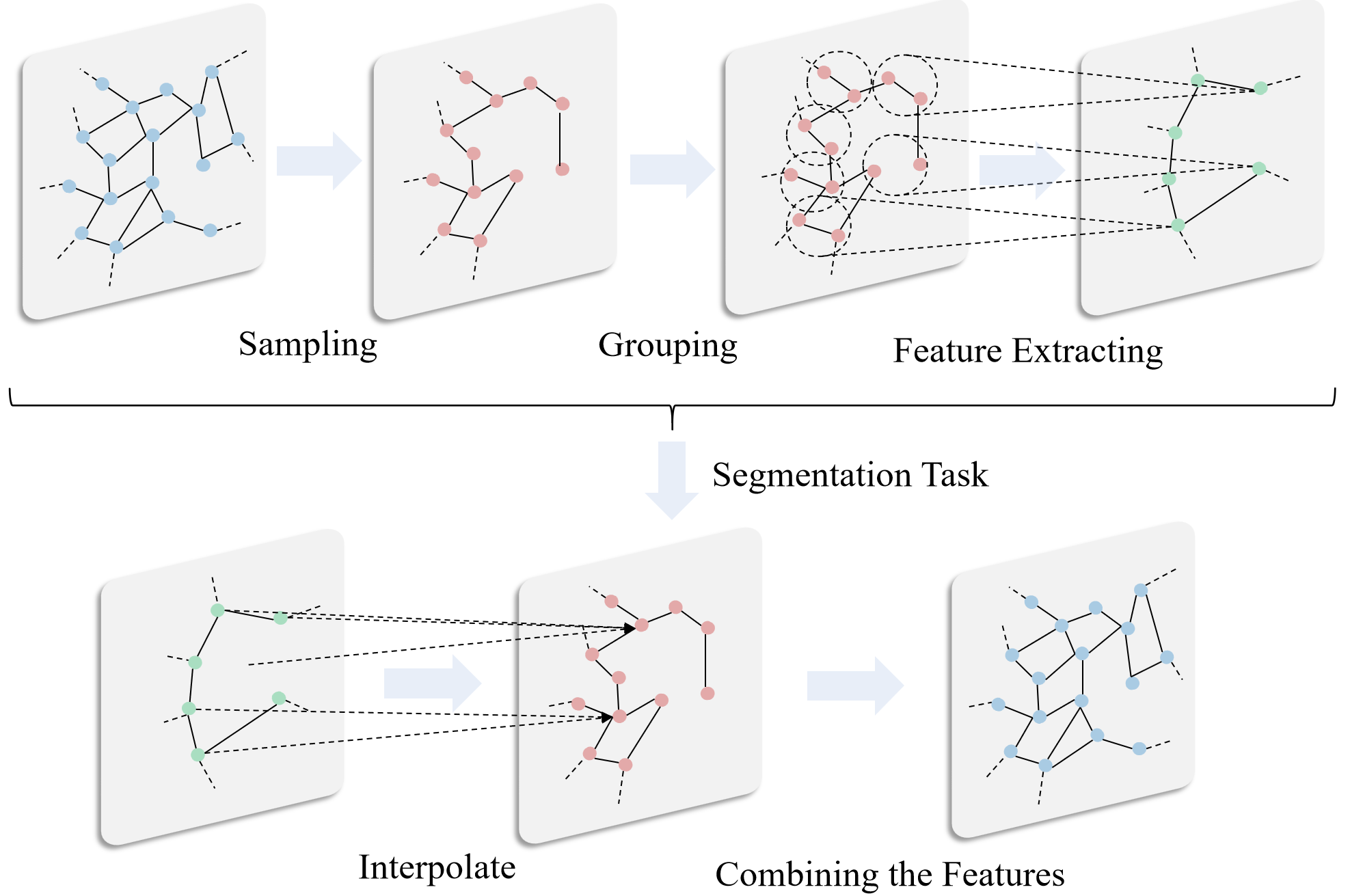}
    \caption{Multi level feature extraction structure.}
    \label{Fig:pointnet}
    \end{figure}
    \item{Point-based Semantic Segmentation}
    
    The unstructured nature of point cloud data challenges the application of 3D Convolutional Neural Networks (CNNs). To address this, researchers map point clouds to regular structures like multi-view projections and voxelized grids for CNN compatibility, though this can cause information loss and higher computational costs.

    PointNet \cite{Char8099499}, the first deep learning framework for direct point cloud processing, uses a multi-layer perceptron (MLP) and T-Net to ensure model invariance to spatial transformations. While PointNet excels in classification by extracting global features via maximum pooling, it struggles with capturing local features, complicating complex scene analysis. PointNet++ \cite{Qi5989836} addresses this by proposing a multi-level feature extraction structure, capturing both local and global features. This framework, involving point cloud downsampling, local domain querying, encoding, upsampling interpolation, and decoding as shown in Fig.\ref{Fig:pointnet}, has inspired many subsequent networks, advancing point cloud analysis. For instance, RandLA-Net \cite{Hu9156466} accelerates computation with random sampling and a more complex local feature encoding module to compensate for geometric detail loss. Point Transformer \cite{Zhao9710703},\cite{Wu20228ece66},\cite{wu20246357} enhances the encoding module with a self-attention mechanism to capture the hidden space of the point cloud.

    In CNNs, convolution extracts local features through receptive fields and weight sharing, reducing parameters and computational complexity. Suitable convolution operations are key for performance enhancement. A-CNN \cite{Koma8953826} uses circular convolution to capture geometric features, while Thomas et al. \cite{Thom9010002} proposed KPConv (rigid) and its deformable version to adapt to geometric shapes of scene objects.

    Additionally, local structures in point cloud data are crucial for analysis. Specific clustering strategies enhance model performance by identifying and extracting these structures. Researchers \cite{Lu9577467},\cite{Yin10025770},\cite{Shuai9410334} proposed optimized aggregation methods for improved clustering efficiency and accuracy.
    
    \item{Unit-sets-based on Semantic Segmentation}
    
    Combining the complementary information-preserving advantages of different unit representations is a major reason why researchers explore semantic segmentation based on unit sets. This comprehensive multi-source information strategy significantly improves the accuracy and credibility of semantic segmentation, positioning unit set-based semantic segmentation as a new focus in the research field.
    There are two main shortcomings in point-by-point semantic segmentation methods: first, the primary segmentation target is not effectively used to supervise the point-by-point clustering process; second, point-by-point feature learning and clustering may lead to segmentation and fragmentation issues when dealing with irregular data \cite{Liang9709996}.
    \begin{itemize}
        \item{Superpoint based}
            \begin{figure}[t]
            \centering
            \includegraphics[width=3.2in]{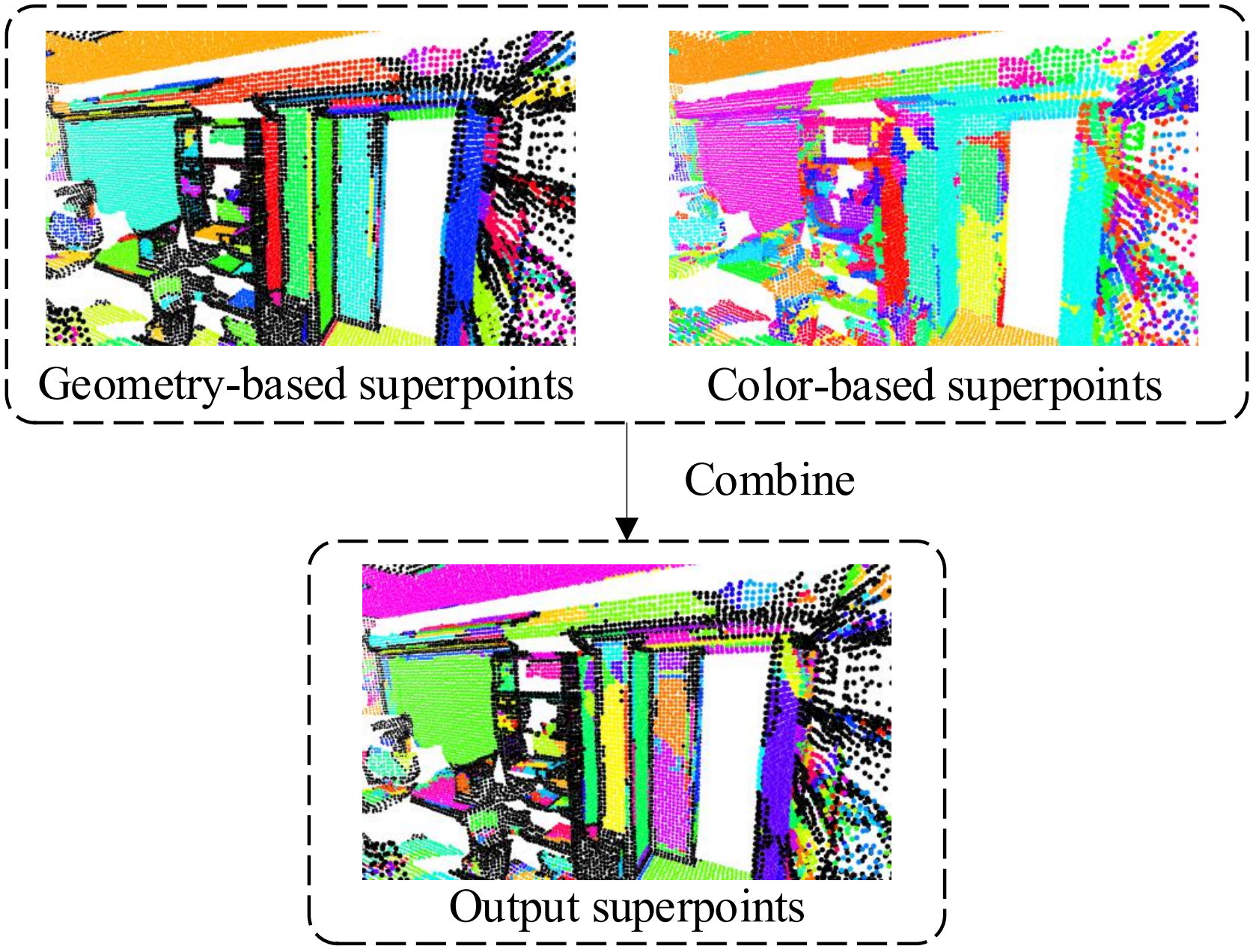}
            \caption{A process of generating superpoints based on geometry and color features\cite{Deng9811904}.}
            \label{Fig:superpoint}
            \end{figure}
            
         To address these challenges, researchers have explored grouping local feature similarity points into sets known as superpoints. Superpoint generation methods vary: Liu et al. \cite{Shi202106931} use spectral clustering, while Deng et al. \cite{Deng9811904} combine geometric and color information through region growing as shown in Fig.\ref{Fig:superpoint}. Superpoint-based semantic segmentation often incorporates unsupervised \cite{Chen1429920}, semi-supervised \cite{Cheng1620350},\cite{Deng9811904}, and weakly supervised \cite{Liu9578763} learning approaches, leveraging limited annotated data to reduce computational complexity and improve segmentation quality. The supervised learning aspects of point cloud semantic segmentation are discussed in \ref{Refined segmentation goal}.

        Landrieu et al. proposed a superpoint graph (SPG) structure for semantic segmentation of large-scale point clouds \cite{Land8578577}, which represents relationships between instance parts and context compactly. Yang et al. \cite{Yang9919408} introduced the Auto-NestedNet architecture, learning optimal nested architectures for point cloud representation using multi-layer features. Recently, Zheng et al. \cite{Zheng3626449} enhanced semantic segmentation performance using multi-scale superpoint networks, obtaining small-scale superpoints based on feature similarity and employing an attention mechanism to upsample and restore resolution, thus propagating multi-scale geometric features from low to high resolution.
        \item{Point-plus-voxel based}

        Point-based representation offers high geometric accuracy but struggles with local neighbor identification due to its disordered nature, while voxel-based representation is more regular but computationally intensive at higher resolutions. In large-scale point cloud segmentation, voxel-based methods typically outperform point-based methods \cite{Xu9709941}. Combining voxel and point features can enhance segmentation quality by providing coarse- and fine-grained spatial information. A comparison of point-based and voxel-based methods is shown in \ref{Fig:point_vs_voxel}
        \begin{figure}[!t]
        \centering
        \includegraphics[width=3.0in]{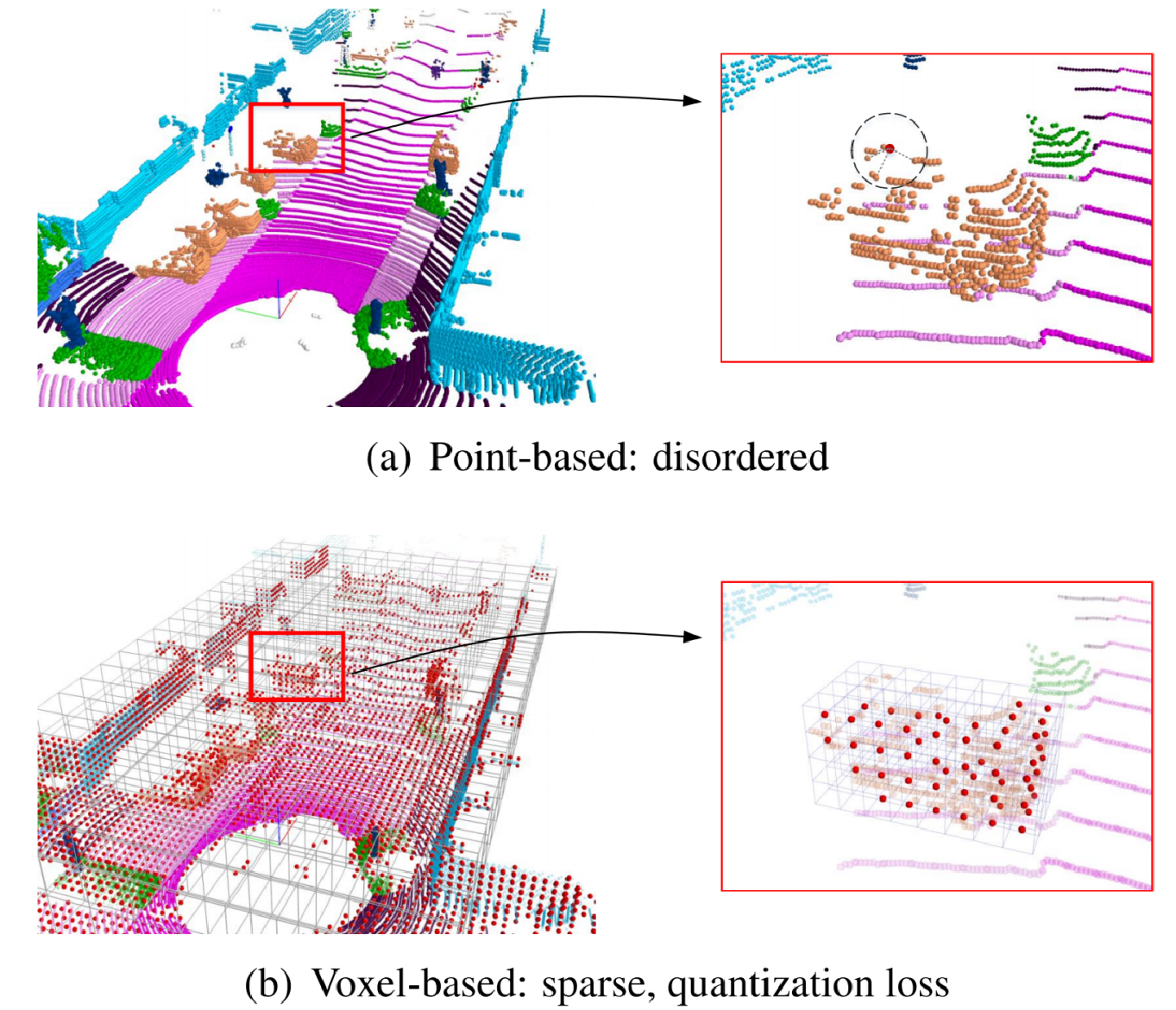}
        \caption{Point-based vs. voxel-based comparison\cite{Xu9709941}.}
        \label{Fig:point_vs_voxel}
        \end{figure}
        The PVCNN model by Liu et al. \cite{Liu7370345} reduces memory consumption by using points as inputs and voxels for efficient neighborhood access in convolution. Xu et al. \cite{Xu9709941} introduced RPVNet, which fuses features from projections, points, and voxels, significantly improving semantic segmentation performance.

        Supervoxels, generated through region growing from seed points \cite{Li9730613}, cluster similar features for better local area representation. Hou et al. \cite{Hou9879674} utilized point-to-voxel knowledge distillation (PVD) to transfer knowledge and compress models, capturing detailed perceptual information at the point level and broader features at the voxel level.

        Furthermore, during the clustering process of point cloud semantic segmentation, dense areas may receive disproportionate attention, leading to the neglect of sparse areas. Chen et al. \cite{Chen1429099} addressed density inconsistency in point cloud segmentation with the PointDC framework, which uses cross-modal distillation and supervoxel clustering. PointDC converts multi-view features into point-based representations and iteratively optimizes semantic features via supervoxel pooling. This approach highlights the effectiveness of knowledge distillation in cross-modal knowledge mapping, as seen in works by \cite{Jiang10220057},\cite{Qiu10160496}, and \cite{Zhang10205029}.
    \end{itemize} 

    In summary, semantic segmentation methods based on unit sets allow pipelines to adaptively segment point clouds based on their spatial complexity, enabling deep learning architectures to mine more refined and long-range interactions. In addition, the linkage between unit collections and different deep learning models also shows certain regularity. Finally, in the process of constructing the local domain, regardless of supervoxels or superpoints, one should try to avoid using only distance constraints because it is impossible to ensure that points located in the constructed domain have the same semantics. 
\end{enumerate}

\subsubsection*{\bf Refined Segmentation Goal}
\label{Refined segmentation goal}
\begin{figure}[!t]
\centering
\includegraphics[width=3.2in]{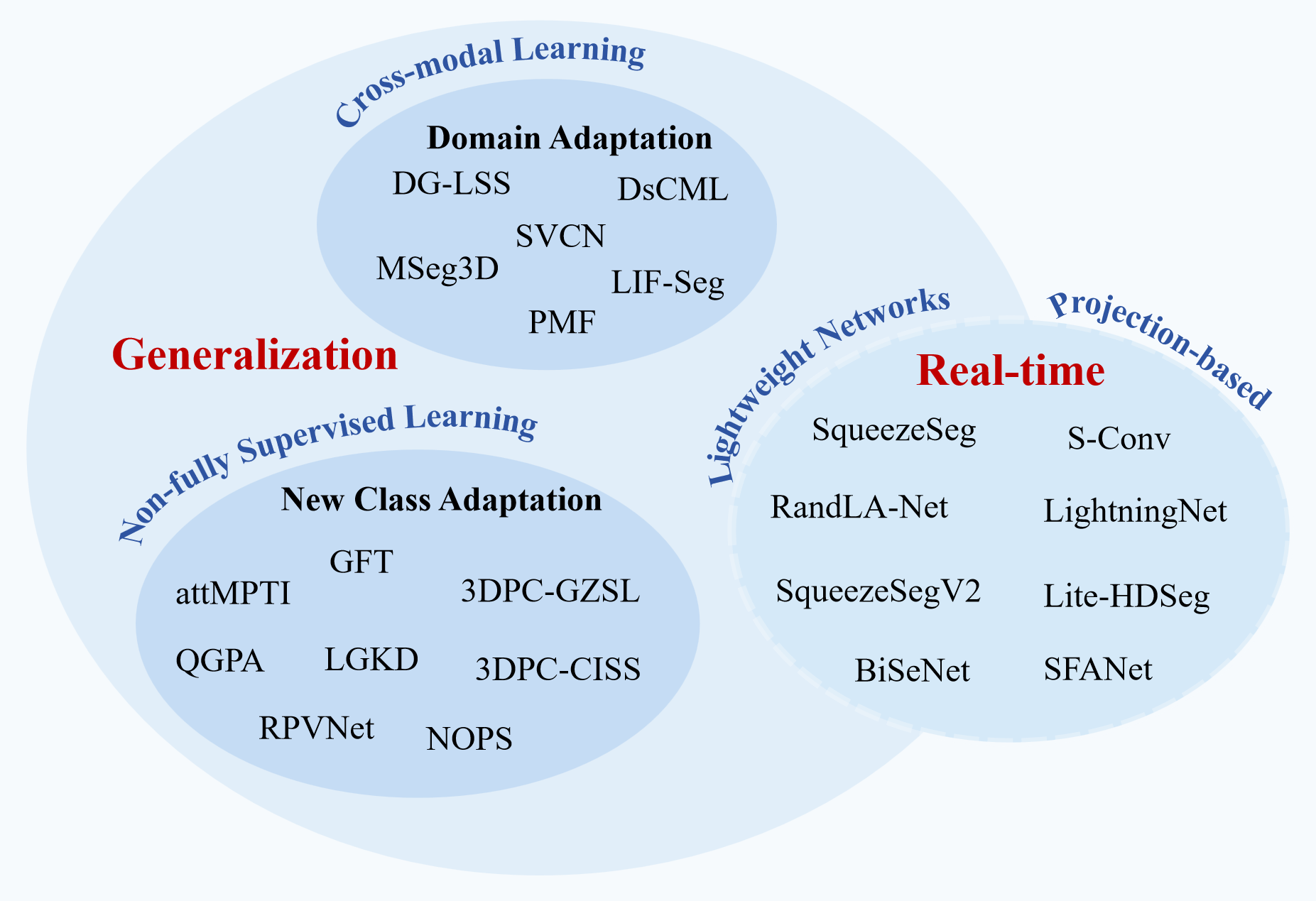}
\caption{The main goals of point cloud semantic segmentation optimization.}
\label{Fig:target}
\end{figure}
In the evaluation of point cloud semantic segmentation methods, an analysis based on refinement goals provides a complementary perspective, enabling researchers to more effectively identify and compare the performance of various models in achieving specific segmentation objectives. As illustrated in Fig.\ref{Fig:target}, we categorize the optimization of point cloud semantic segmentation into two primary directions: generalization and real-time performance. This classification method helps highlight excellent models for specific tasks, provides researchers with a framework for selecting appropriate segmentation methods based on application requirements.
\begin{enumerate}
    \item Generalization of Segmentation

    The generalization goal of point cloud semantic segmentation is to enhance the adaptability of segmentation algorithms to diverse point cloud data inputs and new, previously unprocessed categories.
    \begin{itemize}
        \item Domain Adaptation

        Compared to RGB camera-based technologies, LiDAR-only semantic segmentation methods exhibit greater robustness under varying lighting conditions due to LiDAR’s accurate depth information. However, differences in 3D sampling modes across LiDAR datasets \cite{behley929719307},\cite{chang87488757},\cite{geyer20205223},\cite{huang954960} and changes in sensor configuration can hinder model adaptation, necessitating domain adaptation methods to utilize new unlabeled data and transfer knowledge from labeled data, thus avoiding re-labeling costs.

        Yi et al. \cite{Yi9578920} addressed domain adaptation by converting raw LiDAR data into three-dimensional surface representations, enhancing adaptability. Li et al. \cite{Li10203290} tackled challenges in multi-modal models, such as mode heterogeneity and limited sensor interaction, proposing a method that combines intra-modal feature extraction with inter-modal feature fusion.

        For devices with both RGB and LiDAR sensors, fusing these modalities improves segmentation accuracy. Datasets like KITTI \cite{Geiger6248074} and nuScenes \cite{Caesar9156412} enable this fusion. Zhuang et al. \cite{Zhuang9710693} supplemented depth information by mapping point cloud data to the RGB domain and used a dual-stream network for feature complementation. To address feature density differences between 2D RGB and 3D LiDAR data, Peng et al. \cite{Peng9710520} proposed a dynamic sparse-to-dense cross-modal learning method (DsCML), Zhao et al. \cite{Zhao10128757} developed a coarse-to-fine network LIF-Seg, and Jaritz et al. \cite{Jaritz9737217} introduced new unsupervised cross-modal loss functions, enhancing feature fusion.
        
        To further research, Saltori et al. \cite{Saltori196206} established a domain generalization testbed for LiDAR semantic segmentation (DG-LSS) using diverse datasets and configurations, alongside the baseline LiDOG, to develop more robust models.
        
        \item New Class Adaptation
        
        Deep neural networks excel in semantic segmentation but face challenges due to limited annotated 3D point cloud data, which is costly to obtain. Current methods assume closed-set scenarios, struggling with new, minimally represented categories emerging over time, prompting research into few-shot, zero-shot, and 3D open-vocabulary learning \cite{wu2024632} and class incremental method \cite{Ding70107019}. 
        
        To adapt to new categories, approaches by \cite{Zhao9577428},\cite{Yang1158611596}, and \cite{He10138737} use limited labeled data for segmentation. The disorderly 3D point clouds exacerbate catastrophic forgetting \cite{french128135},\cite{robins123146}, tackled by Geometry-aware Feature Transfer (GFT) \cite{Yang10204829} and Label-Guided Knowledge Distillation (LGKD) \cite{Yang1860118612}. RegionPLC \cite{Yang19829832} effectively builds comprehensive regional point language pairs and performs well in challenging long tail or unannotated scenarios.
        
        Non-fully supervised methods reduce parameters and computational load, maintaining performance standards. Semi-supervised learning \cite{Yang10204829},\cite{Riz10203892},\cite{Kong10205234},\cite{Li10203638},\cite{Xu1809818108} blends limited labels with unlabeled data, while weakly supervised approaches \cite{Li9879771},\cite{Liu1841318422},\cite{Tao9833393},\cite{Chang10035004},\cite{Wei10328634} and unsupervised techniques \cite{Wu8793495},\cite{Zhang10203698},\cite{Rao9736689},\cite{poux2019213},\cite{liu37593768} show promise in semantic segmentation tasks. 
    \end{itemize}
\item Real-time Segmentation

Most point cloud semantic segmentation methods rely on pre-trained models but face challenges in speed, hindering their deployment on mobile platforms. To address this, research focuses on achieving real-time semantic segmentation of point clouds.

Efforts to enhance real-time performance include strategies like reducing image resolution or designing lightweight networks \cite{Yang9102769},\cite{Razani9561171},\cite{kang20211960},\cite{wang20185323} to trim model parameters and sampling dimensions, thus easing computational load. As discussed in Section \ref{Projection-based semantic segmentation}, projection-based techniques leverage mature two-dimensional image processing to lower data complexity, often training models on projected point cloud data for real-time implementation. However, this approach sacrifices contextual and spatial details, impacting segmentation accuracy. Balancing inference speed with segmentation precision remains a central challenge in the field. Recent studies, including SFANet \cite{Weng9583294}, BiSeNet \cite{yu325341}, and DFANet \cite{Li8954459}, employ multi-branch frameworks to integrate high-level and low-level features, aiming to resolve this trade-off effectively.
\end{enumerate}

\subsubsection*{\bf Point Cloud Semantic Segmentation Evaluation Index}
\label{Point cloud semantic segmentation evaluation index}

Average Intersection over Union (mIoU) and Overall Accuracy (OA) are two core indicators when evaluating point cloud semantic segmentation tasks.
\begin{itemize}
  \item mIoU

    mIoU quantifies segmentation performance by calculating the ratio of the intersection and union of the predicted region and the actual region. Specifically, the Intersection over Union (IoU) ratio reflects the degree of overlap between the predicted segmentation area and the real segmentation area. mIoU further provides a comprehensive evaluation by calculating the IoU of each category and taking its average. It is calculated as follows:
    \begin{equation}
  IoU=\frac{TP}{T+P-TP} 
\end{equation}
\begin{equation}
  mIoU=\sum_{i}^nIoU_i 
\end{equation}
Among them, $TP$ represents the positive samples correctly classified by the model, $T$ represents the number of all samples that are actually positive classes, $P$ represents the total number of samples predicted by the model as positive classes, and $n$ represents the number of categories.
\item OA

As a basic evaluation index, OA measures the probability that the sample label predicted by the model is consistent with the real label. It provides a global accuracy measure and is calculated as follows:
\begin{equation}
    OA=\frac{TP+TN}{TP+TN+FP+FN}  
  \end{equation}
  Among them, $TP$ represents the positive sample correctly classified by the model, $FN$ represents the positive sample misclassified by the model, $FP$ represents the negative sample misclassified by the model, and $TN$ represents the negative sample correctly classified by the model.
\end{itemize}

\subsection{\textbf{Semantic-based Point Cloud Compression}}
High-precision sparse point cloud data provide accurate navigation functionalities for autonomous driving systems, while dense point cloud data facilitate the creation of high-fidelity immersive experiences. As the demand for higher precision and denser point cloud data increases, the role of point cloud compression in content generation and transmission becomes increasingly critical. In point cloud compression, both the Video-Based Point Cloud Compression (V-PCC) algorithm for dense point clouds \cite{3dg18030} and the Geometric Point Cloud Compression (G-PCC) algorithm for sparse point clouds \cite{3dg2019text} rely extensively on the rate-distortion optimization (RDO) principle to enhance compression efficiency \cite{Cao9457097}. However, this approach has not fully exploited the semantic information in ordered point cloud data, nor has it effectively adapted and optimized for specific tasks. Moreover, traditional point cloud compression methods primarily aim to achieve visual fidelity for human vision tasks, whereas machine vision tasks prioritize semantic fidelity. Research in this domain remains in its nascent stage.
\begin{figure*}[!t]
\centering
\includegraphics[width=6.5in]{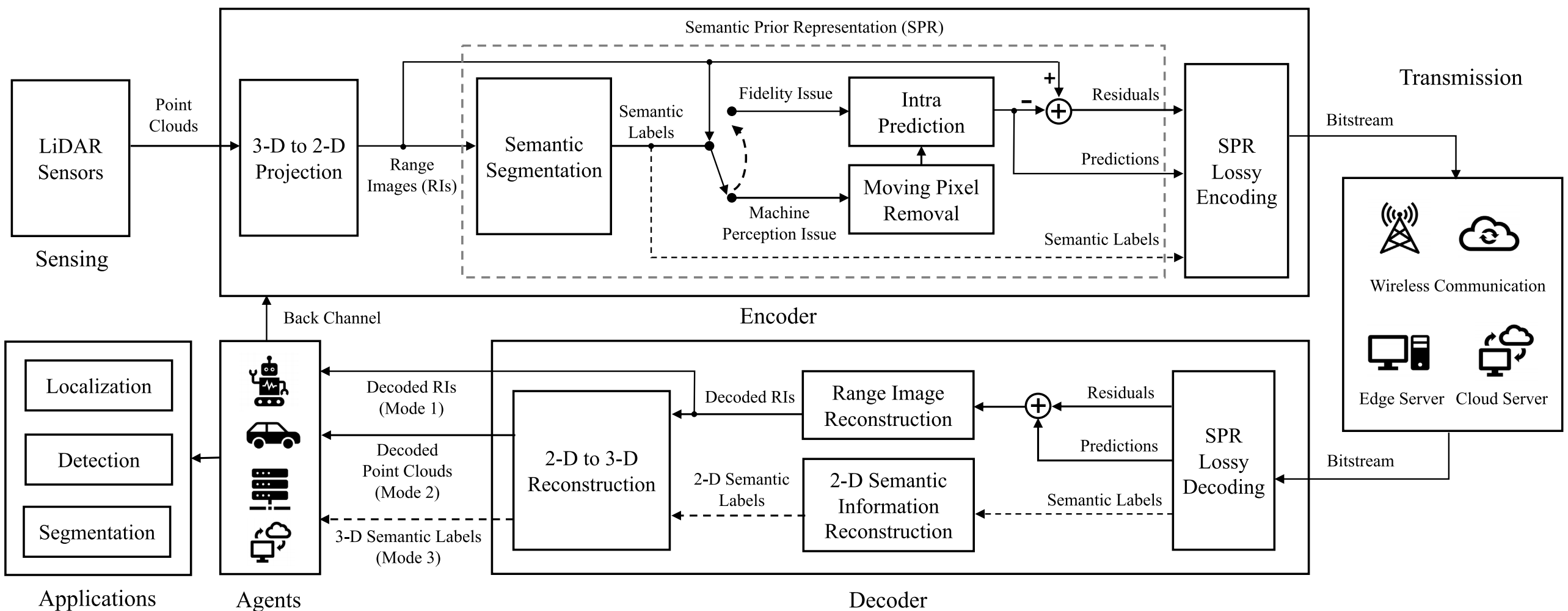}
\caption{The flowchart of real-time scene-aware LiDAR PCC system \cite{Zhao9690112}.}
\label{Fig:scene-aware LiDAR PCC system}
\end{figure*}
\subsubsection*{\bf Research Status of Semantic Combined Compression in 2D}

The application of compression in the context of two-dimensional (2D) semantic information aims primarily at achieving efficient hierarchical transmission and adapting to the requirements of machine vision tasks. This approach can offer valuable insights for the compression of three-dimensional (3D) scenes.

Scalable compression has been demonstrated as an effective method for data representation in 2D contexts, wherein visual signals are encoded into multiple layers. Zhang et al. \cite{Zhang10032603} and Akbari et al. \cite{Akbari8683541} proposed a framework for scalable cross-modal compression that hierarchically represents signals based on semantic segmentation. In this framework, each layer of the scalable flow enhances interpretability through the preceding layer. Similarly, Chang et al. \cite{Chang9428366} utilized semantic segmentation maps as a structural guide to extract deep semantic priors, thereby providing stronger capabilities for detail construction and greater flexibility for advanced visual tasks. This approach achieves a compression ratio of up to 1000:1 while maintaining high-quality visual reconstruction and robust applicability to visual task analysis.

Huang et al. \cite{Huang9894405} introduced a human-computer-friendly video compression scheme, namely Human-Machine Friendly Video Compression (HMFVC) based on Learning Semantic Representation (LSR). This solution establishes an end-to-end video compression framework by extracting semantic information from temporally adjacent frames, optimizing human visual perception quality, machine analysis accuracy, and compression efficiency in a collaborative manner.

In the realm of 2D representation, several initial attempts at compression processing based on semantic information have been made, demonstrating significant potential in enhancing performance.

\subsubsection*{\bf Research Status of Point Cloud Compression based on Semantics}

Recent point cloud compression architectures primarily aim to produce higher quality reconstructed point clouds while optimizing compression efficiency \cite{Akhtar10380494},\cite{Gao10416804},\cite{Song10205051},\cite{Ngu10024999},\cite{Zhou9878865}. However, these methods may not be the most suitable for machine vision tasks. Many image coding technologies designed for machine vision enhance performance specifically for these tasks, often compromising human visual performance. To address the need for a balance between human and computer vision tasks in compression schemes, Ma \cite{Ma10222524} and Liu \cite{Liu10219641} proposed innovative solutions. 

The HM-PCGC method \cite{Ma10222524} employs a pre-trained compression backbone network and a semantic mining module to aggregate multi-task features, achieving superior performance compared to traditional methods. The PCHM-Net \cite{Liu10219641} utilizes a dual-branch structure to simultaneously accomplish two tasks, constructing a comprehensive graph based on sparse point sets to aggregate global information, thereby achieving a better trade-off between the two tasks. Zhao et al. incorporated the scalable coding concept from two-dimensional representation \cite{Zhao9690112}, using semantic prior representation (SPR) and a variable precision lossy coding algorithm to generate the final bit stream, as illustrated in Fig.\ref{Fig:scene-aware LiDAR PCC system}, facilitating real-time compression of point clouds.

In addition to enhancing machine vision and human vision tasks, semantic compression technology also significantly contributes to improving compression efficiency. Sun \cite{Sun9944923} proposed a task-driven sparse point cloud coding framework, SA-LPCC, for autonomous driving scenarios. This framework employs a semantic segmentation network to identify and temporarily shield moving objects, eliminating spatial and temporal redundancy and enhancing compression efficiency. Liu \cite{Liu3654800} et al. introduced a unified framework for jointly compressing various visual and semantic data, including images, point clouds, segmentation maps, object attributes, and relationships. By embedding these data representations into a joint embedding graph based on categories, this framework allows flexible handling of joint compression tasks for diverse visual and semantic data.

Despite significant advancements in semantic-based point cloud compression, existing solutions have not yet achieved comprehensive coverage of all point cloud compression application scenarios compared to traditional compression technologies. Furthermore, the focus of current solutions remains relatively limited in scope.

\subsubsection*{\bf Point Cloud Compression Evaluation Index}
\label{Point cloud compression evaluation index}
Similar to traditional point cloud compression, point cloud compression evaluation indicators based on semantic information can also be measured using compression efficiency and fidelity, but machine vision-oriented solutions need to additionally consider the fidelity of machine vision.

Compression efficiency is usually compared by the number of bits per point (BPP) used to store each point. The lower the BPP, the higher the compression efficiency. In addition, encoding and decoding time is also used as one of the evaluation indicators of compression efficiency. The shorter the time, the higher the efficiency.

In the field of machine vision, the evaluation of fidelity usually uses mean average precision (mAP) as the performance index. For the human visual system, the evaluation of fidelity involves multiple indicators, including parameters such as root mean square error (RMSE) and peak signal-to-noise ratio (PSNR). MPEG defines two additional PSNR metrics for point clouds \cite{schwarz2017766}, namely PSNR D1 and PSNR D2, which correspond to the peak signal-to-noise ratio based on point-to-point distance and point-to-point distance. These metrics collectively reflect the difference between compressed or processed visual content and the original content, allowing for a quantitative analysis of visual fidelity.

\begin{itemize}
   \item Mean Average Precision (mAP)
  
   mAP evaluates the overall performance of an algorithm in multi-category detection by averaging the area under the precision-recall curve (AP) of each category.
   \begin{equation}\label{mAP}
     mAP = \frac{1}{n} \sum_{i=1}^{n} AP_i
   \end{equation}
   Among them, $n$ represents the number of categories, and $AP_i$ is the average accuracy of the i-th category.
   \item Symmetric root mean square (RMSE)
  
   The symmetric root mean square distance uses the original point cloud as a reference to calculate the maximum average error value of the decoded point cloud based on geometric distance.
   \begin{equation}\label{MSE(A,B)}
     MSE(A,B)=\frac{1}{K}\sum \left \| v_{A(i)}-v_{B(k)} \right \|^2
   \end{equation}
   \begin{equation}\label{RMSE(A,B)}
     RMSE(A,B)=\sqrt[]{\frac{MSE(A,B)+MSE(B,A)}{2}}
   \end{equation}
   Among them, $A$ and $B$ represent the original and decoded point clouds respectively; $v_{A(i)}$ and $v_{B(k)}$ represent the original point cloud and the corresponding nearest neighbor decoded point cloud respectively; $K$ is the number of points in the original point cloud.
   \item PSNR-based Geometry Quality Metrics
\end{itemize}
   Based on the characteristics of point cloud data, MPEG proposes a PSNR-based geometric quality measure suitable for 3D point cloud representation. PSNR is obtained from the normalized mean square error MSE, by calculating PSNR in both original and decoded directions to obtain a single symmetric PSNR value for the maximum pooling function.
\begin{itemize}
   \item[-] Point-to-point distance (PSNR D1)
   \begin{equation}\label{D1}
     d_{A,B}^{D1}=\left \| \overrightarrow{e}(i,j) \right \| _2^2
   \end{equation}
   \begin{equation}
     MSE_{D1}=\frac{1}{N_A}\sum_{\forall a_i\in A}d_{A,B}^{D1}
   \end{equation}
   \item[-] Point-to-surface distance (PSNR D2)
   \begin{equation}\label{D2}
     d_{A,B}^{D2}=\left \| \widehat{e}(i,j) \right \| _2^2=(\overrightarrow{e}(i,j)\cdot\overrightarrow{n_j } )^2
   \end{equation}
   \begin{equation}
     MSE_{D2}=\frac{1}{N_A}\sum_{\forall a_i\in A}d_{A,B}^{D2}
   \end{equation}
\end{itemize}

Based on the mean square error (MSE) obtained in the above steps, the corresponding geometric peak signal-to-noise ratio (PSNR) value can be further calculated.
   \begin{equation}\label{PSNR}
     PSNR_{A,B}=10log_{10}(\frac{p_s^2}{MSE_{D1/D2}})
   \end{equation}
   \begin{equation}
     PSNR=max(PSNR_{A,B},PSNR_{B,A})
   \end{equation}
   Among them, $p_s$ represents the signal peak value, and $MSE_{D1/D2}$ is the point-to-point or point-to-point mean square error between all points in the original point cloud $A$ and their corresponding nearest neighbor points in the decoded point cloud $B$. Due to the fact that not all datasets provide the ground truth normals of object surfaces, which are necessary reference values for PSNR D2 calculations, the analysis of PSNR D1 geometric quality based on point-to-point distance is currently more widely used.

\subsection{\textbf{Semantic-based Point Cloud Registration}}
3D point cloud registration is a pivotal process in the domains of computer vision and robotics. This process involves the precise alignment or merging of two or more 3D point cloud datasets into a unified coordinate system. These point clouds may be acquired at different times, from various perspectives, or via different sensors. The objective of registration is to determine the optimal spatial transformation that ensures these point clouds are geometrically consistent or aligned to the desired extent.

\subsubsection*{\bf Current Research Status of Semantic-based Point Cloud Registration}

In the field of point cloud registration, registration algorithms can be divided into two categories: partial registration algorithms and global registration algorithms based on the degree of dependence on the initial transformation parameters. The partial registration algorithm \cite{Besl121791} relies on the heuristic nearest neighbor search strategy to establish the correspondence between point clouds, and can achieve more accurate pose estimation when accurate initial transformation parameters are provided. The global registration \cite{Lei7918612} algorithm refers to a registration method that does not require accurate initial guesses or has low initial guess requirements. It can perform an overall registration operation on two or more point clouds from a macro perspective.
\begin{figure}[!t]
\centering
\includegraphics[width=3.5in]{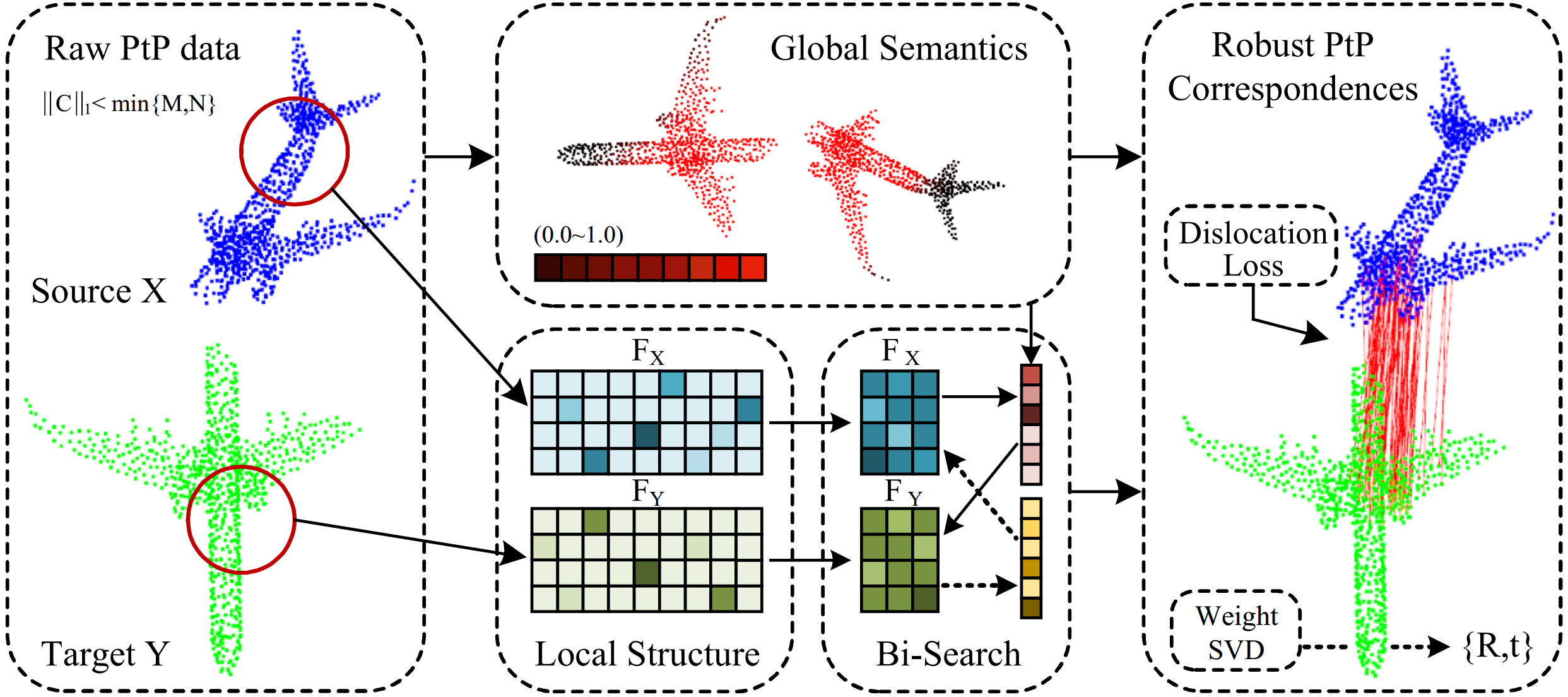}
\caption{Partial-to-Partial (PtP) registration based on semantic-strucual cognition \cite{Shu9860002}.}
\label{Fig:Partial-to-Partial}
\end{figure}
\begin{enumerate}
    \item Partial Registration

    Shu et al. \cite{Shu9860002} implemented a hierarchical partial-to-partial registration strategy, illustrated in Fig.\ref{Fig:Partial-to-Partial}. This approach involves two key phases: the first phase operates at the global semantic level, aiming to generate high-confidence matching candidates; the second phase focuses on the local structure level, seeking to establish robust point-to-point correspondences. 

    Bowman et al. \cite{Bowman7989203} utilized a Gaussian Mixture Model (GMM) to create a correlation model between semantic three-dimensional landmarks and image detection results. Building on this, Hu et al. \cite{Hu9561140} introduced a novel method to enhance point set registration performance by incorporating an additional semantic region association layer, also modeled using GMM. This method employs a hierarchical point set registration strategy, accounting for both external semantic region associations and internal point pair associations. By leveraging GMM to associate semantic regions and integrating the cascaded expectation-maximization (EM) algorithm to address the hierarchical association registration problem, this approach achieves more accurate and robust point cloud registration.
    \item Global Registration
    
    Yin et al. \cite{Yin10160798} developed a semantic global point cloud registration framework by extracting semantic clues using neural networks. This framework provides accurate pose estimation for low-overlap LiDAR scanning in autonomous driving environments and significantly enhances the efficiency and accuracy of the correspondence search algorithm by incorporating semantic information. Similarly, Zhang et al. \cite{Zhang9561929} proposed an innovative semantic point cloud registration framework, which utilizes a hierarchical distributed representation method of features to integrate and extend geometric, color, and semantic information into a non-parametric continuous model. This comprehensive approach achieves more accurate and robust environment mapping and pose estimation.

    In another development, Qiao et al. \cite{Qiao10341394} introduced a low-overlap point cloud registration method known as Pyramid-based Angular-semantic Registration with Outlier Rejection (Pagor). This method employs a distrust and verification strategy, initially using semantic clues for preliminary registration and subsequently utilizing geometric clues to verify and optimize the registration results.

    Furthermore, the Iterative Closest Point (ICP) algorithm \cite{besl586606} is an iterative method for point cloud registration. This algorithm iteratively matches each point in the source point cloud with the closest point in the target point cloud. During each iteration, based on the current point correspondences, the spatial transformation between the two point clouds is calculated, and the source point cloud is updated according to this transformation to reduce the difference between the two point clouds. This process is repeated until predetermined convergence conditions are met. Due to its simplicity and efficiency, the ICP algorithm (global registration) has become a benchmark method in the field of point cloud registration.

    Inspired by the iterative framework of the ICP algorithm, researchers such as Wang et al. \cite{Wang9282862}, Zaganidis et al. \cite{Zaganidis8387438}, Li et al. \cite{Li202109306}, and Truong et al. \cite{Truong8945870} have utilized semantic information to guide the establishment of point correspondences. This approach enhances the accuracy of correspondences between points with the same semantics, thereby improving the overall robustness and precision of point cloud registration.

\end{enumerate}
Current semantic-based point cloud registration work aims to further improve the robustness and convergence of point set registration methods by fusing semantic information extracted from images. In addition, combined with pose estimation technology, multi-scan registration can not only achieve spatial alignment, but also segment and integrate scan data to build a semantic map.
\subsubsection*{\bf Point Cloud Registration Evaluation Index}
\label{Point cloud registration evaluation index}

In the quantitative assessment of point cloud registration, similarity measures are typically employed to evaluate the degree of correspondence between two sets of point clouds. The pertinent quantitative evaluation metrics can be categorized into distance-based and transformation error-based measurement approaches. Distance-based measurement methods are widely regarded as fundamental and commonly utilized evaluation techniques. These methods involve a quantitative analysis of the spatial distance between the transformed source and target point clouds.
\begin{itemize}
   \item Chamfer Distance (CD)
  
   The Chamfer Distance \cite{Fan8099747} is extensively utilized to assess the discrepancy between two point clouds. This metric's calculation typically entails the registration of corresponding points within the two point cloud sets, followed by the computation of the spatial distance between these registered point pairs. Considering the discrete nature of 3D point cloud data, various approximation strategies are frequently employed in practical applications to estimate Chamfer Distances. These strategies facilitate effective surface distance evaluation on discrete point cloud datasets.
   \begin{equation}
     d_{CD}(A,B)=\frac{1}{A} \sum_{x\in A}\min_{y\in B} \left \| x-y \right \|_2^2 +\frac{ 1}{B} \sum_{y\in B}\min_{x\in A} \left \| y-x \right \|_2^2
   \end{equation}
   Among them, $A$ and $B$ denote two sets of 3D point clouds respectively. The first term represents the sum of the minimum distances from any point $x$ in $A$ to the midpoint of $B$, and the second term represents any point in $B$. The sum of the minimum distances from a point $y$ to $A$.
   Lower value of chamfer distance tends to indicate higher morphological consistency between the two sets of point clouds, and it can be inferred that the registration process has higher accuracy.
\begin{figure}[!t]
\centering
\includegraphics[width=3.4in]{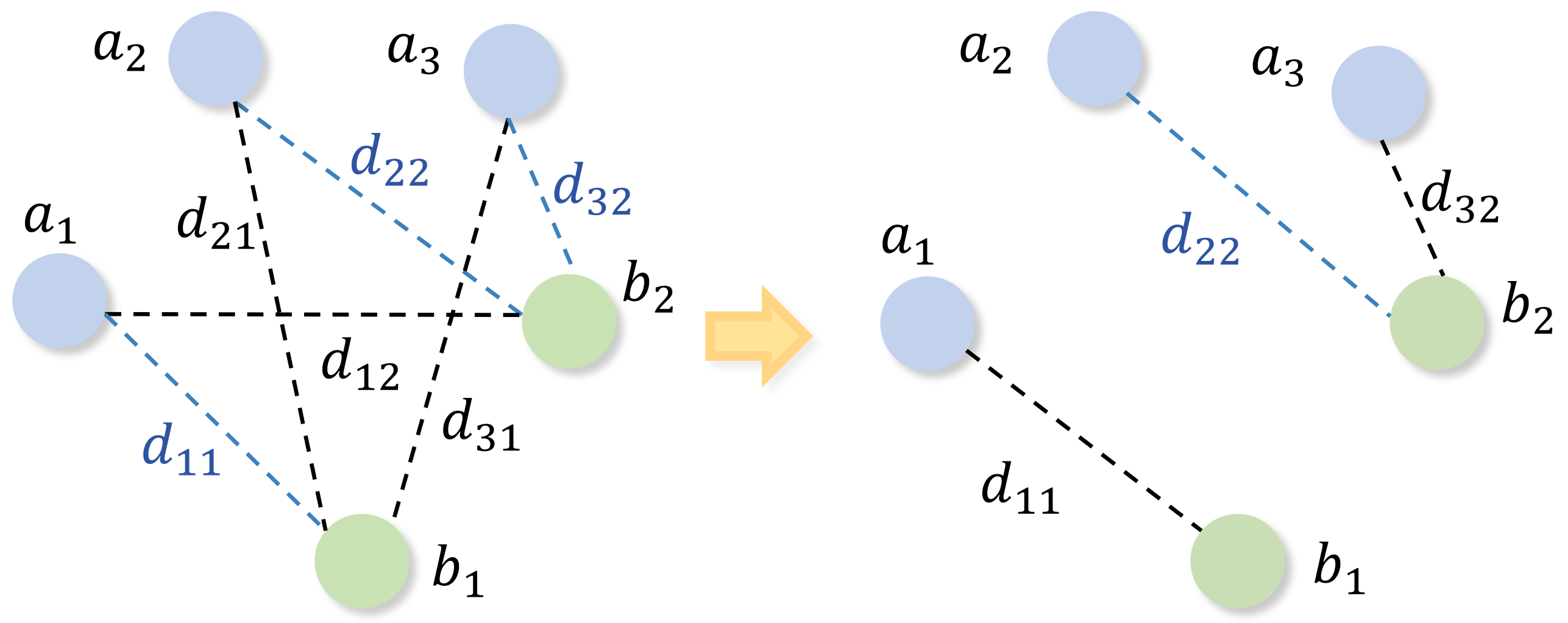}
\caption{Hausdorff distance illustration. Calculate the distance between points of different classes, take the minimum distance value between points of different classes, which is $d_{11}$, $d_{22}$, $d_{32}$, and then take the maximum one as the HD value, which is $d_{22}$.}
\label{Fig:Hausdorff}
\end{figure}
   \item Hausdorff Distance (HD)
  
   Hausdorff distance \cite{Hutt232073} is used to describe the degree of similarity between two sets of points, as shown in Fig.\ref{Fig:Hausdorff}.
   \begin{equation}
     H(A,B)=max(h(A,B),h(B,A))
   \end{equation}
   \begin{equation}
     h(A,B)=\max _{a \in A}\left \{ \min_{b \in B}\left \| a-b \right \| \right \}
   \end{equation}
   \begin{equation}
     h(B,A)=\max _{b \in B}\left \{ \min_{a \in A}\left \| b-a \right \| \right \}
   \end{equation}
   Among them, $A$ and $B$ represent two sets of 3D point clouds respectively, and $\left \| \cdot \right \| $ is the distance paradigm between point set $A$ and point set $B$. (eg: $L2$ or Euclidean distance) 
\end{itemize}
Commonly used quantitative evaluation indicators based on transformation errors include relative translation error, relative rotation error \cite{Pomerleau1393272}, etc. The specific details are as follows:
\begin{itemize}
   \item Relative Translation Error (RTE)
  
   The relative translation error is defined as the difference measure between the observed actual translation vector between two point clouds and the translation vector estimated by the registration algorithm during the point cloud registration process. This error index is used to quantify the accuracy of the registration algorithm in spatial transformation alignment, reflecting the accuracy of the algorithm's estimation of the relative position between point clouds.
   \begin{equation}
     RRE=\sqrt{\Delta x^2+\Delta y^2+\Delta z^2}
   \end{equation}
   Among them, $\Delta x$, $\Delta y$, and $\Delta z$ denote the deviations between the actual translation vector and the predicted translation vector, represented mathematically as $t_{gt} - t_{pre}$.
   \item Relative Rotation Error (RRE)
  
   Relative rotation error is a key performance metric used to evaluate rotational alignment accuracy. This metric quantifies the difference in rotation between the rotation matrix (or quaternion) estimated by the registration algorithm and the actual rotation of the two point clouds. The goal of point cloud registration not only includes spatial translation alignment, but also requires the accuracy of rotation alignment. The calculation of RRE involves comparing the rotation representation output by the registration algorithm with the representation of the true rotation to evaluate the accuracy of the algorithm's estimation of the rotation parameters.
   \begin{equation}
     RTE=arccos(\frac{trace((R_{pre}R_{gt}^{-1})-1)}{2} )
   \end{equation}
   Where $trace()$ denotes the trace of a matrix, which is the sum of its diagonal elements. In this context, $R_{pre}$ represents the rotation matrix estimated by the registration algorithm, while $R_{gt}$ signifies the rotation matrix corresponding to the ground truth. This formula computes the angular difference between the two rotation matrices, with the resulting value expressed in radians.
\end{itemize}
In related academic research, registration algorithms are typically optimized based on the correspondences between point pairs. Consequently, in addition to measuring registration error, the recall metric is often employed to represent the percentage of correctly matched point pairs. Hence, the recall metric is crucial for assessing the robustness of research methods in handling noise and outliers.

\begin{itemize}
   \item Recall Rate (Registration Recall, RR)
  
   When registering each pair of scan data, the registration error introduced by the predicted transformation parameters can be calculated by the following formula \cite{Choy9009829}:
   \begin{equation}
     E_{RMSE}=\sqrt{\frac{1}{\Omega _{gt}}\sum _{(a,b) \in \Omega_{gt}}\left \| b-R_{pre}a- t_{pre}\right \|_2 }
   \end{equation}
   Among them, $\Omega _{gt}$ represents the corresponding true value segment, and $(R_{pre},t_{pre})$ is the transformation parameter rotation matrix and translation vector of attitude estimation.
\end{itemize}

\begin{figure}[!t]
\centering
\includegraphics[width=3.0in]{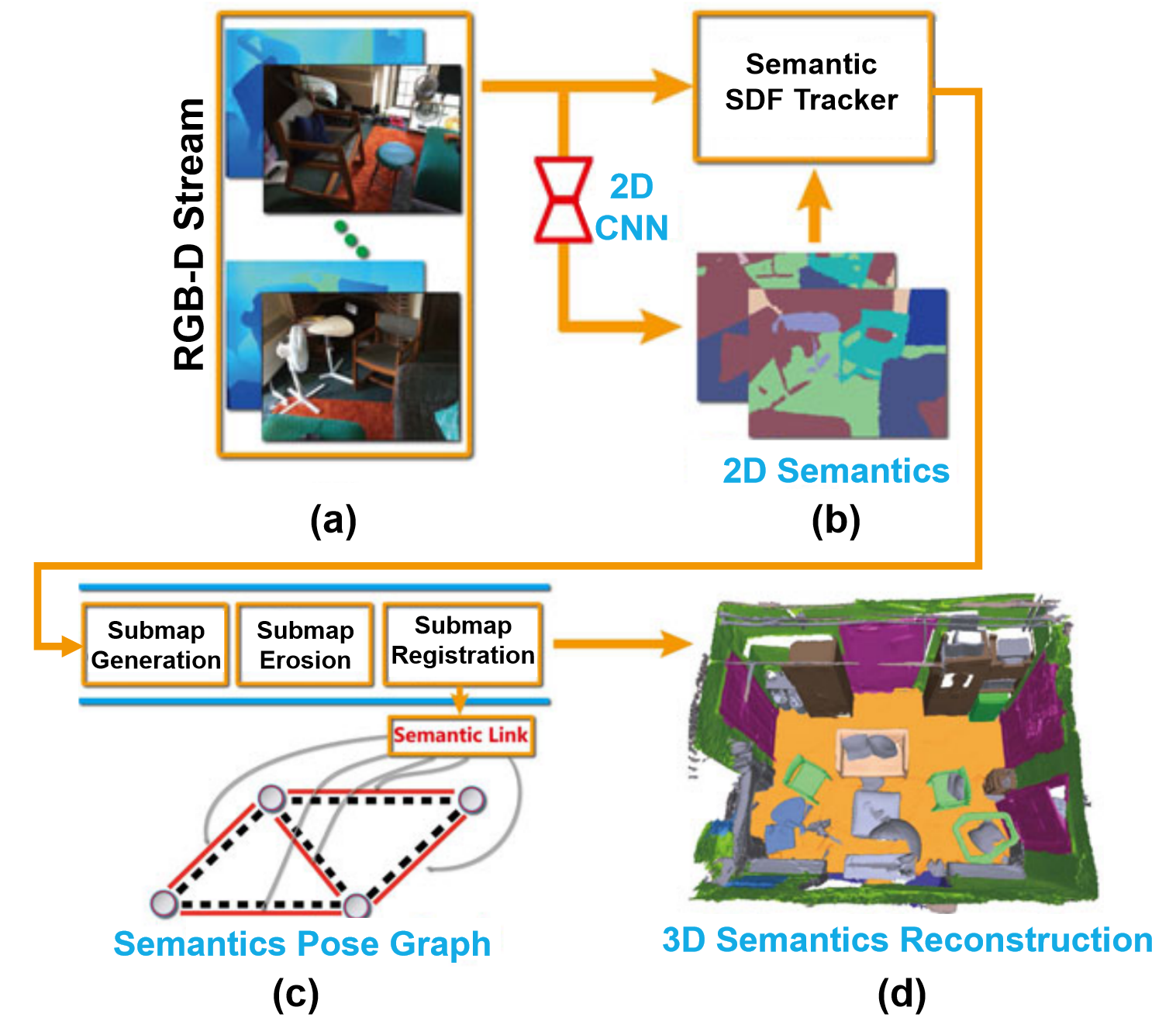}
\caption{Point cloud reconstruction With semantic priors \cite{Huang9662197}. (a) RGB-D data stream, (b) semantic tracker, (c) semantic pose graph, (d) 3D semantic reconstruction.}
\end{figure}

\subsection{\textbf{Semantic-based Point Cloud Reconstruction}}
Point cloud reconstruction entails the process of recovering the geometry and structure of an object from a collection of points distributed in three-dimensional space. These points are generally obtained through techniques such as laser scanning, structured light scanning, or voxel rendering. The primary objective of point cloud reconstruction is to construct an accurate geometric model of the object from these irregularly distributed points.
\subsubsection*{\bf Research Status of Point Cloud Reconstruction based on Semantics}

By analyzing the role of explicit representation and implicit guidance of semantics in other traditional tasks, it can be inferred that the integration of semantic cues can optimize task performance. We observe that similar applications exist in point cloud reconstruction. When leveraging semantic prior information for instruction, the assessment of accuracy is complicated by the difficult-to-measure nature of semantic information.

BuildingFusion \cite{Zheng9286413} is an innovative semantic-aware closed-loop detection method that utilizes semantic information as a positional identifier to enhance robustness against similar positions. Additionally, BuildingFusion can provide real-time semantic and structural information, enabling an immediate understanding of online scenes. Similarly, Huang et al. \cite{Huang9662197} proposed a semantic space model using a continuous metric function to quantify the distance between discrete semantic concepts. In the context of 3D reconstruction, this method employs semantic mapping and registration techniques to establish reliable semantic correspondences and construct a global pose map. By leveraging compact semantic priors, this technique effectively integrates semantic and geometric cues, facilitating real-time 3D reconstruction.

Currently, the exploration of semantic-based reconstruction in point cloud representation is relatively underdeveloped. Compared to other three-dimensional representations, this area requires further investigation to fully realize its potential.
\subsubsection*{\bf Research Status of Other 3D Representation Reconstruction based on Semantics}

Neural Radiance Fields (NeRF) is a technology that demonstrates exceptional performance in the field of 3D scene reconstruction and rendering. It leverages neural networks to learn the lighting and texture information of a scene, enabling the high-quality rendering of 3D images from any viewing angle. As a crucial technology in 3D reconstruction, research on NeRF offers valuable insights for point cloud reconstruction. Furthermore, point clouds can serve as an input data type for the NeRF model, highlighting the significant practical and application value of integrating and developing these technologies.

Before NeRF can generate a three-dimensional scene, it requires the collection of images from multiple perspectives. To minimize the number of images needed as input for NeRF, researchers have explored the strategy of introducing semantic guidance information during the reconstruction process. For instance, SG-NeRF \cite{Qu10219715} employs sparse point clouds as geometric constraints for NeRF optimization and performs semantic predictions on both two-dimensional images and point clouds. This approach guides the search for adjacent neural points during the reconstruction process. By using semantic information, the sampled query points can accurately identify neural points that are structurally adjacent to the query points within a non-uniformly distributed sparse point cloud.

Additionally, in a three-dimensional scene, decoupling instances and attributes within those instances is critical for accurate scene reconstruction. NaviNeRF \cite{Xie1799218002} uses self-supervised learning to identify interpretable semantic directions in reconstructed scenes. This method combines 3D reconstruction technology with potential semantic operations, enabling fine-grained solutions to 3D scene reconstruction challenges.

\subsubsection*{\bf Point Cloud Reconstruction Evaluation Index}

In point cloud reconstruction, commonly used evaluation indicators include chamfer distance (see \ref{Point cloud registration evaluation index}), average intersection over union (see \ref{Point cloud semantic segmentation evaluation index}), RMSE (See \ref{Point cloud compression evaluation index}) etc.

\section{New tasks introduced by semantic information}
The increasing demand for deeper understanding and reasoning in visual tasks represented by 3D point clouds has established point cloud understanding as a crucial research domain. This field includes key components such as scene graph prediction and the integration of 3D vision with natural language. Scene graph prediction focuses on extracting semantic relationships between objects within a point cloud, thereby enabling machines to comprehend the spatial and contextual aspects of a scene more accurately. Meanwhile, the integration of 3D vision and language aims to bridge the gap between visual data and natural language processing, allowing machines to describe and interpret 3D scenes with greater precision.

A significant element within 3D vision and language is 3D dense captioning, which generates detailed descriptions for individual points in a point cloud, providing a comprehensive understanding of the scene's content. Additionally, point cloud-based semantic localization is another critical aspect of this area. It involves the precise identification and localization of specific objects or regions of interest within a point cloud based on their semantic attributes. By integrating these techniques, the field of point cloud understanding contributes to the development of more advanced and intelligent 3D perception systems, ultimately enhancing our capacity to interact with and interpret the 3D world.

\subsection{\textbf{3D Scene Understanding}}
With the advancement of artificial intelligence technology, the understanding of scene content has evolved beyond traditional localized tasks such as object recognition and segmentation. It now encompasses a deeper exploration of objects, their attributes, and the global spatial relationships between objects within a scene \cite{Chang9661322}. In the field of 3D point cloud research, the current mainstream methods for scene understanding primarily focus on scene graph prediction and 3D vision integrated with language. This study provides a detailed explanation and comparative analysis of these two approaches, highlighting their characteristics, advantages, and applicable scenarios when processing three-dimensional spatial data.

\subsubsection*{\bf Scene Graph Prediction}
\label{Scene Graph}
\begin{figure}[!t]
    \centering
    \includegraphics[width=3.5in]{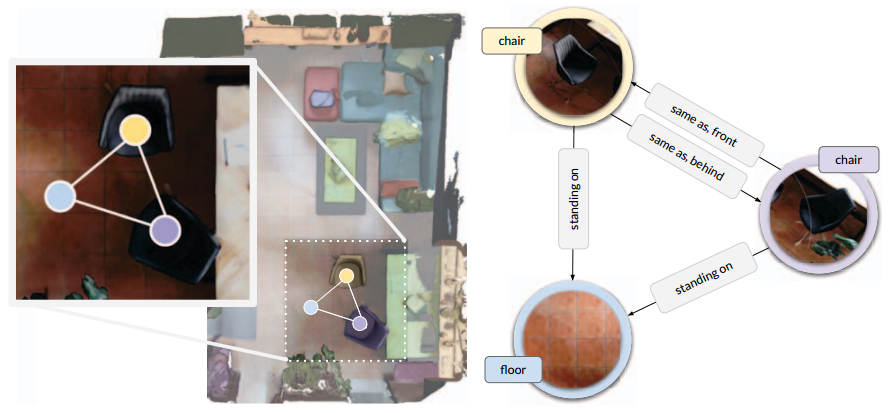}
    \caption{Scene graph prediction: Semantic scene graph inference from point clouds \cite{Wald9156565}.}
    \label{Fig:SSG}
\end{figure}

As a structured representation method, the three-dimensional scene graph is dedicated to describing objects within a scene and their interrelationships. In the scene graph prediction task, the primary objective is to identify object instances within the image (such as humans, vehicles, furniture, etc.) and to predict the pairwise relationships between these objects (e.g., "located on the right of...", "nearby", etc.). Ultimately, a series of (subject, predicate, object) triples are generated. As illustrated in \ref{Fig:SSG}, this representation method emphasizes the semantic analysis of image content and the modeling of relationships between objects.

Wald et al. \cite{Wald9156565} proposed the first learning-based approach for generating semantic scene graphs from 3D point clouds and demonstrated how 3D semantic scene graphs can be utilized for 2D cross-domain retrieval. Furthermore, the team developed a point cloud semantic scene graph dataset, 3DSSG, employing the GNN-based baseline model SGPN. This dataset includes detailed semantics within nodes (instances), encompassing attributes and edges (relationships).

Wu et al. \cite{Wu9578559} developed a method for predicting real-time 3D scene graphs in a 3D environment by incrementally integrating newly observed subgraph predictions into a globally consistent semantic scene graph model.

Zhang et al. \cite{Zhang9578123} introduced a graph autoencoder network capable of automatically learning a set of embedding vectors for each category. These pre-learned embeddings are then utilized for the prediction of three-dimensional scene graphs (3DSSG), enabling the network to identify reliable relational triples from pre-trained knowledge.

To address visual confusion issues, SGGpoint \cite{Zhang209555403} employs prior knowledge for scene graph prediction, acquiring information solely through the structural patterns of semantic categories and rules. After training, the scene graph prediction model integrates geometric features with corresponding knowledge element embeddings to form relational triples.

Building on previous work \cite{Wald9156565}, Wald et al. introduced an embedding-based method \cite{wald20630651} for learning semantic scene graphs from raw 3D point clouds. They further released the RIO10 localization benchmark's scene graph to supplement the large-scale 3D scene graph dataset 3DSSG.

Following the development of these fully supervised tasks, Koch et al. \cite{koch34043414} proposed a self-supervised pre-training method, SGRec3D. SGRec3D uses three-dimensional graph structure reconstruction as an auxiliary pre-training task and enhances the efficiency of the pre-training process by learning the optimal information flow within the graph.

Given the uneven distribution of word segments in the training set triples, which leads to a significant long-tail distribution, VL-SAT \cite{Wang10205194} seeks to utilize visual-linguistic semantic information to enhance the understanding of three-dimensional structures. This approach improves the 3D scene graph prediction model's ability to distinguish long-tail distributions and ambiguous semantic relationship triples. Additionally, VL-SAT demonstrated the generalization capability and feasibility of this strategy in scene graph prediction tasks.

\begin{figure}[!t]
\centering
\includegraphics[width=3.4in]{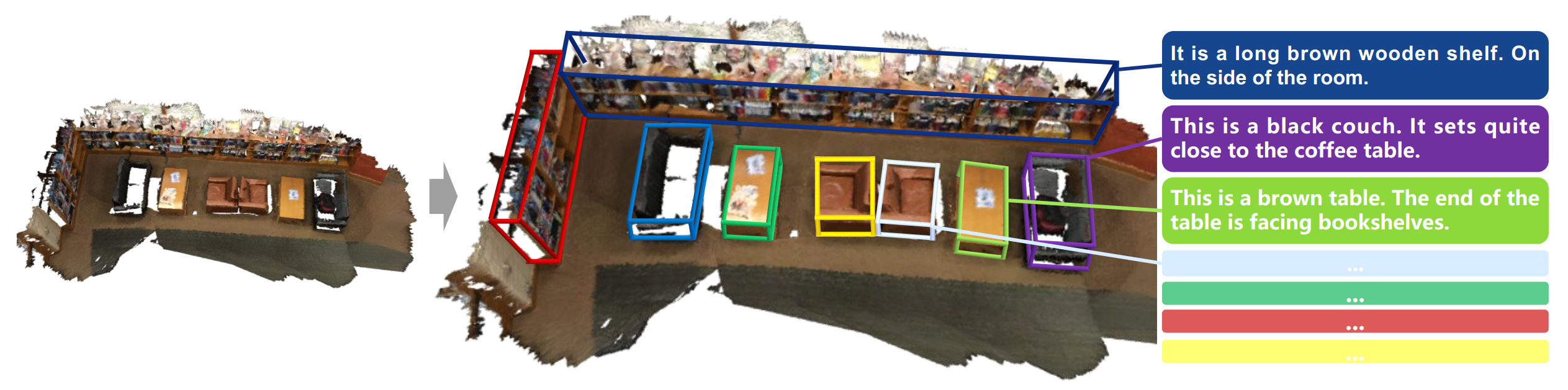}
\caption{3D dense captioning task: localization and description of objects in 3D scenes\cite{Yu10187165}.}
\label{Fig:3DdenseCaption}
\end{figure}

Scene graphs offer a compact representation of scene context, making them suitable for various industrial applications. The success of mapping 3D scenes to scene graphs also opens up the possibility of reversing the process, i.e., generating 3D scenes from scene graphs. Dhamo et al. \cite{Dhamo9710451} were the first to implement the direct generation of 3D representations from scene graphs in an end-to-end manner, using a novel model architecture that shares the layout and shape generation process. Additionally, CommonScenes \cite{zhai202436512}, as a fully generative model, combines variational autoencoders and latent diffusion models to achieve parallel modeling of scene layout and shape distribution.

Currently, research on point cloud scene graph prediction has advanced significantly. Researchers have explored various approaches, such as utilizing 3D scene graphs for cross-domain retrieval between 2D and 3D data, enhancing localization, and incorporating new learning methods to improve prediction accuracy.

\begin{enumerate}
    \item Scene Graph Prediction Evaluation Index

    In 3D point scene graph prediction, the experimental settings and evaluation indicators of 3DSSG \cite{Wald9156565} are mainly followed. For the accuracy evaluation of triplet prediction, top-k precision ($A@k$) and top-k recall ($R@k$) are mainly used as indicators. Additionally, since scene graph prediction can be considered an extension of the semantic segmentation task, some studies have adopted the top-k evaluation approach and introduced metrics such as $F1@k$ to comprehensively assess model.

    \begin{itemize}
   \item Top-k Triple Accuracy (A@k)
  
   For triples (subject, predicate, object), the three partial scores are first multiplied to obtain the triple score, and then the top-k accuracy ($A@k$) is calculated as the evaluation index. This triplet is considered correct only if the subject, predicate, and object are all correct. To solve the problem of long-tail distribution in the database, the average top-k accuracy ($mA@k$) can be calculated, such as the average top-k accuracy of predicates in all predicate classes. The top-k triple accuracy calculation formula is as follows:
   \begin{equation}
     A@k = \frac{trip_{c}}{k}
   \end{equation}
   \begin{equation}
     mA@k = \frac{1}{k}\sum_{i=1}^{k} A@k
   \end{equation}
   Where $trip_{c}$ represents the correct number of triples, and $k$ represents the number of top-k triples.
   \item Top-k Triple Recall Rate (R@k)
  
    Recall is commonly used to evaluate a model's coverage of positive samples; thus, mean recall ($mR@k$) can assess performance in cases of uneven sampling relationships, following a strategy similar to mean average precision ($mA@k$). As with precision, the three partial scores are multiplied to obtain the triplet score, and then top-k recall is calculated. The formula for calculating the top-k triplet recall rate is as follows:
   \begin{equation}
     R@k = \frac{trip_{c}}{GT}
   \end{equation}
   \begin{equation}
     mR@k = \frac{1}{k}\sum_{i=1}^{k} R@k
   \end{equation}
   Where $trip_{c}$ represents the number of correct triples, n represents the number of top-k triples, and $GT$ represents the number of true values.
\end{itemize}
\end{enumerate}

\subsubsection*{\bf 3D Vision with Language}

Currently, tasks related to 3D vision and language, such as 3D dense captioning and 3D point cloud localization, are being explored. 
\begin{figure*}[!t]
\centering
\includegraphics[width=7.0in]{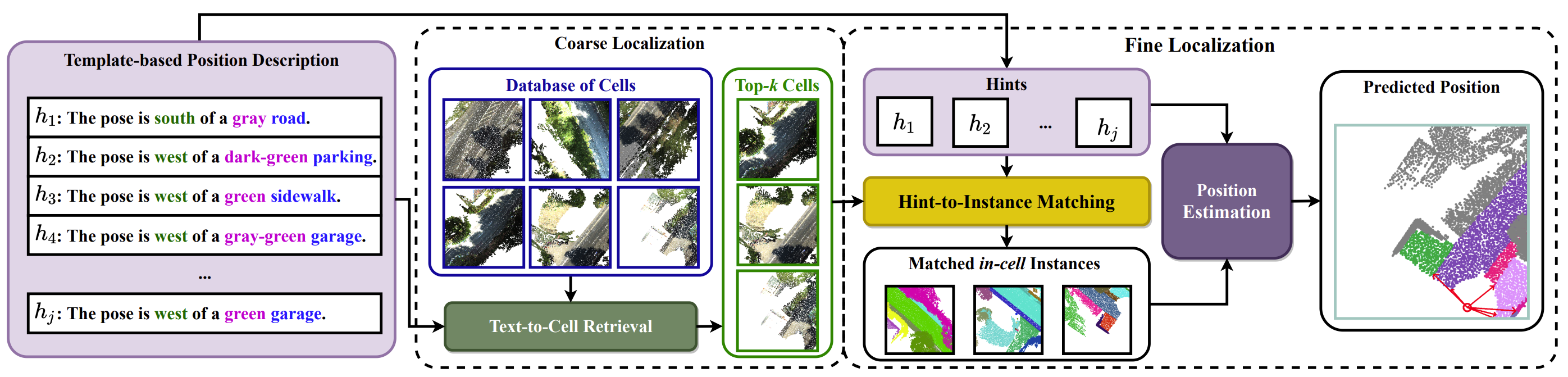}
\caption{Text-to-point-cloud cross-modal localization \cite{Kolm9880174}.}
\label{Fig:3Dlocalization}
\end{figure*}   
\begin{enumerate}
\item 3D Dense Captioning

    3D dense captioning involve describing 3D scenes, such as point clouds, by not only identifying objects within the scene but also generating detailed descriptions of these objects. These descriptions are typically presented in natural language, aiming to provide a richer understanding of the environment for applications like robotics and self-driving cars. The task requires locating and classifying objects in three-dimensional space and generating natural language descriptions, emphasizing the comprehension of three-dimensional spatial information and language generation capabilities.

    In contrast, 2D dense captioning focus on identifying and expressing local visual details of 2D content in multiple natural languages. Chen et al. \cite{Chen9577651} expanded this traditional task by extending the concept of dense captioning from two-dimensional to three-dimensional space, proposing the 3D dense captioning task. Unlike the scene graph prediction task, the 3D dense captioning task prioritizes the target object in the scene and primarily learns the main relationships between the target object and its surrounding objects \cite{Cai9879358}. The input data for this task is point cloud data generated from a three-dimensional scene, and the output includes text descriptions of the target object and the corresponding specific bounding box, as shown in Fig.\ref{Fig:3DdenseCaption}.

    In the 3D dense captioning task, to effectively model complex connections and cross-modal interactions in indoor scenes, researchers have adopted various relationship modeling methods. These methods include, but are not limited to, graph-based methods, Transformer architecture-based methods, and knowledge distillation techniques. Traditional image captioning methods typically rely on the visual features of objects detected from images and generate corresponding natural language descriptions through natural language inference models, such as RNNs or LSTMs. However, these methods often fail to fully exploit the semantic relationships between objects, resulting in limitations in the accuracy of the generated language descriptions.

    Graph-based methods focus on capturing the structural characteristics between objects and their mutual relationships, using nodes to represent objects and edges to represent relative positional relationships. Methods such as Scan2Cap \cite{Chen9577651} and MORE \cite{jiao528545} use semantic scene graphs \cite{Feng9710755} to capture the relative spatial positional relationships between objects (e.g., "top," "front," "left," or "center"). Graph-based methods can facilitate the inference of relationships, allowing for reasoning even when two objects are not directly connected by edges in the graph.

    Transformer architecture-based methods employ self-attention mechanisms to manage long-distance dependencies in sequence data. Specifically, both SpaCap3D \cite{wang220410688} and 3DJCG \cite{Cai9879358} use transformer-based modules to construct relationship models between objects. SpaCap3D utilizes spatially guided transformers to capture relationships between objects. These methods rely on the transformer's self-attention mechanism to establish relationship models between objects and capture long-distance dependencies in the scene. The 3DJCG method enhances attribute features through a feature enhancement module composed of a multi-head self-attention layer and a fully connected layer.

    Additionally, knowledge distillation, as a model compression and knowledge transfer technique, helps reduce model complexity while maintaining performance. For example, X-Trans2Cap \cite{Yuan9879338} applied a knowledge distillation framework with a cross-modal fusion module to facilitate the interaction between 3D object features and multiple modalities.

    Together, these methods provide powerful tools for the 3D dense captioning task, enabling accurate description and explanation of objects and their relationships in indoor scenes.

    \begin{itemize}
        \item 3D Dense captioning Evaluation Index

        The 3D dense captioning evaluation index mainly conducts a comprehensive evaluation from two dimensions: detection and captioning. In order to jointly evaluate the positioning and captioning generation capabilities of the model, the $m_c@kIoU$ metric is generally used, where $m_c$ is a metric specially designed for image captioning, including CIDEr (C) \cite{Vedantam7299087}, BLEU-4 (B -4) \cite{papineni311318}, ROUGE (R) \cite{lin20047481}, METEOR (m) \cite{banerjee6572}, etc. k represents the intersection ratio threshold of the predicted bounding box and the true bounding box, that is, only the parts whose intersection ratio is greater than k are considered. Calculated as follows:
        \begin{equation}
            m_c@kIoU=\frac{1}{N}\sum _{i=1}^{N}m_{ci}
        \end{equation}
        Where N represents the number of all captioned instances in the evaluation dataset, and $m_c$ can be any metric among CIDEr, BLEU-4, ROUGE, and METEOR.
    \end{itemize}


 
    
\item Point Cloud Localization

    Point cloud localization, which focuses on the geometric shape of the scene, offers several advantages over image-based approaches. Unlike images, point clouds are not affected by variations in lighting, weather, or seasons, ensuring consistent representation of geometric structures. With the maturation of image localization research, point cloud localization, particularly deep learning-based 3D localization, has emerged as a prominent area of interest \cite{Uy8578568},\cite{Xia149514967}.  Currently, only a few networks have been proposed for language-based localization in 3D large-scale urban maps.

    Text2Pose \cite{Kolm9880174} is a pioneering approach that aligns objects described in text with their corresponding instances in a point cloud using a coarse-to-fine strategy. However, this method overlooks the relationship between instances and sentences. To address this issue, Wang et al. \cite{Wang250109023} developed the RET network, based on the Transformer architecture, to enhance the discriminative representation of point clouds and natural language queries.

    Inspired by RET, Xia et al. \cite{Xia149514967} tackled the problem of 3D point cloud localization using natural language descriptions by introducing a new neural network, Text2Loc. This network effectively captures the semantic relationships between points and text, achieving state-of-the-art performance. 
\end{enumerate}

\begin{itemize}
    \item Point Cloud Localization Evaluation Index
    
    In point cloud localization, commonly uses top-k recall as  evaluation index, which has been introduced detailly in scene graph prediction ealuation index (see \ref{Scene Graph}).
\end{itemize}

Both tasks are developed in conjunction with deep learning and corresponding tasks in two dimensions. Since the development of these two-dimensional tasks is still ongoing, the evaluation metrics are primarily based on a two-dimensional evaluation system. Although the processes of the two tasks are similar, they have not yet been fully unified.

\subsection{\textbf{Point Cloud Semantic Scene Completion}}
\label{Point cloud semantic scene completion}
In the field of visual scene understanding, humans can mentally complete incomplete objects in a scene by observing their visible parts. For instance, they can infer the presence of a doll based on partially exposed elements. This capability relies on a deep semantic understanding of the scene, allowing observers to estimate the spatial layout of objects using prior knowledge. For autonomous agents, such as navigation or object interaction robots, the ability to infer geometric shapes based on semantic understanding is highly advantageous. A robot with this intuitive capacity can predict the geometry behind a surface from a limited view, enabling it to plan actions without fully exploring the environment.

Recent advancements in three-dimensional deep learning have introduced Semantic Scene Completion (SSC) as an extension of Scene Completion (SC). In SSC, both the semantic information and geometric structure of a scene are inferred simultaneously. The increasing richness of semantic data distinguishes SSC from traditional SC, particularly in terms of input data attributes and sparsity.

\begin{figure}[!t]
    \centering
    \includegraphics[width=3.5in]{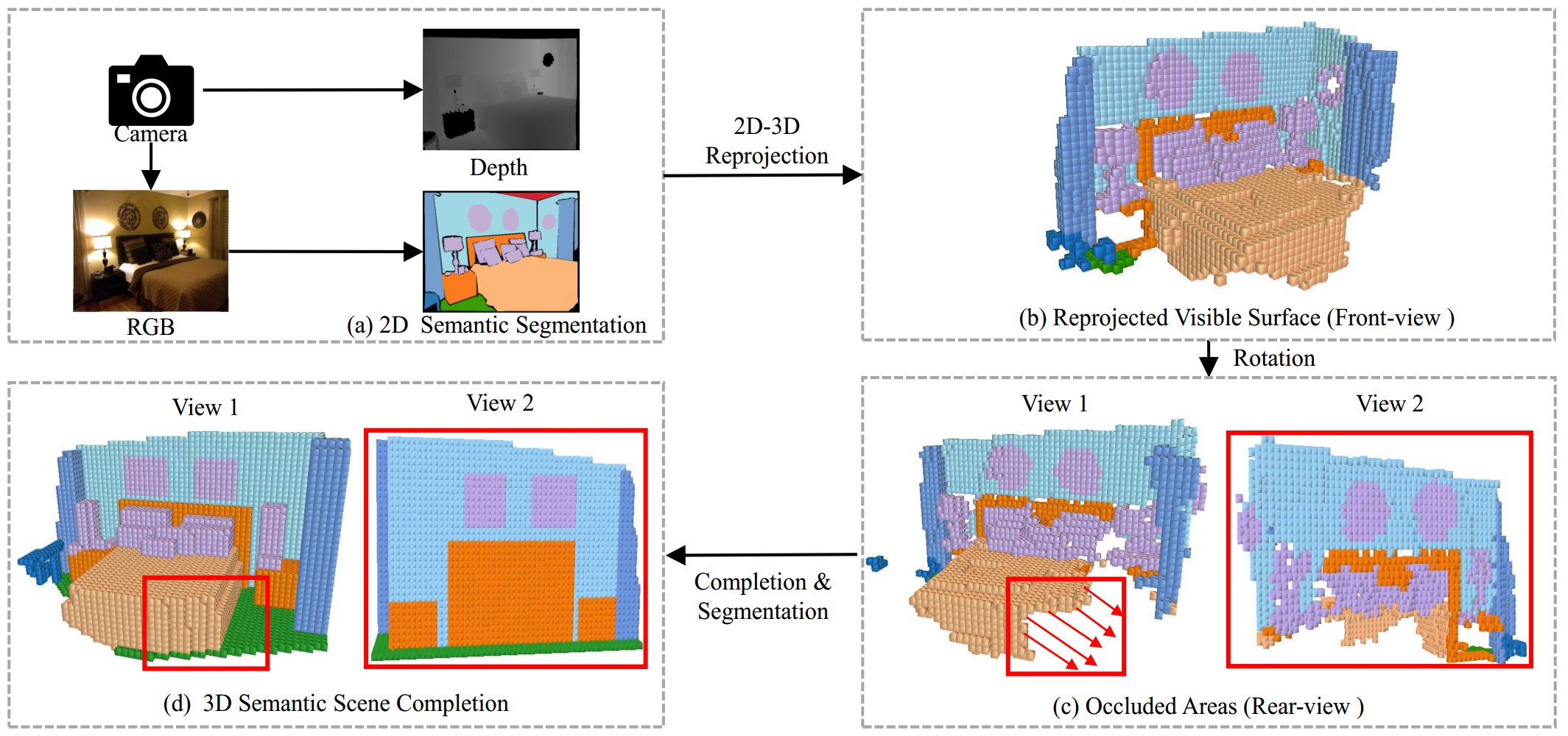}
    \caption{Schematic diagram of semantic scene completion \cite{Li10007036}. (a) 2D semantic segmentation generated for RGB images. (b) and (c) are different views of the point cloud, and (d) is the semantic scene completion result.}
    \label{Fig:SSC}
\end{figure}

Song and colleagues \cite{Song8099511} first introduced an end-to-end model called SSCNet to tackle the semantic scene completion task. This model uses the integration of scene completion and depth map semantic labeling to achieve two objectives simultaneously. They demonstrated that the joint approach outperforms addressing each task separately. The schematic diagram of semantic scene completion is illustrated in Fig.\ref{Fig:SSC}.

Scene Completion (SC) has become a crucial pre-processing step in 3D vision tasks, facilitating downstream processing. Extensive research has been conducted on various representation methods, including point-based and voxel-based approaches. In contrast, Semantic Scene Completion (SSC) is still in its early development stage. Most studies in SSC use depth maps as input and design models and network architectures based on voxel representations, as seen in works by \cite{Li9435044},\cite{Zhang9008381},\cite{Chen9156418},\cite{miao230213540},\cite{li1140211409},\cite{Garb9025517},\cite{Liu20189872},\cite{Wang1031410323}, among others. Unlike voxel-based methods, humans naturally perceive and understand the world through the concept of "instances." Following this approach, Jiang et al. \cite{Jiang202520267} employed instance concepts to guide semantic scene reconstruction.

Previous research in 3D geometry reconstruction often assumes uniform measurements of the scene by sensors, leading to a focus on voxel frameworks and ultimately decoding into voxel labels. However, few studies address directly processing points or generating point-level labels, and point-based semantic scene completion methods remain underdeveloped. Additionally, unlike RGBD cameras, the data density from lidar scanning decreases with distance, causing point cloud data to become uneven. This not only complicates scene completion but also challenges high-resolution downstream tasks, requiring dense 3D structures and associated semantic labels from "sparse" representations.

Based on this, Cheng \cite{Cheng21482161} et al. supplemented the 3D network based on a multi-view fusion strategy to provide robustness to extreme sparsity.
Similarly, An et al. \cite{An10409585} developed a semantic scene completion network ESC-Net for extremely sparse scenes. First, feature map completion technology is utilized to recover accurate spatial features. Secondly, 3D object boundary details are preserved by correcting the occupancy and semantic labels of regions in three-dimensional space and bird's-eye space. Third, a combined network (ESC-Net-D) is used to overcome the dilemma of imbalance of foreground and background GT labels. And can achieve better results in downstream tasks (point cloud registration, target detection).

Deep implicit functions are advantageous for modeling complex relationships, providing continuous representation, and exhibiting strong generalization capabilities, making them a powerful tool in semantic scene completion. Rist et al. \cite{Rist9477025} introduced a novel scene segmentation network based on local depth implicit functions, which constructs a continuous scene representation by locally encoding point clouds at different resolutions. This approach effectively captures the geometric and semantic features of outdoor scenes, avoids the voxelization inherent in traditional methods, does not require spatial discretization, and overcomes the trade-off between detail and coverage.

Similarly, Li et al. \cite{Li2189421904} employed deformable templates based on deep implicit function representations (DDIT) for semantic scene completion, where each instance in the scene is represented segmentally by a template. This method constrains the overall shape while preserving fine details of each instance.

To address the challenge of misclassification in small-sized objects and complex scenes, Xia et al. \cite{Xia10203998} designed kernels of various sizes to aggregate multi-scale features, leveraging rich contextual information. They also introduced a knowledge distillation strategy, optimizing it according to the task's characteristics. By aligning through feature index, this approach ensures that the paired feature relationships learned by the student model match those in the teacher model, facilitating effective transfer of relationship knowledge.

In the development trajectory of the field of semantic scene completion, a series of innovative methods combining new technologies have emerged. These methods can provide important reference for point-level semantic scene completion tasks. Dong et al. \cite{Dong88748883} use multi-view feature synthesis (MVFS) and cross-view transformer (CVTr) based on voxel representation to achieve cross-view modeling of object relationships. To obtain relevant information about the existence of occluded objects. At present, although there are a variety of technical means that can convert only two-dimensional image input into a three-dimensional voxel scene with semantic labels \cite{Li10203337},\cite{Yao94559465},\cite{Xiao10379527}, there has not yet been a method that can directly convert two-dimensional image input into a three-dimensional voxel scene with semantic labels. Lee \cite{Lee2833728347} et al. applied the diffusion model to SSC, achieving outdoor scene generation and improving its applicability in various downstream tasks.

As an emerging research field, there are currently relatively few studies on semantic scene completion for point cloud data, which shows that this field has broad exploration space and potential research value. Judging from the progress of point cloud semantic segmentation, different representations and their combinations (point-based, voxel-based, and unit set-based) show their unique advantages. Therefore, in the semantic scene completion task, fine-grained collaborative optimization with the model can be performed in the scene through semantic relationships to improve overall performance. The performance achieved in the semantic scene completion task reveals that in a multi-task environment, the diversity of feature extraction strategies and the integrated application of cross-modal technologies provide rich inspiration and reference value for different fields.

\subsubsection*{\bf Semantic Scene Completion Evaluation Index}

Semantic scene completion metrics are similar to those used in semantic segmentation, including indicators such as the average intersection over union (mIoU), recall rate (RR), and accuracy (see \ref{Point cloud semantic segmentation evaluation index} and \ref{Point cloud registration evaluation index}).

Precision is a performance metric in machine learning and information retrieval that measures how often a model correctly predicts a positive class. In the context of semantic scene completion, precision typically refers to the proportion of predicted positive pixels that are indeed positive. The formula for calculating precision is as follows:
\begin{equation}
   Precision=\frac{TP}{TP+FP}
\end{equation}
True Positives (TP) are pixels correctly predicted as positive, while False Positives (FP) are pixels incorrectly predicted as positive. Precision measures the proportion of predicted positive pixels that are actually positive, emphasizing the minimization of FP. However, precision alone does not fully assess model performance since it ignores False Negatives (FN), which are actual positive pixels not predicted as such. Therefore, metrics like recall are also used to provide a comprehensive evaluation of the model's performance.

In the current field of semantic scene completion, the evaluation indicators used show a certain degree of homogeneity, mainly focusing on quantifying the accuracy of semantic label classification. However, there is currently a lack of dedicated evaluation indicators for a comprehensive assessment of the quality of scene completion, especially the measurement of the degree of scene completion.
\begin{figure}[!t]
    \centering
    \includegraphics[width=3in]{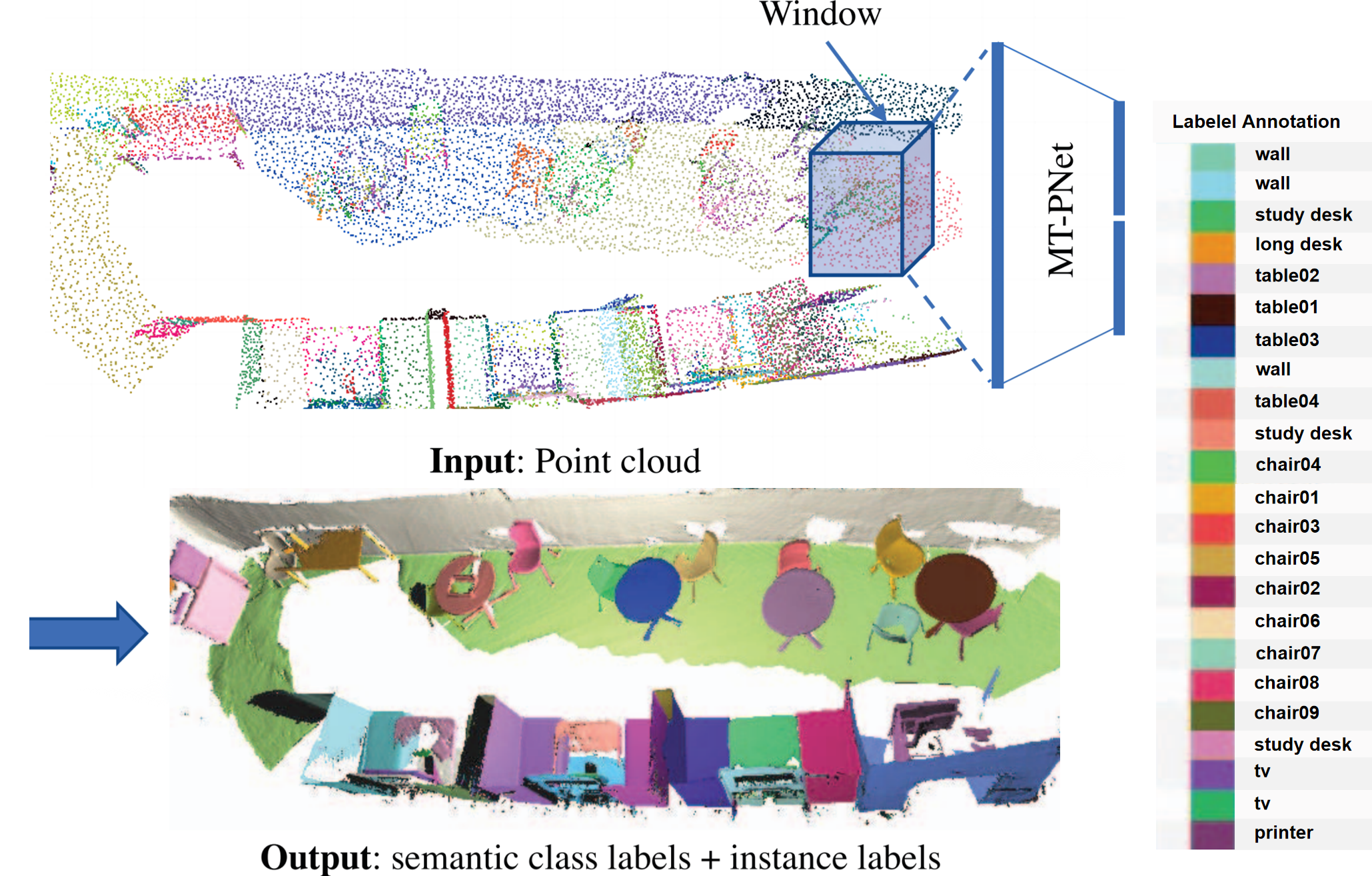}
    \caption{Schematic diagram of semantic instance joint segmentation method \cite{Pham8953238}.}
    \label{Fig:pointcloud understanding}
\end{figure}

\subsection{\textbf{Point Cloud Understanding}}
In this article, we distinguish point cloud understanding from point cloud scene understanding. The latter involves perceiving, interpreting, and reasoning about the environment to identify objects, their relationships, and the overall layout and structure of the point cloud scene. Point cloud scene understanding typically includes subtasks such as object detection, semantic segmentation, pose estimation, and scene graph generation. We posit that point cloud understanding encompasses point cloud scene understanding but extends beyond it. It is a new, integrated task based on traditional point cloud tasks that leverages shared underlying features across multiple information dimensions and modalities to enhance flexibility. Similar to the underlying logic of point cloud semantic scene completion (see \ref{Point cloud semantic scene completion}), point cloud understanding reorganizes the subtasks in scene understanding, optimizing tasks with high coupling or complementarity. Before the concept of semantic scene completion was introduced, the collaborative optimization of semantic segmentation and scene completion was categorized under point cloud scene understanding, as noted in \cite{Nie9578585}. Furthermore, the exsiting researchs has linked the development of point cloud backbone networks to point cloud understanding tasks. The main reason is that after optimizing the network structure, the underlying features of the point cloud can be more accurately captured, achieving a more comprehensive and in-depth understanding.

Semantic information serves two main roles in point cloud understanding. First, it acts as a cross-task coherence mechanism, facilitating effective integration and collaboration between different tasks. Second, as a low-dimensional representation derived from high-dimensional features, it aids in the training process of cross-modal learning, enhancing the model's ability to understand point cloud data.
\subsubsection*{\bf Integration Tasks in Point Clouds}

In addition to the point cloud semantic scene completion technology that has gradually attracted attention recently, we have noticed the diversity of integration tasks. Therefore, we provide a comprehensive overview of this type of tasks. Semantics plays the role of implicit feature communication in integration tasks and provides clues for inductive reasoning in the interaction of different tasks.

\cite{Pham8953238},\cite{Wang8953321},\cite{Hu8590720},\cite{Zhao9932589} and others developed a semantic instance joint segmentation method that simultaneously performs instance and semantic segmentation tasks on 3D point clouds. This approach leverages the interdependency between object instances and object semantics, as depicted in the figure \ref{Fig:pointcloud understanding}. \cite{Pham8953238} introduced a Multi Task Point Network (MT-PNet), which embeds 3D points into high-dimensional feature vectors after predicting their object categories in a point cloud. These vectors are then clustered based on object semantics, promoting proximity among similar objects and enhancing discrimination between different objects. Similarly, \cite{Wang8953321}'s semantic-aware point-level instance embedding method improves instance segmentation through semantic segmentation. To address issues of under-segmentation and over-segmentation, \cite{Hu8590720} divided the point cloud into surface patches and proposed a new patch segmentation and classification framework with multi-scale processing. This approach combines unsupervised clustering of patches with supervised methods to enhance the model's generalization capabilities. \cite{Zhao9932589} employs joint instance and semantic segmentation (JISS) modules to enable mutual utilization between the two tasks, sharing multiple layers of features through a common underlying backbone network. JISS converts semantic features into instance embedding space and then fuses these transformed features with instance features to facilitate instance segmentation.


Fang et al. \cite{fang202414563} proposed a general framework called PIC for 3D visual cues, which includes separate models for four tasks: point cloud reconstruction, denoising, registration, and part segmentation, along with a multi-task head. The framework employs a shared backbone model using a contextual learning paradigm to interpret 3D point clouds. The performance of this framework on all four tasks is comparable to that of single-task models and surpasses existing multi-task models.

\subsubsection*{\bf{Multi-modality}}
Beyond integration tasks, research in point cloud understanding also explores the sharing mechanism of cross-modal information at the underlying feature level. Afham et al. \cite{Afham99029912} integrated feature consistency relationships within and between modalities to achieve two-dimensional to three-dimensional correspondence using end-to-end self-supervised learning.

In the realm of two-dimensional representation, the CLIP method \cite{Radford87488763} was pioneering in aligning images and semantic concepts, demonstrating that knowledge from different modalities can significantly enhance conceptual understanding. Building on this, Zhang et al. \cite{Zhang9878980} proposed PointCLIP, a method for point clouds. PointCLIP first converts the 3D point cloud into a series of depth maps and then directly applies CLIP for zero-shot 3D classification. Similarly, CrossNet \cite{Wu10147273} combines 2D and 3D data for joint learning to achieve deep feature mining.

Xue et al. \cite{Xue10203465} introduced a method for learning a unified representation of language, images, and point clouds (ULIP) to improve the understanding of three-dimensional scenes with limited data. ULIP, as a backbone network, has demonstrated superior performance. Unlike PointCLIP, the ULIP method focuses on learning a unified representation across images, semantic concepts, and point clouds, significantly enhancing the comprehension of three-dimensional scenes.

Despite advancements in multi-modal exploration in the three-dimensional field, there is \textbf{currently a lack of authoritative datasets that integrate semantics, images, and point clouds. Most methods rely on existing two-dimensional datasets for independent expansion}.

\subsubsection*{\bf{Network Architecture Optimization}}
Moreover, optimizing the neural network architecture for point cloud feature extraction is a significant area of research in enhancing point cloud understanding. Improvements in network design can enable the extraction of features at multiple levels, leading to better performance across multiple tasks. Therefore, Transformers \cite{DuanS238839},\cite{Park2181823} and Mamba \cite{zhang2403007},\cite{liu24036467}, which have garnered significant interest in recent years, have been applied to point cloud data with promising results.

\section{Conclusion}

In this paper, we review the latest developments in semantics in point clouds, focusing on traditional tasks and emerging tasks introduced by semantics. Semantic information serves two primary roles in point clouds: explicit representation and implicit derivation. Explicit representations include semantic segmentation, 3D dense captioning, and scene graph prediction. Implicit derivation encompasses tasks such as point cloud compression, registration, reconstruction, and semantic scene completion.

Different tasks in point cloud processing often provide mutual references. Due to the high cost of obtaining labeled 3D point cloud data, there is a growing interest in applying semi-supervised and unsupervised methods, following the success of fully supervised approaches. Furthermore, novel tasks in 2D representation are being transferred to 3D point cloud representation. The progress in point cloud semantics is influenced by advancements in both 2D representation and other 3D representations, such as NeRF. Successful semantic-related tasks in other domains are being adapted to 3D point cloud representations. Additionally, multi-modal and cross-dimensional information enhances various models.

We conducted an in-depth analysis of each task, covering its contributions, challenges, recent advances in low-dimensional or other 3D representations, and evaluation metrics. Most emerging tasks still use semantic segmentation metrics, emphasizing segmentation accuracy. For instance, tasks like semantic scene completion and scene graph prediction inherit semantic segmentation metrics, primarily focusing on label accuracy. However, there are gaps in evaluating the depth of completion tasks and scene graph prediction.

This study systematically reviews relevant datasets from multiple task perspectives. The application of semantic information in 3D point cloud representation has not been thoroughly summarized. Similar tasks may be described using different concepts, leading to academic inconsistencies. For example, research similar to "semantic scene completion" existed before the term was clearly defined, often described as "scene understanding." Even within the same conceptual framework, there can be significant differences and a lack of strict definition. For instance, "point cloud understanding" might refer to developing backbone networks for extracting point cloud features or extending to the comprehensive application of multi-modal information. This ambiguity makes it challenging to classify related research clearly, even using keywords.

These challenges contribute to differing interpretations among researchers. This study aims to clarify relevant concepts to promote coherence and clear communication in future academic research. Considering the significant interest in large language models and diffusion models in recent years, we systematically searched and evaluated their application in each task. Our detailed discussion and review of semantic information processing in point clouds indicate that the application of semantic information in 3D point cloud representation will expand and deepen.

Future research in explicit representation will delve deeper into tasks such as semantic segmentation, 3D dense captioning, and scene graph prediction, with a focus on accuracy, real-time performance, and robustness. In the field of implicit derivation, there will be an increased emphasis on the extraction and utilization of point cloud features. Cross-modal information fusion will present new opportunities for point cloud understanding tasks, such as multi-modal scene understanding and cross-modal information retrieval. As new tasks emerge, datasets and evaluation metrics will be refined to better support task evaluation.

In conclusion, as technology advances and applications deepen, the use of semantic information in 3D point cloud representation will become more widespread and efficient. Future research will focus on enhancing task performance, integrating cross-modal information, and driving innovation and breakthroughs in point cloud processing.

 




\bibliographystyle{IEEEtran}
\bibliography{ref.bib}

\begin{thebibliography}{100}
\providecommand{\url}[1]{#1}
\csname url@samestyle\endcsname
\providecommand{\newblock}{\relax}
\providecommand{\bibinfo}[2]{#2}
\providecommand{\BIBentrySTDinterwordspacing}{\spaceskip=0pt\relax}
\providecommand{\BIBentryALTinterwordstretchfactor}{4}
\providecommand{\BIBentryALTinterwordspacing}{\spaceskip=\fontdimen2\font plus
\BIBentryALTinterwordstretchfactor\fontdimen3\font minus \fontdimen4\font\relax}
\providecommand{\BIBforeignlanguage}[2]{{%
\expandafter\ifx\csname l@#1\endcsname\relax
\typeout{** WARNING: IEEEtran.bst: No hyphenation pattern has been}%
\typeout{** loaded for the language `#1'. Using the pattern for}%
\typeout{** the default language instead.}%
\else
\language=\csname l@#1\endcsname
\fi
#2}}
\providecommand{\BIBdecl}{\relax}
\BIBdecl

\bibitem{Kähler7165673}
O.~Kähler, V.~Adrian~Prisacariu, C.~Yuheng~Ren, X.~Sun, P.~Torr, and D.~Murray, ``Very high frame rate volumetric integration of depth images on mobile devices,'' \emph{IEEE Transactions on Visualization and Computer Graphics}, vol.~21, no.~11, pp. 1241--1250, 2015.

\bibitem{Laine5620900}
S.~Laine and T.~Karras, ``Efficient sparse voxel octrees,'' \emph{IEEE Transactions on Visualization and Computer Graphics}, vol.~17, no.~8, pp. 1048--1059, 2011.

\bibitem{Chia284492}
C.-W. Liao and G.~Medioni, ``Surface approximation of a cloud of 3d points,'' in \emph{Proceedings of 1994 IEEE 2nd CAD-Based Vision Workshop}, 1994, pp. 274--281.

\bibitem{CATMULL1978350}
\BIBentryALTinterwordspacing
E.~Catmull and J.~Clark, ``Recursively generated b-spline surfaces on arbitrary topological meshes,'' \emph{Computer-Aided Design}, vol.~10, no.~6, pp. 350--355, 1978. [Online]. Available: \url{https://www.sciencedirect.com/science/article/pii/0010448578901100}
\BIBentrySTDinterwordspacing

\bibitem{Catmull1974ASA}
E.~E. Catmull, ``A subdivision algorithm for computer display of curved surfaces.''\hskip 1em plus 0.5em minus 0.4em\relax The University of Utah, 1974, aAI7504786.

\bibitem{NeRF58452824}
\BIBentryALTinterwordspacing
B.~Mildenhall, P.~P. Srinivasan, M.~Tancik, J.~T. Barron, R.~Ramamoorthi, and R.~Ng, ``Nerf: Representing scenes as neural radiance fields for view synthesis,'' in \emph{Computer Vision – ECCV 2020: 16th European Conference, Glasgow, UK, August 23–28, 2020, Proceedings, Part I}.\hskip 1em plus 0.5em minus 0.4em\relax Berlin, Heidelberg: Springer-Verlag, 2020, p. 405–421. [Online]. Available: \url{https://doi.org/10.1007/978-3-030-58452-8_24}
\BIBentrySTDinterwordspacing

\bibitem{ALZUBI201520}
\BIBentryALTinterwordspacing
A.~Alzu'bi, A.~Amira, and N.~Ramzan, ``Semantic content-based image retrieval: A comprehensive study,'' \emph{Journal of Visual Communication and Image Representation}, vol.~32, pp. 20--54, 2015. [Online]. Available: \url{https://www.sciencedirect.com/science/article/pii/S1047320315001327}
\BIBentrySTDinterwordspacing

\bibitem{Mafla9423139}
A.~Mafla, R.~S. Rezende, L.~Gómez, D.~Larlus, and D.~Karatzas, ``Stacmr: Scene-text aware cross-modal retrieval,'' in \emph{2021 IEEE Winter Conference on Applications of Computer Vision (WACV)}, 2021, pp. 2219--2229.

\bibitem{Chai10377456}
W.~Chai, X.~Guo, G.~Wang, and Y.~Lu, ``Stablevideo: Text-driven consistency-aware diffusion video editing,'' in \emph{2023 IEEE/CVF International Conference on Computer Vision (ICCV)}, 2023, pp. 22\,983--22\,993.

\bibitem{Gu10041185}
Y.~Gu, H.~Xu, Y.~Quan, W.~Chen, and J.~Zheng, ``Orsi salient object detection via bidimensional attention and full-stage semantic guidance,'' \emph{IEEE Transactions on Geoscience and Remote Sensing}, vol.~61, pp. 1--13, 2023.

\bibitem{Chang9428366}
J.~Chang, Z.~Zhao, L.~Yang, C.~Jia, J.~Zhang, and S.~Ma, ``Thousand to one: Semantic prior modeling for conceptual coding,'' in \emph{2021 IEEE International Conference on Multimedia and Expo (ICME)}, 2021, pp. 1--6.

\bibitem{Qi9930878}
Q.~Qi, K.~Li, H.~Zheng, X.~Gao, G.~Hou, and K.~Sun, ``Sguie-net: Semantic attention guided underwater image enhancement with multi-scale perception,'' \emph{IEEE Transactions on Image Processing}, vol.~31, pp. 6816--6830, 2022.

\bibitem{Wang10091192}
S.~Wang, Z.~Wang, H.~Li, J.~Chang, W.~Ouyang, and Q.~Tian, ``Semantic-guided information alignment network for fine-grained image recognition,'' \emph{IEEE Transactions on Circuits and Systems for Video Technology}, vol.~33, no.~11, pp. 6558--6570, 2023.

\bibitem{Zhao9690112}
L.~Zhao, K.-K. Ma, Z.~Liu, Q.~Yin, and J.~Chen, ``Real-time scene-aware lidar point cloud compression using semantic prior representation,'' \emph{IEEE Transactions on Circuits and Systems for Video Technology}, vol.~32, no.~8, pp. 5623--5637, 2022.

\bibitem{Shu9860002}
Y.~Shu, Z.~Hou, B.~Xiao, X.~Bi, X.~Luan, and W.~Li, ``Partial-to-partial point cloud registration based on multi-level semantic-structural cognition,'' in \emph{2022 IEEE International Conference on Multimedia and Expo (ICME)}, 2022, pp. 1--6.

\bibitem{Wang252615362}
\BIBentryALTinterwordspacing
T.~Wang, Q.~Wang, H.-B. Ai, and L.~Zhang, ``Semantics-and-primitives-guided indoor 3d reconstruction from point clouds,'' \emph{Remote. Sens.}, vol.~14, p. 4820, 2022. [Online]. Available: \url{https://api.semanticscholar.org/CorpusID:252615362}
\BIBentrySTDinterwordspacing

\bibitem{Cai9879358}
D.~Cai, L.~Zhao, J.~Zhang, L.~Sheng, and D.~Xu, ``3djcg: A unified framework for joint dense captioning and visual grounding on 3d point clouds,'' in \emph{2022 IEEE/CVF Conference on Computer Vision and Pattern Recognition (CVPR)}, 2022, pp. 16\,443--16\,452.

\bibitem{Wang10205194}
Z.~Wang, B.~Cheng, L.~Zhao, D.~Xu, Y.~Tang, and L.~Sheng, ``Vl-sat: Visual-linguistic semantics assisted training for 3d semantic scene graph prediction in point cloud,'' in \emph{2023 IEEE/CVF Conference on Computer Vision and Pattern Recognition (CVPR)}, 2023, pp. 21\,560--21\,569.

\bibitem{Li10007036}
J.~Li, Q.~Song, X.~Yan, Y.~Chen, and R.~Huang, ``From front to rear: 3d semantic scene completion through planar convolution and attention-based network,'' \emph{IEEE Transactions on Multimedia}, vol.~25, pp. 8294--8307, 2023.

\bibitem{Xue10203465}
L.~Xue, M.~Gao, C.~Xing, R.~Martín-Martín, J.~Wu, C.~Xiong, R.~Xu, J.~C. Niebles, and S.~Savarese, ``Ulip: Learning a unified representation of language, images, and point clouds for 3d understanding,'' in \emph{2023 IEEE/CVF Conference on Computer Vision and Pattern Recognition (CVPR)}, 2023, pp. 1179--1189.

\bibitem{Uy9009007}
M.~A. Uy, Q.-H. Pham, B.-S. Hua, T.~Nguyen, and S.-K. Yeung, ``Revisiting point cloud classification: A new benchmark dataset and classification model on real-world data,'' in \emph{2019 IEEE/CVF International Conference on Computer Vision (ICCV)}, 2019, pp. 1588--1597.

\bibitem{Shi9018080}
S.~Shi, Z.~Wang, J.~Shi, X.~Wang, and H.~Li, ``From points to parts: 3d object detection from point cloud with part-aware and part-aggregation network,'' \emph{IEEE Transactions on Pattern Analysis and Machine Intelligence}, vol.~43, no.~8, pp. 2647--2664, 2021.

\bibitem{cao20193d}
C.~Cao, M.~Preda, and T.~Zaharia, ``3d point cloud compression: A survey,'' in \emph{Proceedings of the 24th International Conference on 3D Web Technology}, 2019, pp. 1--9.

\bibitem{Qin10076895}
Z.~Qin, H.~Yu, C.~Wang, Y.~Guo, Y.~Peng, S.~Ilic, D.~Hu, and K.~Xu, ``Geotransformer: Fast and robust point cloud registration with geometric transformer,'' \emph{IEEE Transactions on Pattern Analysis and Machine Intelligence}, vol.~45, no.~8, pp. 9806--9821, 2023.

\bibitem{Far10301359}
A.~Farshian, M.~Götz, G.~Cavallaro, C.~Debus, M.~Nießner, J.~A. Benediktsson, and A.~Streit, ``Deep-learning-based 3-d surface reconstruction—a survey,'' \emph{Proceedings of the IEEE}, vol. 111, no.~11, pp. 1464--1501, 2023.

\bibitem{Huang20202413}
\BIBentryALTinterwordspacing
Z.~Huang, Y.~Wen, Z.~Wang, J.~Ren, and K.~Jia, ``Surface reconstruction from point clouds: A survey and a benchmark,'' \emph{ArXiv}, vol. abs/2205.02413, 2022. [Online]. Available: \url{https://api.semanticscholar.org/CorpusID:248525170}
\BIBentrySTDinterwordspacing

\bibitem{Ding10203570}
R.~Ding, J.~Yang, C.~Xue, W.~Zhang, S.~Bai, and X.~Qi, ``Pla: Language-driven open-vocabulary 3d scene understanding,'' in \emph{2023 IEEE/CVF Conference on Computer Vision and Pattern Recognition (CVPR)}, 2023, pp. 7010--7019.

\bibitem{Chang9661322}
X.~Chang, P.~Ren, P.~Xu, Z.~Li, X.~Chen, and A.~Hauptmann, ``A comprehensive survey of scene graphs: Generation and application,'' \emph{IEEE Transactions on Pattern Analysis and Machine Intelligence}, vol.~45, no.~1, pp. 1--26, 2023.

\bibitem{Xia149514967}
Y.~Xia, L.~Shi, Z.~Ding, J.~F. Henriques, and D.~Cremers, in \emph{Proceedings of the IEEE/CVF Conference on Computer Vision and Pattern Recognition (CVPR)}, June 2024, pp. 14\,958--14\,967.

\bibitem{Li9435044}
J.~Li, P.~Wang, K.~Han, and Y.~Liu, ``Anisotropic convolutional neural networks for rgb-d based semantic scene completion,'' \emph{IEEE Transactions on Pattern Analysis and Machine Intelligence}, vol.~44, no.~11, pp. 8125--8138, 2022.

\bibitem{hacke170403847}
T.~Hackel, N.~Savinov, L.~Ladicky, J.~D. Wegner, K.~Schindler, and M.~Pollefeys, ``Semantic3d.net: A new large-scale point cloud classification benchmark,'' \emph{arXiv preprint arXiv:1704.03847}, 2017.

\bibitem{behley929719307}
J.~Behley, M.~Garbade, A.~Milioto, J.~Quenzel, S.~Behnke, C.~Stachniss, and J.~Gall, ``Semantickitti: A dataset for semantic scene understanding of lidar sequences,'' in \emph{Proceedings of the IEEE/CVF international conference on computer vision}, 2019, pp. 9297--9307.

\bibitem{geiger12311237}
A.~Geiger, P.~Lenz, C.~Stiller, and R.~Urtasun, ``Vision meets robotics: The kitti dataset,'' \emph{The International Journal of Robotics Research}, vol.~32, no.~11, pp. 1231--1237, 2013.

\bibitem{armeni15341543}
I.~Armeni, O.~Sener, A.~R. Zamir, H.~Jiang, I.~Brilakis, M.~Fischer, and S.~Savarese, ``3d semantic parsing of large-scale indoor spaces,'' in \emph{Proceedings of the IEEE conference on computer vision and pattern recognition}, 2016, pp. 1534--1543.

\bibitem{pan687693}
Y.~Pan, B.~Gao, J.~Mei, S.~Geng, C.~Li, and H.~Zhao, ``Semanticposs: A point cloud dataset with large quantity of dynamic instances,'' in \emph{2020 IEEE Intelligent Vehicles Symposium (IV)}.\hskip 1em plus 0.5em minus 0.4em\relax IEEE, 2020, pp. 687--693.

\bibitem{Hu9578083}
Q.~Hu, B.~Yang, S.~Khalid, W.~Xiao, N.~Trigoni, and A.~Markham, ``Towards semantic segmentation of urban-scale 3d point clouds: A dataset, benchmarks and challenges,'' in \emph{2021 IEEE/CVF Conference on Computer Vision and Pattern Recognition (CVPR)}, 2021, pp. 4975--4985.

\bibitem{chang170906158}
A.~Chang, A.~Dai, T.~Funkhouser, M.~Halber, M.~Niessner, M.~Savva, S.~Song, A.~Zeng, and Y.~Zhang, ``Matterport3d: Learning from rgb-d data in indoor environments,'' \emph{arXiv preprint arXiv:1709.06158}, 2017.

\bibitem{hua92101}
B.-S. Hua, Q.-H. Pham, D.~T. Nguyen, M.-K. Tran, L.-F. Yu, and S.-K. Yeung, ``Scenenn: A scene meshes dataset with annotations,'' in \emph{2016 fourth international conference on 3D vision (3DV)}.\hskip 1em plus 0.5em minus 0.4em\relax Ieee, 2016, pp. 92--101.

\bibitem{Chang2015ShapeNetAI}
\BIBentryALTinterwordspacing
A.~X. Chang, T.~A. Funkhouser, L.~J. Guibas, P.~Hanrahan, Q.-X. Huang, Z.~Li, S.~Savarese, M.~Savva, S.~Song, H.~Su, J.~Xiao, L.~Yi, and F.~Yu, ``Shapenet: An information-rich 3d model repository,'' \emph{ArXiv}, vol. abs/1512.03012, 2015. [Online]. Available: \url{https://api.semanticscholar.org/CorpusID:2554264}
\BIBentrySTDinterwordspacing

\bibitem{wu19121920}
Z.~Wu, S.~Song, A.~Khosla, F.~Yu, L.~Zhang, X.~Tang, and J.~Xiao, ``3d shapenets: A deep representation for volumetric shapes,'' in \emph{Proceedings of the IEEE conference on computer vision and pattern recognition}, 2015, pp. 1912--1920.

\bibitem{dai58285839}
A.~Dai, A.~X. Chang, M.~Savva, M.~Halber, T.~Funkhouser, and M.~Nie{\ss}ner, ``Scannet: Richly-annotated 3d reconstructions of indoor scenes,'' in \emph{Proceedings of the IEEE conference on computer vision and pattern recognition}, 2017, pp. 5828--5839.

\bibitem{langlois40084015}
P.-A. Langlois, Y.~Xiao, A.~Boulch, and R.~Marlet, ``Vasad: a volume and semantic dataset for building reconstruction from point clouds,'' in \emph{2022 26th International Conference on Pattern Recognition (ICPR)}.\hskip 1em plus 0.5em minus 0.4em\relax IEEE, 2022, pp. 4008--4015.

\bibitem{Wald9156565}
J.~Wald, H.~Dhamo, N.~Navab, and F.~Tombari, ``Learning 3d semantic scene graphs from 3d indoor reconstructions,'' in \emph{2020 IEEE/CVF Conference on Computer Vision and Pattern Recognition (CVPR)}, 2020, pp. 3960--3969.

\bibitem{achlioptas422440}
P.~Achlioptas, A.~Abdelreheem, F.~Xia, M.~Elhoseiny, and L.~Guibas, ``Referit3d: Neural listeners for fine-grained 3d object identification in real-world scenes,'' in \emph{Computer Vision--ECCV 2020: 16th European Conference, Glasgow, UK, August 23--28, 2020, Proceedings, Part I 16}.\hskip 1em plus 0.5em minus 0.4em\relax Springer, 2020, pp. 422--440.

\bibitem{chen202221}
D.~Z. Chen, A.~X. Chang, and M.~Nie{\ss}ner, ``Scanrefer: 3d object localization in rgb-d scans using natural language,'' in \emph{European conference on computer vision}.\hskip 1em plus 0.5em minus 0.4em\relax Springer, 2020, pp. 202--221.

\bibitem{silberman746760}
N.~Silberman, D.~Hoiem, P.~Kohli, and R.~Fergus, ``Indoor segmentation and support inference from rgbd images,'' in \emph{Computer Vision--ECCV 2012: 12th European Conference on Computer Vision, Florence, Italy, October 7-13, 2012, Proceedings, Part V 12}.\hskip 1em plus 0.5em minus 0.4em\relax Springer, 2012, pp. 746--760.

\bibitem{song808816}
S.~Song and J.~Xiao, ``Deep sliding shapes for amodal 3d object detection in rgb-d images,'' in \emph{Proceedings of the IEEE conference on computer vision and pattern recognition}, 2016, pp. 808--816.

\bibitem{avetisyan26142623}
A.~Avetisyan, M.~Dahnert, A.~Dai, M.~Savva, A.~X. Chang, and M.~Nie{\ss}ner, ``Scan2cad: Learning cad model alignment in rgb-d scans,'' in \emph{Proceedings of the IEEE/CVF Conference on computer vision and pattern recognition}, 2019, pp. 2614--2623.

\bibitem{NEURIPS2020_cd3afef9}
\BIBentryALTinterwordspacing
X.~Wang, R.~Zhang, T.~Kong, L.~Li, and C.~Shen, ``Solov2: Dynamic and fast instance segmentation,'' in \emph{Advances in Neural Information Processing Systems}, H.~Larochelle, M.~Ranzato, R.~Hadsell, M.~Balcan, and H.~Lin, Eds., vol.~33.\hskip 1em plus 0.5em minus 0.4em\relax Curran Associates, Inc., 2020, pp. 17\,721--17\,732. [Online]. Available: \url{https://proceedings.neurips.cc/paper_files/paper/2020/file/cd3afef9b8b89558cd56638c3631868a-Paper.pdf}
\BIBentrySTDinterwordspacing

\bibitem{Wang9536421}
X.~Wang, R.~Zhang, C.~Shen, T.~Kong, and L.~Li, ``Solo: A simple framework for instance segmentation,'' \emph{IEEE Transactions on Pattern Analysis and Machine Intelligence}, vol.~44, no.~11, pp. 8587--8601, 2022.

\bibitem{Sun10030147}
W.~Sun, D.~Rebain, R.~Liao, V.~Tankovich, S.~Yazdani, K.~M. Yi, and A.~Tagliasacchi, ``Neuralbf: Neural bilateral filtering for top-down instance segmentation on point clouds,'' in \emph{2023 IEEE/CVF Winter Conference on Applications of Computer Vision (WACV)}, 2023, pp. 551--560.

\bibitem{Matu7353481}
D.~Maturana and S.~Scherer, ``Voxnet: A 3d convolutional neural network for real-time object recognition,'' in \emph{2015 IEEE/RSJ International Conference on Intelligent Robots and Systems (IROS)}, 2015, pp. 922--928.

\bibitem{Le8579057}
T.~Le and Y.~Duan, ``Pointgrid: A deep network for 3d shape understanding,'' in \emph{2018 IEEE/CVF Conference on Computer Vision and Pattern Recognition}, 2018, pp. 9204--9214.

\bibitem{Rieg8100184}
G.~Riegler, A.~O. Ulusoy, and A.~Geiger, ``Octnet: Learning deep 3d representations at high resolutions,'' in \emph{2017 IEEE Conference on Computer Vision and Pattern Recognition (CVPR)}, 2017, pp. 6620--6629.

\bibitem{Hu9710530}
Z.~Hu, X.~Bai, J.~Shang, R.~Zhang, J.~Dong, X.~Wang, G.~Sun, H.~Fu, and C.-L. Tai, ``Vmnet: Voxel-mesh network for geodesic-aware 3d semantic segmentation,'' in \emph{2021 IEEE/CVF International Conference on Computer Vision (ICCV)}, 2021, pp. 15\,468--15\,478.

\bibitem{Wang171517168}
Y.~Wang, Y.~Chen, X.~Liao, L.~Fan, and Z.~Zhang, ``Panoocc: Unified occupancy representation for camera-based 3d panoptic segmentation,'' in \emph{Proceedings of the IEEE/CVF Conference on Computer Vision and Pattern Recognition (CVPR)}, June 2024, pp. 17\,158--17\,168.

\bibitem{Tang220919732}
\BIBentryALTinterwordspacing
H.~Tang, Z.~Liu, S.~Zhao, Y.~Lin, J.~Lin, H.~Wang, and S.~Han, ``Searching efficient 3d architectures with sparse point-voxel convolution,'' \emph{ArXiv}, vol. abs/2007.16100, 2020. [Online]. Available: \url{https://api.semanticscholar.org/CorpusID:220919732}
\BIBentrySTDinterwordspacing

\bibitem{Zhang20010028}
S.~Zhang, X.~Fei, and Y.~Duan, ``Geoauxnet: Towards universal 3d representation learning for multi-sensor point clouds,'' in \emph{Proceedings of the IEEE/CVF Conference on Computer Vision and Pattern Recognition (CVPR)}, June 2024, pp. 20\,019--20\,028.

\bibitem{lawin2095107}
F.~J. Lawin, M.~Danelljan, P.~Tosteberg, G.~Bhat, F.~S. Khan, and M.~Felsberg, ``Deep projective 3d semantic segmentation,'' in \emph{Computer Analysis of Images and Patterns: 17th International Conference, CAIP 2017, Ystad, Sweden, August 22-24, 2017, Proceedings, Part I 17}.\hskip 1em plus 0.5em minus 0.4em\relax Springer, 2017, pp. 95--107.

\bibitem{Rong9430559}
M.~Rong, H.~Cui, Z.~Hu, H.~Jiang, H.~Liu, and S.~Shen, ``Active learning based 3d semantic labeling from images and videos,'' \emph{IEEE Transactions on Circuits and Systems for Video Technology}, vol.~32, no.~12, pp. 8101--8115, 2022.

\bibitem{Rong10158507}
M.~Rong, H.~Cui, and S.~Shen, ``Efficient 3d scene semantic segmentation via active learning on rendered 2d images,'' \emph{IEEE Transactions on Image Processing}, vol.~32, pp. 3521--3535, 2023.

\bibitem{alons5435439}
I.~Alonso, L.~Riazuelo, L.~Montesano, and A.~C. Murillo, ``3d-mininet: Learning a 2d representation from point clouds for fast and efficient 3d lidar semantic segmentation,'' \emph{IEEE Robotics and Automation Letters}, vol.~5, no.~4, pp. 5432--5439, 2020.

\bibitem{Milio8967762}
A.~Milioto, I.~Vizzo, J.~Behley, and C.~Stachniss, ``Rangenet ++: Fast and accurate lidar semantic segmentation,'' in \emph{2019 IEEE/RSJ International Conference on Intelligent Robots and Systems (IROS)}, 2019, pp. 4213--4220.

\bibitem{Wu8462926}
B.~Wu, A.~Wan, X.~Yue, and K.~Keutzer, ``Squeezeseg: Convolutional neural nets with recurrent crf for real-time road-object segmentation from 3d lidar point cloud,'' in \emph{2018 IEEE International Conference on Robotics and Automation (ICRA)}, 2018, pp. 1887--1893.

\bibitem{Wu8793495}
B.~Wu, X.~Zhou, S.~Zhao, X.~Yue, and K.~Keutzer, ``Squeezesegv2: Improved model structure and unsupervised domain adaptation for road-object segmentation from a lidar point cloud,'' in \emph{2019 International Conference on Robotics and Automation (ICRA)}, 2019, pp. 4376--4382.

\bibitem{xu28846293}
C.~Xu, B.~Wu, Z.~Wang, W.~Zhan, P.~Vajda, K.~Keutzer, and M.~Tomizuka, ``Squeezesegv3: Spatially-adaptive convolution for efficient point-cloud segmentation,'' in \emph{Computer Vision--ECCV 2020: 16th European Conference, Glasgow, UK, August 23--28, 2020, Proceedings, Part XXVIII 16}.\hskip 1em plus 0.5em minus 0.4em\relax Springer, 2020, pp. 1--19.

\bibitem{cortinhal207222}
T.~Cortinhal, G.~Tzelepis, and E.~Erdal~Aksoy, ``Salsanext: Fast, uncertainty-aware semantic segmentation of lidar point clouds,'' in \emph{Advances in Visual Computing: 15th International Symposium, ISVC 2020, San Diego, CA, USA, October 5--7, 2020, Proceedings, Part II 15}.\hskip 1em plus 0.5em minus 0.4em\relax Springer, 2020, pp. 207--222.

\bibitem{wang20185323}
Y.~Wang, T.~Shi, P.~Yun, L.~Tai, and M.~Liu, ``Pointseg: Real-time semantic segmentation based on 3d lidar point cloud,'' \emph{arXiv preprint arXiv:1807.06288}, 2018.

\bibitem{Rong10119167}
M.~Rong and S.~Shen, ``3d semantic segmentation of aerial photogrammetry models based on orthographic projection,'' \emph{IEEE Transactions on Circuits and Systems for Video Technology}, vol.~33, no.~12, pp. 7425--7437, 2023.

\bibitem{Jiang10220057}
F.~Jiang, H.~Gao, S.~Qiu, H.~Zhang, R.~Wan, and J.~Pu, ``Knowledge distillation from 3d to bird's-eye-view for lidar semantic segmentation,'' in \emph{2023 IEEE International Conference on Multimedia and Expo (ICME)}, 2023, pp. 402--407.

\bibitem{ng20205242}
M.~H. Ng, K.~Radia, J.~Chen, D.~Wang, I.~Gog, and J.~E. Gonzalez, ``Bev-seg: Bird's eye view semantic segmentation using geometry and semantic point cloud,'' \emph{arXiv preprint arXiv:2006.11436}, 2020.

\bibitem{chen2033331}
Q.~Chen and X.~Qi, ``Residual graph convolutional network for bird's-eye-view semantic segmentation,'' in \emph{Proceedings of the IEEE/CVF Winter Conference on Applications of Computer Vision}, 2024, pp. 3324--3331.

\bibitem{Liu10124335}
J.~Liu, Z.~Cao, J.~Yang, X.~Liu, Y.~Yang, and Z.~Qu, ``Bird's-eye-view semantic segmentation with two-stream compact depth transformation and feature rectification,'' \emph{IEEE Transactions on Intelligent Vehicles}, vol.~8, no.~11, pp. 4546--4558, 2023.

\bibitem{Ando10204428}
A.~Ando, S.~Gidaris, A.~Bursuc, G.~Puy, A.~Boulch, and R.~Marlet, ``Rangevit: Towards vision transformers for 3d semantic segmentation in autonomous driving,'' in \emph{2023 IEEE/CVF Conference on Computer Vision and Pattern Recognition (CVPR)}, 2023, pp. 5240--5250.

\bibitem{dosov201011929}
A.~Dosovitskiy, L.~Beyer, A.~Kolesnikov, D.~Weissenborn, X.~Zhai, T.~Unterthiner, M.~Dehghani, M.~Minderer, G.~Heigold, S.~Gelly \emph{et~al.}, ``An image is worth 16x16 words: Transformers for image recognition at scale,'' \emph{arXiv preprint arXiv:2010.11929}, 2020.

\bibitem{Char8099499}
R.~Q. Charles, H.~Su, M.~Kaichun, and L.~J. Guibas, ``Pointnet: Deep learning on point sets for 3d classification and segmentation,'' in \emph{2017 IEEE Conference on Computer Vision and Pattern Recognition (CVPR)}, 2017, pp. 77--85.

\bibitem{Qi5989836}
\BIBentryALTinterwordspacing
C.~R. Qi, L.~Yi, H.~Su, and L.~J. Guibas, ``Pointnet++: Deep hierarchical feature learning on point sets in a metric space,'' in \emph{Advances in Neural Information Processing Systems}, I.~Guyon, U.~V. Luxburg, S.~Bengio, H.~Wallach, R.~Fergus, S.~Vishwanathan, and R.~Garnett, Eds., vol.~30.\hskip 1em plus 0.5em minus 0.4em\relax Curran Associates, Inc., 2017. [Online]. Available: \url{https://proceedings.neurips.cc/paper_files/paper/2017/file/d8bf84be3800d12f74d8b05e9b89836f-Paper.pdf}
\BIBentrySTDinterwordspacing

\bibitem{Hu9156466}
Q.~Hu, B.~Yang, L.~Xie, S.~Rosa, Y.~Guo, Z.~Wang, N.~Trigoni, and A.~Markham, ``Randla-net: Efficient semantic segmentation of large-scale point clouds,'' in \emph{2020 IEEE/CVF Conference on Computer Vision and Pattern Recognition (CVPR)}, 2020, pp. 11\,105--11\,114.

\bibitem{Zhao9710703}
H.~Zhao, L.~Jiang, J.~Jia, P.~Torr, and V.~Koltun, ``Point transformer,'' in \emph{2021 IEEE/CVF International Conference on Computer Vision (ICCV)}, 2021, pp. 16\,239--16\,248.

\bibitem{Wu20228ece66}
\BIBentryALTinterwordspacing
X.~Wu, Y.~Lao, L.~Jiang, X.~Liu, and H.~Zhao, ``Point transformer v2: Grouped vector attention and partition-based pooling,'' in \emph{Advances in Neural Information Processing Systems}, S.~Koyejo, S.~Mohamed, A.~Agarwal, D.~Belgrave, K.~Cho, and A.~Oh, Eds., vol.~35.\hskip 1em plus 0.5em minus 0.4em\relax Curran Associates, Inc., 2022, pp. 33\,330--33\,342. [Online]. Available: \url{https://proceedings.neurips.cc/paper_files/paper/2022/file/d78ece6613953f46501b958b7bb4582f-Paper-Conference.pdf}
\BIBentrySTDinterwordspacing

\bibitem{wu20246357}
X.~Wu, L.~Jiang, P.-S. Wang, Z.~Liu, X.~Liu, Y.~Qiao, W.~Ouyang, T.~He, and H.~Zhao, ``Point transformer v3: Simpler, faster, stronger,'' in \emph{2024 IEEE/CVF Conference on Computer Vision and Pattern Recognition (CVPR)}, 2024.

\bibitem{Koma8953826}
A.~Komarichev, Z.~Zhong, and J.~Hua, ``A-cnn: Annularly convolutional neural networks on point clouds,'' in \emph{2019 IEEE/CVF Conference on Computer Vision and Pattern Recognition (CVPR)}, 2019, pp. 7413--7422.

\bibitem{Thom9010002}
H.~Thomas, C.~R. Qi, J.-E. Deschaud, B.~Marcotegui, F.~Goulette, and L.~Guibas, ``Kpconv: Flexible and deformable convolution for point clouds,'' in \emph{2019 IEEE/CVF International Conference on Computer Vision (ICCV)}, 2019, pp. 6410--6419.

\bibitem{Lu9577467}
T.~Lu, L.~Wang, and G.~Wu, ``Cga-net: Category guided aggregation for point cloud semantic segmentation,'' in \emph{2021 IEEE/CVF Conference on Computer Vision and Pattern Recognition (CVPR)}, 2021, pp. 11\,688--11\,697.

\bibitem{Yin10025770}
F.~Yin, Z.~Huang, T.~Chen, G.~Luo, G.~Yu, and B.~Fu, ``Dcnet: Large-scale point cloud semantic segmentation with discriminative and efficient feature aggregation,'' \emph{IEEE Transactions on Circuits and Systems for Video Technology}, vol.~33, no.~8, pp. 4083--4095, 2023.

\bibitem{Shuai9410334}
H.~Shuai, X.~Xu, and Q.~Liu, ``Backward attentive fusing network with local aggregation classifier for 3d point cloud semantic segmentation,'' \emph{IEEE Transactions on Image Processing}, vol.~30, pp. 4973--4984, 2021.

\bibitem{Liang9709996}
Z.~Liang, Z.~Li, S.~Xu, M.~Tan, and K.~Jia, ``Instance segmentation in 3d scenes using semantic superpoint tree networks,'' in \emph{2021 IEEE/CVF International Conference on Computer Vision (ICCV)}, 2021, pp. 2763--2772.

\bibitem{Deng9811904}
S.~Deng, Q.~Dong, B.~Liu, and Z.~Hu, ``Superpoint-guided semi-supervised semantic segmentation of 3d point clouds,'' in \emph{2022 International Conference on Robotics and Automation (ICRA)}, 2022, pp. 9214--9220.

\bibitem{Shi202106931}
\BIBentryALTinterwordspacing
X.~Shi, X.~Xu, K.~Chen, L.~Cai, C.-S. Foo, and K.~Jia, ``Label-efficient point cloud semantic segmentation: An active learning approach,'' \emph{ArXiv}, vol. abs/2101.06931, 2021. [Online]. Available: \url{https://api.semanticscholar.org/CorpusID:231632271}
\BIBentrySTDinterwordspacing

\bibitem{Chen1429920}
Z.~Chen, H.~Xu, W.~Chen, Z.~Zhou, H.~Xiao, B.~Sun, X.~Xie, and W.~kang, ``Pointdc: Unsupervised semantic segmentation of 3d point clouds via cross-modal distillation and super-voxel clustering,'' in \emph{Proceedings of the IEEE/CVF International Conference on Computer Vision (ICCV)}, October 2023, pp. 14\,290--14\,299.

\bibitem{Cheng1620350}
M.~Cheng, L.~Hui, J.~Xie, and J.~Yang, ``Sspc-net: Semi-supervised semantic 3d point cloud segmentation network,'' in \emph{Proceedings of the AAAI Conference on Artificial Intelligence(AAAI)}, vol.~35, no.~2, 2021, pp. 1140--1147.

\bibitem{Liu9578763}
Z.~Liu, X.~Qi, and C.-W. Fu, ``One thing one click: A self-training approach for weakly supervised 3d semantic segmentation,'' in \emph{2021 IEEE/CVF Conference on Computer Vision and Pattern Recognition (CVPR)}, 2021, pp. 1726--1736.

\bibitem{Land8578577}
L.~Landrieu and M.~Simonovsky, ``Large-scale point cloud semantic segmentation with superpoint graphs,'' in \emph{2018 IEEE/CVF Conference on Computer Vision and Pattern Recognition}, 2018, pp. 4558--4567.

\bibitem{Yang9919408}
F.~Yang, X.~Li, and J.~Shen, ``Nested architecture search for point cloud semantic segmentation,'' \emph{IEEE Transactions on Image Processing}, vol.~32, pp. 2889--2900, 2023.

\bibitem{Zheng3626449}
\BIBentryALTinterwordspacing
F.~Zheng, L.~Hui, J.~Xie, and H.~Zhang, ``Multi-scale superpoint network for 3d point cloud semantic segmentation,'' in \emph{Proceedings of the 5th ACM International Conference on Multimedia in Asia}, ser. MMAsia '23.\hskip 1em plus 0.5em minus 0.4em\relax New York, NY, USA: Association for Computing Machinery, 2024. [Online]. Available: \url{https://doi.org/10.1145/3595916.3626449}
\BIBentrySTDinterwordspacing

\bibitem{Xu9709941}
J.~Xu, R.~Zhang, J.~Dou, Y.~Zhu, J.~Sun, and S.~Pu, ``Rpvnet: A deep and efficient range-point-voxel fusion network for lidar point cloud segmentation,'' in \emph{2021 IEEE/CVF International Conference on Computer Vision (ICCV)}, 2021, pp. 16\,004--16\,013.

\bibitem{Liu7370345}
\BIBentryALTinterwordspacing
Z.~Liu, H.~Tang, Y.~Lin, and S.~Han, ``Point-voxel cnn for efficient 3d deep learning,'' in \emph{Advances in Neural Information Processing Systems}, H.~Wallach, H.~Larochelle, A.~Beygelzimer, F.~d\textquotesingle Alch\'{e}-Buc, E.~Fox, and R.~Garnett, Eds., vol.~32.\hskip 1em plus 0.5em minus 0.4em\relax Curran Associates, Inc., 2019, p. 965–975. [Online]. Available: \url{https://proceedings.neurips.cc/paper_files/paper/2019/file/5737034557ef5b8c02c0e46513b98f90-Paper.pdf}
\BIBentrySTDinterwordspacing

\bibitem{Li9730613}
J.~Li, M.~Li, Z.~Li, and S.~Peng, ``Super-voxel graph guided 3d point cloud denoising,'' in \emph{2022 14th International Conference on Computer Research and Development (ICCRD)}, 2022, pp. 276--280.

\bibitem{Hou9879674}
Y.~Hou, X.~Zhu, Y.~Ma, C.~C. Loy, and Y.~Li, ``Point-to-voxel knowledge distillation for lidar semantic segmentation,'' in \emph{2022 IEEE/CVF Conference on Computer Vision and Pattern Recognition (CVPR)}, 2022, pp. 8469--8478.

\bibitem{Chen1429099}
Z.~Chen, H.~Xu, W.~Chen, Z.~Zhou, H.~Xiao, B.~Sun, X.~Xie, and W.~kang, ``Pointdc: Unsupervised semantic segmentation of 3d point clouds via cross-modal distillation and super-voxel clustering,'' in \emph{Proceedings of the IEEE/CVF International Conference on Computer Vision (ICCV)}, October 2023, pp. 14\,290--14\,299.

\bibitem{Qiu10160496}
S.~Qiu, F.~Jiang, H.~Zhang, X.~Xue, and J.~Pu, ``Multi-to-single knowledge distillation for point cloud semantic segmentation,'' in \emph{2023 IEEE International Conference on Robotics and Automation (ICRA)}, 2023, pp. 9303--9309.

\bibitem{Zhang10205029}
L.~Zhang, R.~Dong, H.-S. Tai, and K.~Ma, ``Pointdistiller: Structured knowledge distillation towards efficient and compact 3d detection,'' in \emph{2023 IEEE/CVF Conference on Computer Vision and Pattern Recognition (CVPR)}, 2023, pp. 21\,791--21\,801.

\bibitem{chang87488757}
M.-F. Chang, J.~Lambert, P.~Sangkloy, J.~Singh, S.~Bak, A.~Hartnett, D.~Wang, P.~Carr, S.~Lucey, D.~Ramanan \emph{et~al.}, ``Argoverse: 3d tracking and forecasting with rich maps,'' in \emph{Proceedings of the IEEE/CVF conference on computer vision and pattern recognition}, 2019, pp. 8748--8757.

\bibitem{geyer20205223}
J.~Geyer, Y.~Kassahun, M.~Mahmudi, X.~Ricou, R.~Durgesh, A.~S. Chung, L.~Hauswald, V.~H. Pham, M.~M{\"u}hlegg, S.~Dorn \emph{et~al.}, ``A2d2: Audi autonomous driving dataset,'' \emph{arXiv preprint arXiv:2004.06320}, 2020.

\bibitem{huang954960}
X.~Huang, X.~Cheng, Q.~Geng, B.~Cao, D.~Zhou, P.~Wang, Y.~Lin, and R.~Yang, ``The apolloscape dataset for autonomous driving,'' in \emph{Proceedings of the IEEE conference on computer vision and pattern recognition workshops}, 2018, pp. 954--960.

\bibitem{Yi9578920}
L.~Yi, B.~Gong, and T.~Funkhouser, ``Complete \& label: A domain adaptation approach to semantic segmentation of lidar point clouds,'' in \emph{2021 IEEE/CVF Conference on Computer Vision and Pattern Recognition (CVPR)}, 2021, pp. 15\,358--15\,368.

\bibitem{Li10203290}
J.~Li, H.~Dai, H.~Han, and Y.~Ding, ``Mseg3d: Multi-modal 3d semantic segmentation for autonomous driving,'' in \emph{2023 IEEE/CVF Conference on Computer Vision and Pattern Recognition (CVPR)}, 2023, pp. 21\,694--21\,704.

\bibitem{Geiger6248074}
A.~Geiger, P.~Lenz, and R.~Urtasun, ``Are we ready for autonomous driving? the kitti vision benchmark suite,'' in \emph{2012 IEEE Conference on Computer Vision and Pattern Recognition}, 2012, pp. 3354--3361.

\bibitem{Caesar9156412}
H.~Caesar, V.~Bankiti, A.~H. Lang, S.~Vora, V.~E. Liong, Q.~Xu, A.~Krishnan, Y.~Pan, G.~Baldan, and O.~Beijbom, ``nuscenes: A multimodal dataset for autonomous driving,'' in \emph{2020 IEEE/CVF Conference on Computer Vision and Pattern Recognition (CVPR)}, 2020, pp. 11\,618--11\,628.

\bibitem{Zhuang9710693}
Z.~Zhuang, R.~Li, K.~Jia, Q.~Wang, Y.~Li, and M.~Tan, ``Perception-aware multi-sensor fusion for 3d lidar semantic segmentation,'' in \emph{2021 IEEE/CVF International Conference on Computer Vision (ICCV)}, 2021, pp. 16\,260--16\,270.

\bibitem{Peng9710520}
D.~Peng, Y.~Lei, W.~Li, P.~Zhang, and Y.~Guo, ``Sparse-to-dense feature matching: Intra and inter domain cross-modal learning in domain adaptation for 3d semantic segmentation,'' in \emph{2021 IEEE/CVF International Conference on Computer Vision (ICCV)}, 2021, pp. 7088--7097.

\bibitem{Zhao10128757}
L.~Zhao, H.~Zhou, X.~Zhu, X.~Song, H.~Li, and W.~Tao, ``Lif-seg: Lidar and camera image fusion for 3d lidar semantic segmentation,'' \emph{IEEE Transactions on Multimedia}, pp. 1--11, 2023.

\bibitem{Jaritz9737217}
M.~Jaritz, T.-H. Vu, R.~de~Charette, Ã.~Wirbel, and P.~Pérez, ``Cross-modal learning for domain adaptation in 3d semantic segmentation,'' \emph{IEEE Transactions on Pattern Analysis and Machine Intelligence}, vol.~45, no.~2, pp. 1533--1544, 2023.

\bibitem{Saltori196206}
C.~Saltori, A.~Osep, E.~Ricci, and L.~Leal-Taix\'e, ``Walking your lidog: A journey through multiple domains for lidar semantic segmentation,'' in \emph{Proceedings of the IEEE/CVF International Conference on Computer Vision (ICCV)}, October 2023, pp. 196--206.

\bibitem{wu2024632}
J.~Wu, X.~Li, S.~Xu, H.~Yuan, H.~Ding, Y.~Yang, X.~Li, J.~Zhang, Y.~Tong, X.~Jiang \emph{et~al.}, ``Towards open vocabulary learning: A survey,'' \emph{IEEE Transactions on Pattern Analysis and Machine Intelligence}, 2024.

\bibitem{Ding70107019}
R.~Ding, J.~Yang, C.~Xue, W.~Zhang, S.~Bai, and X.~Qi, ``Pla: Language-driven open-vocabulary 3d scene understanding,'' in \emph{Proceedings of the IEEE/CVF Conference on Computer Vision and Pattern Recognition (CVPR)}, June 2023, pp. 7010--7019.

\bibitem{Zhao9577428}
N.~Zhao, T.-S. Chua, and G.~H. Lee, ``Few-shot 3d point cloud semantic segmentation,'' in \emph{2021 IEEE/CVF Conference on Computer Vision and Pattern Recognition (CVPR)}, 2021, pp. 8869--8878.

\bibitem{Yang1158611596}
Y.~Yang, M.~Hayat, Z.~Jin, H.~Zhu, and Y.~Lei, ``Zero-shot point cloud segmentation by semantic-visual aware synthesis,'' in \emph{Proceedings of the IEEE/CVF International Conference on Computer Vision (ICCV)}, October 2023, pp. 11\,586--11\,596.

\bibitem{He10138737}
S.~He, X.~Jiang, W.~Jiang, and H.~Ding, ``Prototype adaption and projection for few- and zero-shot 3d point cloud semantic segmentation,'' \emph{IEEE Transactions on Image Processing}, vol.~32, pp. 3199--3211, 2023.

\bibitem{french128135}
R.~M. French, ``Catastrophic forgetting in connectionist networks,'' \emph{Trends in cognitive sciences}, vol.~3, no.~4, pp. 128--135, 1999.

\bibitem{robins123146}
A.~Robins, ``Catastrophic forgetting, rehearsal and pseudorehearsal,'' \emph{Connection Science}, vol.~7, no.~2, pp. 123--146, 1995.

\bibitem{Yang10204829}
Y.~Yang, M.~Hayat, Z.~Jin, C.~Ren, and Y.~Lei, ``Geometry and uncertainty-aware 3d point cloud class-incremental semantic segmentation,'' in \emph{2023 IEEE/CVF Conference on Computer Vision and Pattern Recognition (CVPR)}, 2023, pp. 21\,759--21\,768.

\bibitem{Yang1860118612}
Z.~Yang, R.~Li, E.~Ling, C.~Zhang, Y.~Wang, D.~Huang, K.~T. Ma, M.~Hur, and G.~Lin, ``Label-guided knowledge distillation for continual semantic segmentation on 2d images and 3d point clouds,'' in \emph{Proceedings of the IEEE/CVF International Conference on Computer Vision (ICCV)}, October 2023, pp. 18\,601--18\,612.

\bibitem{Yang19829832}
J.~Yang, R.~Ding, W.~Deng, Z.~Wang, and X.~Qi, ``Regionplc: Regional point-language contrastive learning for open-world 3d scene understanding,'' in \emph{Proceedings of the IEEE/CVF Conference on Computer Vision and Pattern Recognition (CVPR)}, June 2024, pp. 19\,823--19\,832.

\bibitem{Riz10203892}
L.~Riz, C.~Saltori, E.~Ricci, and F.~Poiesi, ``Novel class discovery for 3d point cloud semantic segmentation,'' in \emph{2023 IEEE/CVF Conference on Computer Vision and Pattern Recognition (CVPR)}, 2023, pp. 9393--9402.

\bibitem{Kong10205234}
L.~Kong, J.~Ren, L.~Pan, and Z.~Liu, ``Lasermix for semi-supervised lidar semantic segmentation,'' in \emph{2023 IEEE/CVF Conference on Computer Vision and Pattern Recognition (CVPR)}, 2023, pp. 21\,706--21\,716.

\bibitem{Li10203638}
L.~Li, H.~P.~H. Shum, and T.~P. Breckon, ``Less is more: Reducing task and model complexity for 3d point cloud semantic segmentation,'' in \emph{2023 IEEE/CVF Conference on Computer Vision and Pattern Recognition (CVPR)}, 2023, pp. 9361--9371.

\bibitem{Xu1809818108}
Z.~Xu, B.~Yuan, S.~Zhao, Q.~Zhang, and X.~Gao, ``Hierarchical point-based active learning for semi-supervised point cloud semantic segmentation,'' in \emph{Proceedings of the IEEE/CVF International Conference on Computer Vision (ICCV)}, October 2023, pp. 18\,098--18\,108.

\bibitem{Li9879771}
M.~Li, Y.~Xie, Y.~Shen, B.~Ke, R.~Qiao, B.~Ren, S.~Lin, and L.~Ma, ``Hybridcr: Weakly-supervised 3d point cloud semantic segmentation via hybrid contrastive regularization,'' in \emph{2022 IEEE/CVF Conference on Computer Vision and Pattern Recognition (CVPR)}, 2022, pp. 14\,910--14\,919.

\bibitem{Liu1841318422}
L.~Liu, Z.~Zhuang, S.~Huang, X.~Xiao, T.~Xiang, C.~Chen, J.~Wang, and M.~Tan, ``Cpcm: Contextual point cloud modeling for weakly-supervised point cloud semantic segmentation,'' in \emph{Proceedings of the IEEE/CVF International Conference on Computer Vision (ICCV)}, October 2023, pp. 18\,413--18\,422.

\bibitem{Tao9833393}
A.~Tao, Y.~Duan, Y.~Wei, J.~Lu, and J.~Zhou, ``Seggroup: Seg-level supervision for 3d instance and semantic segmentation,'' \emph{IEEE Transactions on Image Processing}, vol.~31, pp. 4952--4965, 2022.

\bibitem{Chang10035004}
X.~Chang, H.~Pan, W.~Sun, and H.~Gao, ``A multi-phase camera-lidar fusion network for 3d semantic segmentation with weak supervision,'' \emph{IEEE Transactions on Circuits and Systems for Video Technology}, vol.~33, no.~8, pp. 3737--3746, 2023.

\bibitem{Wei10328634}
J.~Wei, G.~Lin, K.-H. Yap, F.~Liu, and T.-Y. Hung, ``Dense supervision propagation for weakly supervised semantic segmentation on 3d point clouds,'' \emph{IEEE Transactions on Circuits and Systems for Video Technology}, pp. 1--1, 2024.

\bibitem{Zhang10203698}
Z.~Zhang, B.~Yang, B.~Wang, and B.~Li, ``Growsp: Unsupervised semantic segmentation of 3d point clouds,'' in \emph{2023 IEEE/CVF Conference on Computer Vision and Pattern Recognition (CVPR)}, 2023, pp. 17\,619--17\,629.

\bibitem{Rao9736689}
Y.~Rao, J.~Lu, and J.~Zhou, ``Pointglr: Unsupervised structural representation learning of 3d point clouds,'' \emph{IEEE Transactions on Pattern Analysis and Machine Intelligence}, vol.~45, no.~2, pp. 2193--2207, 2023.

\bibitem{poux2019213}
F.~Poux and R.~Billen, ``Voxel-based 3d point cloud semantic segmentation: Unsupervised geometric and relationship featuring vs deep learning methods,'' \emph{ISPRS International Journal of Geo-Information}, vol.~8, no.~5, p. 213, 2019.

\bibitem{liu37593768}
J.~Liu, Z.~Yu, T.~P. Breckon, and H.~P. Shum, ``U3ds3: Unsupervised 3d semantic scene segmentation,'' in \emph{Proceedings of the IEEE/CVF Winter Conference on Applications of Computer Vision}, 2024, pp. 3759--3768.

\bibitem{Yang9102769}
K.~Yang, S.~Bi, and M.~Dong, ``Lightningnet: Fast and accurate semantic segmentation for autonomous driving based on 3d lidar point cloud,'' in \emph{2020 IEEE International Conference on Multimedia and Expo (ICME)}, 2020, pp. 1--6.

\bibitem{Razani9561171}
R.~Razani, R.~Cheng, E.~Taghavi, and L.~Bingbing, ``Lite-hdseg: Lidar semantic segmentation using lite harmonic dense convolutions,'' in \emph{2021 IEEE International Conference on Robotics and Automation (ICRA)}, 2021, pp. 9550--9556.

\bibitem{kang20211960}
D.~Kang, A.~Wong, B.~Lee, and J.~Kim, ``Real-time semantic segmentation of 3d point cloud for autonomous driving,'' \emph{Electronics}, vol.~10, no.~16, p. 1960, 2021.

\bibitem{Weng9583294}
X.~Weng, Y.~Yan, S.~Chen, J.-H. Xue, and H.~Wang, ``Stage-aware feature alignment network for real-time semantic segmentation of street scenes,'' \emph{IEEE Transactions on Circuits and Systems for Video Technology}, vol.~32, no.~7, pp. 4444--4459, 2022.

\bibitem{yu325341}
C.~Yu, J.~Wang, C.~Peng, C.~Gao, G.~Yu, and N.~Sang, ``Bisenet: Bilateral segmentation network for real-time semantic segmentation,'' in \emph{Proceedings of the European conference on computer vision (ECCV)}, 2018, pp. 325--341.

\bibitem{Li8954459}
H.~Li, P.~Xiong, H.~Fan, and J.~Sun, ``Dfanet: Deep feature aggregation for real-time semantic segmentation,'' in \emph{2019 IEEE/CVF Conference on Computer Vision and Pattern Recognition (CVPR)}, 2019, pp. 9514--9523.

\bibitem{3dg18030}
D.~Group \emph{et~al.}, ``Text of iso/iec cd 23090-5: Video-based point cloud compression,'' \emph{ISO/IEC JTC1/SC29/WG11 Doc. N18030}.

\bibitem{3dg2019text}
------, ``Text of iso/iec cd 23090-9 geometry-based point cloud compression,'' \emph{ISO/IEC JTC1/SC29/WG11 Doc. N18478}, 2019.

\bibitem{Cao9457097}
C.~Cao, M.~Preda, V.~Zakharchenko, E.~S. Jang, and T.~Zaharia, ``Compression of sparse and dense dynamic point clouds—methods and standards,'' \emph{Proceedings of the IEEE}, vol. 109, no.~9, pp. 1537--1558, 2021.

\bibitem{Zhang10032603}
P.~Zhang, S.~Wang, M.~Wang, J.~Li, X.~Wang, and S.~Kwong, ``Rethinking semantic image compression: Scalable representation with cross-modality transfer,'' \emph{IEEE Transactions on Circuits and Systems for Video Technology}, vol.~33, no.~8, pp. 4441--4445, 2023.

\bibitem{Akbari8683541}
M.~Akbari, J.~Liang, and J.~Han, ``Dsslic: Deep semantic segmentation-based layered image compression,'' in \emph{ICASSP 2019 - 2019 IEEE International Conference on Acoustics, Speech and Signal Processing (ICASSP)}, 2019, pp. 2042--2046.

\bibitem{Huang9894405}
Z.~Huang, C.~Jia, S.~Wang, and S.~Ma, ``Hmfvc: A human-machine friendly video compression scheme,'' \emph{IEEE Transactions on Circuits and Systems for Video Technology}, pp. 1--1, 2022.

\bibitem{Akhtar10380494}
A.~Akhtar, Z.~Li, and G.~Van~der Auwera, ``Inter-frame compression for dynamic point cloud geometry coding,'' \emph{IEEE Transactions on Image Processing}, vol.~33, pp. 584--594, 2024.

\bibitem{Gao10416804}
L.~Gao, Z.~Li, L.~Hou, Y.~Xu, and J.~Sun, ``Occupancy-assisted attribute artifact reduction for video-based point cloud compression,'' \emph{IEEE Transactions on Broadcasting}, pp. 1--14, 2024.

\bibitem{Song10205051}
R.~Song, C.~Fu, S.~Liu, and G.~Li, ``Efficient hierarchical entropy model for learned point cloud compression,'' in \emph{2023 IEEE/CVF Conference on Computer Vision and Pattern Recognition (CVPR)}, 2023, pp. 14\,368--14\,377.

\bibitem{Ngu10024999}
D.~T. Nguyen and A.~Kaup, ``Lossless point cloud geometry and attribute compression using a learned conditional probability model,'' \emph{IEEE Transactions on Circuits and Systems for Video Technology}, vol.~33, no.~8, pp. 4337--4348, 2023.

\bibitem{Zhou9878865}
X.~Zhou, C.~R. Qi, Y.~Zhou, and D.~Anguelov, ``Riddle: Lidar data compression with range image deep delta encoding,'' in \emph{2022 IEEE/CVF Conference on Computer Vision and Pattern Recognition (CVPR)}, 2022, pp. 17\,191--17\,200.

\bibitem{Ma10222524}
X.~Ma, Y.~Xu, X.~Zhang, L.~Tang, K.~Zhang, and L.~Zhang, ``Hm-pcgc: A human-machine balanced point cloud geometry compression scheme,'' in \emph{2023 IEEE International Conference on Image Processing (ICIP)}, 2023, pp. 2265--2269.

\bibitem{Liu10219641}
L.~Liu, Z.~Hu, and J.~Zhang, ``Pchm-net: A new point cloud compression framework for both human vision and machine vision,'' in \emph{2023 IEEE International Conference on Multimedia and Expo (ICME)}, 2023, pp. 1997--2002.

\bibitem{Sun9944923}
X.~Sun, M.~Wang, J.~Du, Y.~Sun, S.~S. Cheng, and W.~Xie, ``A task-driven scene-aware lidar point cloud coding framework for autonomous vehicles,'' \emph{IEEE Transactions on Industrial Informatics}, vol.~19, no.~8, pp. 8731--8742, 2023.

\bibitem{Liu3654800}
\BIBentryALTinterwordspacing
S.~Liu, W.~Lin, Y.~Chen, Y.~Zhang, W.~Dai, J.~See, and H.-K. Xiong, ``A unified framework for jointly compressing visual and semantic data,'' \emph{ACM Trans. Multimedia Comput. Commun. Appl.}, vol.~20, no.~7, may 2024. [Online]. Available: \url{https://doi.org/10.1145/3654800}
\BIBentrySTDinterwordspacing

\bibitem{schwarz2017766}
S.~Schwarz, G.~Martin-Cocher, D.~Flynn, and M.~Budagavi, ``Common test conditions for point cloud compression,'' \emph{Document ISO/IEC JTC1/SC29/WG11 w17766, Ljubljana, Slovenia}, 2018.

\bibitem{Besl121791}
P.~Besl and N.~D. McKay, ``A method for registration of 3-d shapes,'' \emph{IEEE Transactions on Pattern Analysis and Machine Intelligence}, vol.~14, no.~2, pp. 239--256, 1992.

\bibitem{Lei7918612}
H.~Lei, G.~Jiang, and L.~Quan, ``Fast descriptors and correspondence propagation for robust global point cloud registration,'' \emph{IEEE Transactions on Image Processing}, vol.~26, no.~8, pp. 3614--3623, 2017.

\bibitem{Bowman7989203}
S.~L. Bowman, N.~Atanasov, K.~Daniilidis, and G.~J. Pappas, ``Probabilistic data association for semantic slam,'' in \emph{2017 IEEE International Conference on Robotics and Automation (ICRA)}, 2017, pp. 1722--1729.

\bibitem{Hu9561140}
L.~Hu, J.~Wei, Z.~Ouyang, and L.~Kneip, ``Point set registration with semantic region association using cascaded expectation maximization,'' in \emph{2021 IEEE International Conference on Robotics and Automation (ICRA)}, 2021, pp. 11\,234--11\,240.

\bibitem{Yin10160798}
P.~Yin, S.~Yuan, H.~Cao, X.~Ji, S.~Zhang, and L.~Xie, ``Segregator: Global point cloud registration with semantic and geometric cues,'' in \emph{2023 IEEE International Conference on Robotics and Automation (ICRA)}, 2023, pp. 2848--2854.

\bibitem{Zhang9561929}
R.~Zhang, T.-Y. Lin, C.~E. Lin, S.~A. Parkison, W.~Clark, J.~W. Grizzle, R.~M. Eustice, and M.~Ghaffari, ``A new framework for registration of semantic point clouds from stereo and rgb-d cameras,'' in \emph{2021 IEEE International Conference on Robotics and Automation (ICRA)}, 2021, pp. 12\,214--12\,221.

\bibitem{Qiao10341394}
Z.~Qiao, Z.~Yu, H.~Yin, and S.~Shen, ``Pyramid semantic graph-based global point cloud registration with low overlap,'' in \emph{2023 IEEE/RSJ International Conference on Intelligent Robots and Systems (IROS)}, 2023, pp. 11\,202--11\,209.

\bibitem{besl586606}
P.~J. Besl and N.~D. McKay, ``Method for registration of 3-d shapes,'' in \emph{Sensor fusion IV: control paradigms and data structures}, vol. 1611.\hskip 1em plus 0.5em minus 0.4em\relax Spie, 1992, pp. 586--606.

\bibitem{Wang9282862}
Q.~Wang, Y.~Yang, T.~Wan, and S.~Du, ``Robust point set registration based on semantic information,'' in \emph{2020 IEEE International Conference on Systems, Man, and Cybernetics (SMC)}, 2020, pp. 2553--2558.

\bibitem{Zaganidis8387438}
A.~Zaganidis, L.~Sun, T.~Duckett, and G.~Cielniak, ``Integrating deep semantic segmentation into 3-d point cloud registration,'' \emph{IEEE Robotics and Automation Letters}, vol.~3, no.~4, pp. 2942--2949, 2018.

\bibitem{Li202109306}
\BIBentryALTinterwordspacing
Q.~Li, C.~Wang, C.~Wen, and X.~Li, ``Deepsir: Deep semantic iterative registration for lidar point clouds,'' \emph{Pattern Recognit.}, vol. 137, p. 109306, 2023. [Online]. Available: \url{https://api.semanticscholar.org/CorpusID:255662237}
\BIBentrySTDinterwordspacing

\bibitem{Truong8945870}
G.~Truong, S.~Z. Gilani, S.~M.~S. Islam, and D.~Suter, ``Fast point cloud registration using semantic segmentation,'' in \emph{2019 Digital Image Computing: Techniques and Applications (DICTA)}, 2019, pp. 1--8.

\bibitem{Fan8099747}
H.~Fan, H.~Su, and L.~Guibas, ``A point set generation network for 3d object reconstruction from a single image,'' in \emph{2017 IEEE Conference on Computer Vision and Pattern Recognition (CVPR)}, 2017, pp. 2463--2471.

\bibitem{Hutt232073}
D.~Huttenlocher, G.~Klanderman, and W.~Rucklidge, ``Comparing images using the hausdorff distance,'' \emph{IEEE Transactions on Pattern Analysis and Machine Intelligence}, vol.~15, no.~9, pp. 850--863, 1993.

\bibitem{Pomerleau1393272}
F.~Pomerleau, F.~Colas, R.~Siegwart, and S.~Magnenat, ``Comparing icp variants on real-world data sets,'' \emph{Auton. Robots}, vol.~34, no.~3, p. 133–148, apr 2013.

\bibitem{Choy9009829}
C.~Choy, J.~Park, and V.~Koltun, ``Fully convolutional geometric features,'' in \emph{2019 IEEE/CVF International Conference on Computer Vision (ICCV)}, 2019, pp. 8957--8965.

\bibitem{Huang9662197}
S.-S. Huang, H.~Chen, J.~Huang, H.~Fu, and S.-M. Hu, ``Real-time globally consistent 3d reconstruction with semantic priors,'' \emph{IEEE Transactions on Visualization and Computer Graphics}, vol.~29, no.~4, pp. 1977--1991, 2023.

\bibitem{Zheng9286413}
T.~Zheng, G.~Zhang, L.~Han, L.~Xu, and L.~Fang, ``Buildingfusion: Semantic-aware structural building-scale 3d reconstruction,'' \emph{IEEE Transactions on Pattern Analysis and Machine Intelligence}, vol.~44, no.~5, pp. 2328--2345, 2022.

\bibitem{Qu10219715}
Y.~Qu, Y.~Wang, and Y.~Qi, ``Sg-nerf: Semantic-guided point-based neural radiance fields,'' in \emph{2023 IEEE International Conference on Multimedia and Expo (ICME)}, 2023, pp. 570--575.

\bibitem{Xie1799218002}
B.~Xie, B.~Li, Z.~Zhang, J.~Dong, X.~Jin, J.~Yang, and W.~Zeng, ``Navinerf: Nerf-based 3d representation disentanglement by latent semantic navigation,'' in \emph{Proceedings of the IEEE/CVF International Conference on Computer Vision (ICCV)}, October 2023, pp. 17\,992--18\,002.

\bibitem{Wu9578559}
S.-C. Wu, J.~Wald, K.~Tateno, N.~Navab, and F.~Tombari, ``Scenegraphfusion: Incremental 3d scene graph prediction from rgb-d sequences,'' in \emph{2021 IEEE/CVF Conference on Computer Vision and Pattern Recognition (CVPR)}, 2021, pp. 7511--7521.

\bibitem{Zhang9578123}
C.~Zhang, J.~Yu, Y.~Song, and W.~Cai, ``Exploiting edge-oriented reasoning for 3d point-based scene graph analysis,'' in \emph{2021 IEEE/CVF Conference on Computer Vision and Pattern Recognition (CVPR)}, 2021, pp. 9700--9710.

\bibitem{Zhang209555403}
\BIBentryALTinterwordspacing
S.~Zhang, s.~li, A.~Hao, and H.~Qin, ``Knowledge-inspired 3d scene graph prediction in point cloud,'' in \emph{Advances in Neural Information Processing Systems}, M.~Ranzato, A.~Beygelzimer, Y.~Dauphin, P.~Liang, and J.~W. Vaughan, Eds., vol.~34.\hskip 1em plus 0.5em minus 0.4em\relax Curran Associates, Inc., 2021, pp. 18\,620--18\,632. [Online]. Available: \url{https://proceedings.neurips.cc/paper_files/paper/2021/file/9a555403384fc12f931656dea910e334-Paper.pdf}
\BIBentrySTDinterwordspacing

\bibitem{wald20630651}
J.~Wald, N.~Navab, and F.~Tombari, ``Learning 3d semantic scene graphs with instance embeddings,'' \emph{International Journal of Computer Vision}, vol. 130, no.~3, pp. 630--651, 2022.

\bibitem{koch34043414}
S.~Koch, P.~Hermosilla, N.~Vaskevicius, M.~Colosi, and T.~Ropinski, ``Sgrec3d: Self-supervised 3d scene graph learning via object-level scene reconstruction,'' in \emph{Proceedings of the IEEE/CVF Winter Conference on Applications of Computer Vision}, 2024, pp. 3404--3414.

\bibitem{Yu10187165}
T.~Yu, X.~Lin, S.~Wang, W.~Sheng, Q.~Huang, and J.~Yu, ``A comprehensive survey of 3d dense captioning: Localizing and describing objects in 3d scenes,'' \emph{IEEE Transactions on Circuits and Systems for Video Technology}, pp. 1--1, 2023.

\bibitem{Dhamo9710451}
H.~Dhamo, F.~Manhardt, N.~Navab, and F.~Tombari, ``Graph-to-3d: End-to-end generation and manipulation of 3d scenes using scene graphs,'' in \emph{2021 IEEE/CVF International Conference on Computer Vision (ICCV)}, 2021, pp. 16\,332--16\,341.

\bibitem{zhai202436512}
G.~Zhai, E.~P. {\"O}rnek, S.-C. Wu, Y.~Di, F.~Tombari, N.~Navab, and B.~Busam, ``Commonscenes: Generating commonsense 3d indoor scenes with scene graphs,'' \emph{Advances in Neural Information Processing Systems}, vol.~36, 2024.

\bibitem{Kolm9880174}
M.~Kolmet, Q.~Zhou, A.~Ošep, and L.~Leal-Taixé, ``Text2pos: Text-to-point-cloud cross-modal localization,'' in \emph{2022 IEEE/CVF Conference on Computer Vision and Pattern Recognition (CVPR)}, 2022, pp. 6677--6686.

\bibitem{Chen9577651}
D.~Z. Chen, A.~Gholami, M.~Nießner, and A.~X. Chang, ``Scan2cap: Context-aware dense captioning in rgb-d scans,'' in \emph{2021 IEEE/CVF Conference on Computer Vision and Pattern Recognition (CVPR)}, 2021, pp. 3192--3202.

\bibitem{jiao528545}
Y.~Jiao, S.~Chen, Z.~Jie, J.~Chen, L.~Ma, and Y.-G. Jiang, ``More: Multi-order relation mining for dense captioning in 3d scenes,'' in \emph{European Conference on Computer Vision}.\hskip 1em plus 0.5em minus 0.4em\relax Springer, 2022, pp. 528--545.

\bibitem{Feng9710755}
M.~Feng, Z.~Li, Q.~Li, L.~Zhang, X.~Zhang, G.~Zhu, H.~Zhang, Y.~Wang, and A.~Mian, ``Free-form description guided 3d visual graph network for object grounding in point cloud,'' in \emph{2021 IEEE/CVF International Conference on Computer Vision (ICCV)}, 2021, pp. 3702--3711.

\bibitem{wang220410688}
H.~Wang, C.~Zhang, J.~Yu, and W.~Cai, ``Spatiality-guided transformer for 3d dense captioning on point clouds,'' \emph{arXiv preprint arXiv:2204.10688}, 2022.

\bibitem{Yuan9879338}
Z.~Yuan, X.~Yan, Y.~Liao, Y.~Guo, G.~Li, S.~Cui, and Z.~Li, ``X -trans2cap: Cross-modal knowledge transfer using transformer for 3d dense captioning,'' in \emph{2022 IEEE/CVF Conference on Computer Vision and Pattern Recognition (CVPR)}, 2022, pp. 8553--8563.

\bibitem{Vedantam7299087}
R.~Vedantam, C.~L. Zitnick, and D.~Parikh, ``Cider: Consensus-based image description evaluation,'' in \emph{2015 IEEE Conference on Computer Vision and Pattern Recognition (CVPR)}, 2015, pp. 4566--4575.

\bibitem{papineni311318}
K.~Papineni, S.~Roukos, T.~Ward, and W.-J. Zhu, ``Bleu: a method for automatic evaluation of machine translation,'' in \emph{Proceedings of the 40th annual meeting of the Association for Computational Linguistics}, 2002, pp. 311--318.

\bibitem{lin20047481}
C.-Y. Lin, ``Rouge: A package for automatic evaluation of summaries,'' in \emph{Text summarization branches out}, 2004, pp. 74--81.

\bibitem{banerjee6572}
S.~Banerjee and A.~Lavie, ``Meteor: An automatic metric for mt evaluation with improved correlation with human judgments,'' in \emph{Proceedings of the acl workshop on intrinsic and extrinsic evaluation measures for machine translation and/or summarization}, 2005, pp. 65--72.

\bibitem{Uy8578568}
M.~A. Uy and G.~H. Lee, ``Pointnetvlad: Deep point cloud based retrieval for large-scale place recognition,'' in \emph{2018 IEEE/CVF Conference on Computer Vision and Pattern Recognition}, 2018, pp. 4470--4479.

\bibitem{Wang250109023}
\BIBentryALTinterwordspacing
G.~Wang, H.~Fan, and M.~Kankanhalli, ``Text to point cloud localization with relation-enhanced transformer,'' \emph{Proceedings of the AAAI Conference on Artificial Intelligence}, vol.~37, no.~2, pp. 2501--2509, Jun. 2023. [Online]. Available: \url{https://ojs.aaai.org/index.php/AAAI/article/view/25347}
\BIBentrySTDinterwordspacing

\bibitem{Song8099511}
S.~Song, F.~Yu, A.~Zeng, A.~X. Chang, M.~Savva, and T.~Funkhouser, ``Semantic scene completion from a single depth image,'' in \emph{2017 IEEE Conference on Computer Vision and Pattern Recognition (CVPR)}, 2017, pp. 190--198.

\bibitem{Zhang9008381}
P.~Zhang, W.~Liu, Y.~Lei, H.~Lu, and X.~Yang, ``Cascaded context pyramid for full-resolution 3d semantic scene completion,'' in \emph{2019 IEEE/CVF International Conference on Computer Vision (ICCV)}, 2019, pp. 7800--7809.

\bibitem{Chen9156418}
X.~Chen, K.-Y. Lin, C.~Qian, G.~Zeng, and H.~Li, ``3d sketch-aware semantic scene completion via semi-supervised structure prior,'' in \emph{2020 IEEE/CVF Conference on Computer Vision and Pattern Recognition (CVPR)}, 2020, pp. 4192--4201.

\bibitem{miao230213540}
R.~Miao, W.~Liu, M.~Chen, Z.~Gong, W.~Xu, C.~Hu, and S.~Zhou, ``Occdepth: A depth-aware method for 3d semantic scene completion,'' 2023.

\bibitem{li1140211409}
S.~Li, C.~Zou, Y.~Li, X.~Zhao, and Y.~Gao, ``Attention-based multi-modal fusion network for semantic scene completion,'' in \emph{Proceedings of the AAAI Conference on Artificial Intelligence}, vol.~34, no.~07, 2020, pp. 11\,402--11\,409.

\bibitem{Garb9025517}
M.~Garbade, Y.-T. Chen, J.~Sawatzky, and J.~Gall, ``Two stream 3d semantic scene completion,'' in \emph{2019 IEEE/CVF Conference on Computer Vision and Pattern Recognition Workshops (CVPRW)}, 2019, pp. 416--425.

\bibitem{Liu20189872}
\BIBentryALTinterwordspacing
S.~Liu, Y.~HU, Y.~Zeng, Q.~Tang, B.~Jin, Y.~Han, and X.~Li, ``See and think: Disentangling semantic scene completion,'' in \emph{Advances in Neural Information Processing Systems}, S.~Bengio, H.~Wallach, H.~Larochelle, K.~Grauman, N.~Cesa-Bianchi, and R.~Garnett, Eds., vol.~31.\hskip 1em plus 0.5em minus 0.4em\relax Curran Associates, Inc., 2018. [Online]. Available: \url{https://proceedings.neurips.cc/paper_files/paper/2018/file/9872ed9fc22fc182d371c3e9ed316094-Paper.pdf}
\BIBentrySTDinterwordspacing

\bibitem{Wang1031410323}
F.~Wang, Q.~Sun, D.~Zhang, and J.~Tang, ``Unleashing network potentials for semantic scene completion,'' in \emph{Proceedings of the IEEE/CVF Conference on Computer Vision and Pattern Recognition (CVPR)}, June 2024, pp. 10\,314--10\,323.

\bibitem{Jiang202520267}
H.~Jiang, T.~Cheng, N.~Gao, H.~Zhang, T.~Lin, W.~Liu, and X.~Wang, ``Symphonize 3d semantic scene completion with contextual instance queries,'' in \emph{Proceedings of the IEEE/CVF Conference on Computer Vision and Pattern Recognition (CVPR)}, June 2024, pp. 20\,258--20\,267.

\bibitem{Cheng21482161}
R.~Cheng, C.~Agia, Y.~Ren, X.~Li, and L.~Bingbing, ``S3cnet: A sparse semantic scene completion network for lidar point clouds,'' in \emph{Proceedings of the 2020 Conference on Robot Learning}, ser. Proceedings of Machine Learning Research, J.~Kober, F.~Ramos, and C.~Tomlin, Eds., vol. 155, 16--18 Nov 2021, pp. 2148--2161.

\bibitem{An10409585}
P.~An, D.~Zhu, S.~Quan, J.~Ding, J.~Ma, Y.~Yang, and Q.~Liu, ``Esc-net: Alleviating triple sparsity on 3d lidar point clouds for extreme sparse scene completion,'' \emph{IEEE Transactions on Multimedia}, pp. 1--12, 2024.

\bibitem{Rist9477025}
C.~B. Rist, D.~Emmerichs, M.~Enzweiler, and D.~M. Gavrila, ``Semantic scene completion using local deep implicit functions on lidar data,'' \emph{IEEE Transactions on Pattern Analysis and Machine Intelligence}, vol.~44, no.~10, pp. 7205--7218, 2022.

\bibitem{Li2189421904}
H.~Li, J.~Dong, B.~Wen, M.~Gao, T.~Huang, Y.-H. Liu, and D.~Cremers, ``Ddit: Semantic scene completion via deformable deep implicit templates,'' in \emph{Proceedings of the IEEE/CVF International Conference on Computer Vision (ICCV)}, October 2023, pp. 21\,894--21\,904.

\bibitem{Xia10203998}
Z.~Xia, Y.~Liu, X.~Li, X.~Zhu, Y.~Ma, Y.~Li, Y.~Hou, and Y.~Qiao, ``Scpnet: Semantic scene completion on point cloud,'' in \emph{2023 IEEE/CVF Conference on Computer Vision and Pattern Recognition (CVPR)}, 2023, pp. 17\,642--17\,651.

\bibitem{Dong88748883}
H.~Dong, E.~Ma, L.~Wang, M.~Wang, W.~Xie, Q.~Guo, P.~Li, L.~Liang, K.~Yang, and D.~Lin, ``Cvsformer: Cross-view synthesis transformer for semantic scene completion,'' in \emph{Proceedings of the IEEE/CVF International Conference on Computer Vision (ICCV)}, October 2023, pp. 8874--8883.

\bibitem{Li10203337}
Y.~Li, Z.~Yu, C.~Choy, C.~Xiao, J.~M. Alvarez, S.~Fidler, C.~Feng, and A.~Anandkumar, ``Voxformer: Sparse voxel transformer for camera-based 3d semantic scene completion,'' in \emph{2023 IEEE/CVF Conference on Computer Vision and Pattern Recognition (CVPR)}, 2023, pp. 9087--9098.

\bibitem{Yao94559465}
J.~Yao, C.~Li, K.~Sun, Y.~Cai, H.~Li, W.~Ouyang, and H.~Li, ``Ndc-scene: Boost monocular 3d semantic scene completion in normalized device coordinates space,'' in \emph{Proceedings of the IEEE/CVF International Conference on Computer Vision (ICCV)}, October 2023, pp. 9455--9465.

\bibitem{Xiao10379527}
H.~Xiao, H.~Xu, W.~Kang, and Y.~Li, ``Instance-aware monocular 3d semantic scene completion,'' \emph{IEEE Transactions on Intelligent Transportation Systems}, pp. 1--12, 2024.

\bibitem{Lee2833728347}
J.~Lee, S.~Lee, C.~Jo, W.~Im, J.~Seon, and S.-E. Yoon, ``Semcity: Semantic scene generation with triplane diffusion,'' in \emph{Proceedings of the IEEE/CVF Conference on Computer Vision and Pattern Recognition (CVPR)}, June 2024, pp. 28\,337--28\,347.

\bibitem{Pham8953238}
Q.-H. Pham, T.~Nguyen, B.-S. Hua, G.~Roig, and S.-K. Yeung, ``Jsis3d: Joint semantic-instance segmentation of 3d point clouds with multi-task pointwise networks and multi-value conditional random fields,'' in \emph{2019 IEEE/CVF Conference on Computer Vision and Pattern Recognition (CVPR)}, 2019, pp. 8819--8828.

\bibitem{Nie9578585}
Y.~Nie, J.~Hou, X.~Han, and M.~Nießner, ``Rfd-net: Point scene understanding by semantic instance reconstruction,'' in \emph{2021 IEEE/CVF Conference on Computer Vision and Pattern Recognition (CVPR)}, 2021, pp. 4606--4616.

\bibitem{Wang8953321}
X.~Wang, S.~Liu, X.~Shen, C.~Shen, and J.~Jia, ``Associatively segmenting instances and semantics in point clouds,'' in \emph{2019 IEEE/CVF Conference on Computer Vision and Pattern Recognition (CVPR)}, 2019, pp. 4091--4100.

\bibitem{Hu8590720}
S.-M. Hu, J.-X. Cai, and Y.-K. Lai, ``Semantic labeling and instance segmentation of 3d point clouds using patch context analysis and multiscale processing,'' \emph{IEEE Transactions on Visualization and Computer Graphics}, vol.~26, no.~7, pp. 2485--2498, 2020.

\bibitem{Zhao9932589}
L.~Zhao and W.~Tao, ``Jsnet++: Dynamic filters and pointwise correlation for 3d point cloud instance and semantic segmentation,'' \emph{IEEE Transactions on Circuits and Systems for Video Technology}, vol.~33, no.~4, pp. 1854--1867, 2023.

\bibitem{fang202414563}
Z.~Fang, X.~Li, X.~Li, J.~M. Buhmann, C.~C. Loy, and M.~Liu, ``Explore in-context learning for 3d point cloud understanding,'' \emph{Advances in Neural Information Processing Systems}, vol.~36, 2024.

\bibitem{Afham99029912}
M.~Afham, I.~Dissanayake, D.~Dissanayake, A.~Dharmasiri, K.~Thilakarathna, and R.~Rodrigo, ``Crosspoint: Self-supervised cross-modal contrastive learning for 3d point cloud understanding,'' in \emph{Proceedings of the IEEE/CVF Conference on Computer Vision and Pattern Recognition (CVPR)}, June 2022, pp. 9902--9912.

\bibitem{Radford87488763}
A.~Radford, J.~W. Kim, C.~Hallacy, A.~Ramesh, G.~Goh, S.~Agarwal, G.~Sastry, A.~Askell, P.~Mishkin, J.~Clark, G.~Krueger, and I.~Sutskever, ``Learning transferable visual models from natural language supervision,'' in \emph{Proceedings of the 38th International Conference on Machine Learning}, ser. Proceedings of Machine Learning Research, M.~Meila and T.~Zhang, Eds., vol. 139.\hskip 1em plus 0.5em minus 0.4em\relax PMLR, 18--24 Jul 2021, pp. 8748--8763.

\bibitem{Zhang9878980}
R.~Zhang, Z.~Guo, W.~Zhang, K.~Li, X.~Miao, B.~Cui, Y.~Qiao, P.~Gao, and H.~Li, ``Pointclip: Point cloud understanding by clip,'' in \emph{2022 IEEE/CVF Conference on Computer Vision and Pattern Recognition (CVPR)}, 2022, pp. 8542--8552.

\bibitem{Wu10147273}
Y.~Wu, J.~Liu, M.~Gong, P.~Gong, X.~Fan, A.~K. Qin, Q.~Miao, and W.~Ma, ``Self-supervised intra-modal and cross-modal contrastive learning for point cloud understanding,'' \emph{IEEE Transactions on Multimedia}, vol.~26, pp. 1626--1638, 2024.

\bibitem{DuanS238839}
\BIBentryALTinterwordspacing
L.~Duan, S.~Zhao, N.~Xue, M.~Gong, G.-S. Xia, and D.~Tao, ``Condaformer: Disassembled transformer with local structure enhancement for 3d point cloud understanding,'' in \emph{Advances in Neural Information Processing Systems}, A.~Oh, T.~Naumann, A.~Globerson, K.~Saenko, M.~Hardt, and S.~Levine, Eds., vol.~36.\hskip 1em plus 0.5em minus 0.4em\relax Curran Associates, Inc., 2023, pp. 23\,886--23\,901. [Online]. Available: \url{https://proceedings.neurips.cc/paper_files/paper/2023/file/4b4f1272c73d5afd222b6dd3391c3f77-Paper-Conference.pdf}
\BIBentrySTDinterwordspacing

\bibitem{Park2181823}
J.~Park, S.~Lee, S.~Kim, Y.~Xiong, and H.~J. Kim, ``Self-positioning point-based transformer for point cloud understanding,'' in \emph{Proceedings of the IEEE/CVF Conference on Computer Vision and Pattern Recognition (CVPR)}, June 2023, pp. 21\,814--21\,823.

\bibitem{zhang2403007}
\BIBentryALTinterwordspacing
T.~Zhang, X.~Li, H.~Yuan, S.~Ji, and S.~Yan, ``Point cloud mamba: Point cloud learning via state space model,'' 2024. [Online]. Available: \url{https://arxiv.org/abs/2403.00762}
\BIBentrySTDinterwordspacing

\bibitem{liu24036467}
\BIBentryALTinterwordspacing
J.~Liu, R.~Yu, Y.~Wang, Y.~Zheng, T.~Deng, W.~Ye, and H.~Wang, ``Point mamba: A novel point cloud backbone based on state space model with octree-based ordering strategy,'' 2024. [Online]. Available: \url{https://arxiv.org/abs/2403.06467}
\BIBentrySTDinterwordspacing

\end{thebibliography}
\vfill

\end{document}